\documentclass[runningheads]{llncs}

\usepackage{eccv}

\usepackage{eccvabbrv}

\usepackage{graphicx}
\usepackage{booktabs}
\usepackage{multirow}
\usepackage{colortbl}
\usepackage{amsmath}
\usepackage{subcaption}
\usepackage{wrapfig}
\usepackage{tabularx}
\usepackage{pifont}
\usepackage{xcolor}
\usepackage[bookmarks=true,linkcolor=orange,colorlinks = true, citecolor=orange]{hyperref}
\usepackage{tcolorbox}
\tcbuselibrary{skins, breakable}
\usepackage{float}

\definecolor{myblue}{rgb}{0.9, 0.95, 1}
\newcommand{\equalcontrib}{\textsuperscript{*}}
\newcommand{\corrauthor}{\textsuperscript{$\dagger$}}

\usepackage[accsupp]{axessibility}  

\usepackage{orcidlink}

\begin{document}

\title{Plug-and-Play Traffic Element Awareness for End-to-End Autonomous Driving} 

\titlerunning{TE-Aware End-to-End Autonomous Driving}

\author{Zongzheng Zhang\inst{1}\equalcontrib\orcidlink{0009-0007-6909-1587} \and
Jijun Wang\inst{1}\equalcontrib\orcidlink{0009-0004-8497-2030} \and
Saining Zhang\inst{1} \and 
Wang Shuo\inst{2} \and \\ 
Yiru Wang\inst{2} \and
Hai Yang\inst{2} \and
Yang Chen\inst{2} \and
Yuwen Heng \inst{2}\orcidlink{0000-0003-3793-4811} \and \\
Hao Sun\inst{2} \and
Anqing Jiang\inst{2}\orcidlink{0000-0001-9187-830X} \and
Hao Zhao\inst{1}\corrauthor\orcidlink{0000-0001-7903-581X}
}

\authorrunning{Z.~Zhang, Ji.~Wang et al.}

\institute{Institute for AI Industry Research (AIR), Tsinghua University, Beijing, China \and
Bosch Corporate Research, Shanghai, China\\
\equalcontrib Equal contribution.
\quad
\corrauthor Corresponding author.\\
\url{https://zzongzheng0918.github.io/TE-Aware-E2E-AD/} \\
}

\maketitle
\vspace{-4mm}
\begin{center}
    \captionsetup{type=figure}
    \includegraphics[width=1.0\linewidth]{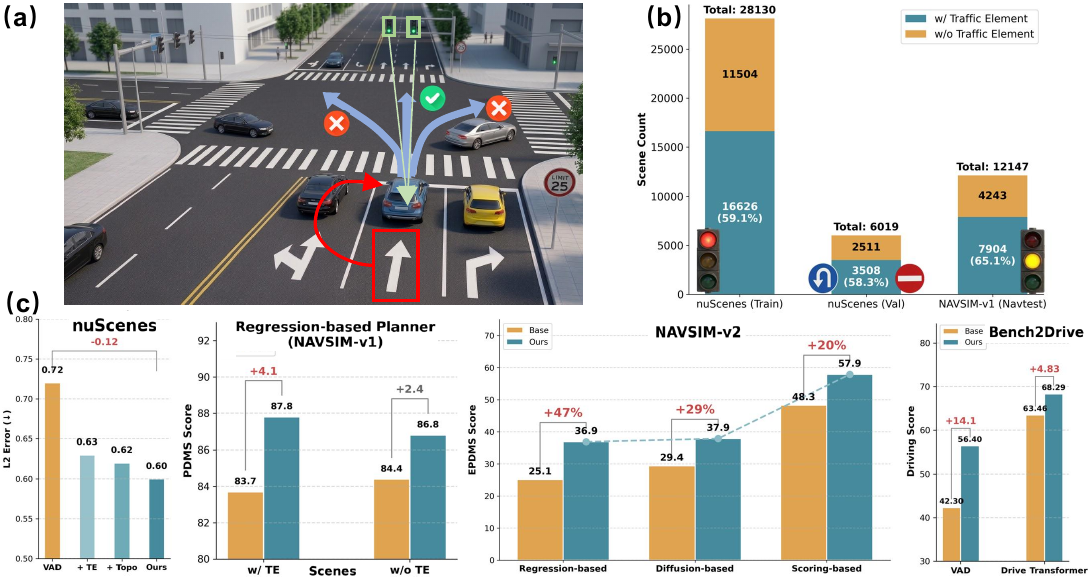}
    \vspace{-6mm}
    \captionof{figure}{\textbf{TE-Aware End-to-End Planning.} (a) In complex intersections, correct planning requires recognizing rule-critical traffic elements to avoid topologically plausible but rule-violating maneuvers. (b) Traffic elements are \textbf{pervasive}: across three benchmarks, over 55\% of scenes contain at least one traffic element. (c) Leveraging traffic elements (and topology) yields \textbf{consistent improvements} across representative end-to-end planners in both open-loop and closed-loop evaluations.
    }\label{fig:teaser}
\end{center}
\vspace{-7mm}

\begin{abstract}
   Traffic elements such as traffic lights and road signs play a fundamental role in human driving decisions and should naturally influence end-to-end driving performance. However, existing end-to-end driving research predominantly focuses on dynamic road participants (e.g., vehicles and pedestrians), while the role of traffic elements remains largely unexplored. To date, the community lacks a systematic study quantifying how traffic elements affect end-to-end driving models. This gap stems from two main challenges: first, existing public datasets rarely provide structured annotations for traffic elements; second, modern end-to-end driving systems vary widely in architectures and training paradigms, making conclusions drawn from a single method difficult to generalize. In this work, we present the first systematic investigation of \textbf{traffic element awareness} for end-to-end autonomous driving. We begin by constructing a unified research infrastructure by augmenting multiple public driving datasets with comprehensive traffic element annotations. To ensure broad applicability across diverse model families, we intentionally adopt a minimal and universal integration design, allowing traffic element signals to be incorporated into existing pipelines in a \textbf{plug-and-play} manner with negligible architectural modification. We evaluate this design across a wide spectrum of modern paradigms, including perception–prediction-planning pipelines, vision-language-action models (VLA), regression-based planners, diffusion-based policies, and trajectory scoring frameworks. Experiments are conducted on multiple widely used benchmarks, including nuScenes, NAVSIM-v1, NAVSIM-v2, and Bench2Drive. Across all paradigms and datasets, our simple integration consistently improves driving performance, demonstrating that traffic element awareness provides a robust and generalizable signal for end-to-end driving systems. Notably, on the challenging NAVSIM-v2 benchmark, our approach significantly boosts the performance of state-of-the-art architectures and data pipelines, establishing a new state of the art.
\end{abstract}
\section{Introduction}
\label{sec:intro}

End-to-end (E2E) autonomous driving has rapidly moved from a research prototype to a deployable paradigm. Today’s systems span a wide spectrum of architectural and training choices: ranging from perception–prediction–planning hybrids~\cite{hu2023planning, jiang2023vad}, to pure regression-based planners~\cite{chitta2022transfuser, jia2025drivetransformer, sun2025sparsedrive}, to diffusion policies~\cite{liao2025diffusiondrive, zou2025diffusiondrivev2, xing2025goalflow} and trajectory scoring frameworks~\cite{li2024hydra, liu2025takead}, and recently to VLM/VLA-style agentic planners~\cite{fu2025orion, tian2024drivevlm, zhou2025autovla, wang2025omnidrive}. Despite this diversity, a surprisingly fundamental factor in human driving is still under-discussed in the E2E literature: traffic elements (TE), such as traffic lights and road signs. From first principles, TE are not \textit{optional context}: they are regulatory signals that directly constrain the feasible action space (e.g., stop/go, forbidden turns, lane-level permissions) and thus should causally shape planning decisions. Yet, mainstream E2E benchmarks and methods overwhelmingly center their perception objectives and representations on dynamic agents (vehicles and pedestrians) and dense geometric cues, leaving the role of TE largely unquantified. The community does not have a systematic answer to a simple question: \textit{how much does traffic-element awareness actually matter for E2E driving performance, across modern paradigms?}

Fig.~\ref{fig:teaser} motivates why this question cannot be dismissed as a corner case. Fig.~\ref{fig:teaser}a illustrates a typical intersection where multiple signals and signs jointly determine what actions are legal and safe. Meanwhile, TE are prevalent rather than long-tailed. As shown in Fig.~\ref{fig:teaser}b, a majority of scenes in widely used datasets contain traffic elements: nuScenes~\cite{caesar2020nuscenes} includes TE in 59.1\% of training scenes (16,626 / 28,130) and 58.3\% of validation scenes (3,508 / 6,019), while NAVSIM-v1 (\texttt{navtest})~\cite{dauner2024navsim} contains TE in 65.1\% of scenes (7,904 / 12,147). This prevalence means that even within academic benchmarks, TE are not a rare special case; and in real-world deployment, E2E systems inevitably interact with traffic lights/signs continuously. The lack of systematic TE studies is therefore not due to irrelevance, but rather due to two practical barriers: (i) missing structured TE annotations in most public datasets~\cite{dauner2024navsim, cao2025navsimv2, caesar2020nuscenes}, and (ii) the heterogeneity of E2E driving systems, where conclusions drawn from a single architecture or training recipe are difficult to generalize.

To close this gap, we present the first systematic investigation of \textbf{plug-and-play traffic element awareness for E2E autonomous driving}. Our goal is deliberately not to propose yet another highly customized planner, but to build a unified research infrastructure and derive generalizable conclusions that hold across model families. Concretely, we (1) augment multiple mainstream driving benchmarks with comprehensive TE annotations; (2) design a minimal universal integration mechanism, implemented as a lightweight auxiliary 3D traffic element supervision (and optional topology conditioning when available~\cite{wang2023openlane}), so that TE signals can be incorporated into existing pipelines in a plug-and-play manner with negligible architectural modifications; and (3) evaluate this integration across a broad set of representative paradigms, including regression-based planners, perception–planning pipelines, diffusion-based policies, trajectory scoring frameworks, and VLM/VLA-style models, on nuScenes~\cite{caesar2020nuscenes}, NAVSIM-v1~\cite{dauner2024navsim}, NAVSIM-v2~\cite{cao2025navsimv2}, and the closed-loop Bench2Drive~\cite{jia2024bench2drive} benchmark.

Our experiments reveal a consistent and practically important conclusion: traffic element awareness yields stable improvements across datasets and paradigms, even under a minimal integration design. As previewed in Fig.~\ref{fig:teaser}c, adding TE supervision reduces open-loop trajectory error on nuScenes, boosts NAVSIM performance for all three paradigm planners, and improves closed-loop driving metrics on Bench2Drive. Interestingly, on the highly challenging NAVSIM-v2 benchmark, TE awareness produces large, consistent gains across representative families (regression / diffusion / scoring), and remains effective even when combined with strong modern data engines~\cite{tian2025simscale} (e.g., pairing a state-of-the-art scoring-based backbone with large-scale simulated co-training data) continues to improve performance and establishes a new state of the art in our evaluation. 

Beyond aggregate gains, we further conduct targeted ablations to explain why TE supervision works and how to integrate it robustly. We find that (i) explicit 3D TE representations outperform 2D-only cues for planning; (ii) selectively modeling TE depth is more effective than feeding global depth maps, suggesting TE distills the most decision-relevant spatial cues; (iii) traffic signs (not only lights) are crucial and often overlooked; and (iv) for sparse TE targets, an independent prediction head, focal loss, and max-preserving pooling are key to avoiding gradient domination by dense background. When lane topology is available, we show that encoding ego-relevant topology into a compact conditioning signal (language-style embeddings), which is more effective than GNN-based graph encoding~\cite{schlichtkrull2018gcn}, complements local TE cues. Together, these results argue that TE awareness is a broadly useful and underutilized signal for E2E driving.
\vspace{-0.5cm}
\section{Related Work}
\label{sec:related work}
\vspace{-0.3cm}

\subsection{End-to-End Autonomous Driving}

End-to-end autonomous driving methods can be broadly categorized into several paradigms according to how driving decisions are produced. Perception–planning methods~\cite{hu2022stp3, hu2023planning, jiang2023vad, chen2024vadv2, sun2025sparsedrive, song2025momad, jia2023thinktwice, jia2023driveadapter, kirby2026drivoR, gao2026uniuncer} construct intermediate scene representations (e.g.,vectorized map) and then perform motion planning based on them. Regression-based methods directly predict trajectories or control commands from sensor observations, including~\cite{chitta2022transfuser ,wu2022trajectory, li2024bevplanner, wozniak2025prix, yuan2024drama, guo2025uad, paradrive, jia2025drivetransformer}. Trajectory-scoring methods~\cite{li2025ztrs, li2025generalized, yao2025drivesuprim, sima2025centaur, li2024hydra} generate multiple candidate trajectories and select the optimal one through a scoring module. Generative planning methods~\cite{liao2025diffusiondrive, zou2025diffusiondrivev2, xing2025goalflow, zheng2024genad, song2025breaking, wang2026meanfuser, liu2025guideflow, liu2025bridgedrive}, including diffusion/flow-based approaches, model trajectory distributions with generative models. Recently, vision–language–action (VLA) methods~\cite{wang2025unified, chen2025drivinggpt, sima2024drivelm, zhou2025autovla, jiang2024senna, hwang2024emma,wang2025omnidrive, li2025drivevla, jiang2025diffvla, chi2025impromptu, ding2024hintad} integrate visual perception, language understanding, and driving actions in a unified framework. In addition, world-model-based approaches~\cite{zheng2025world4drive, li2024ssr, shi2025drivex, li2024law} learn latent environment dynamics for decision making. We evaluate representative methods from these categories to verify the plug-and-play generality of our approach.

\vspace{-0.3cm}
\label{sec: preliminaries}
\begin{wrapfigure}{r}{0.45\textwidth}
    \vspace{-20pt}
    \centering
    \includegraphics[width=1\linewidth]{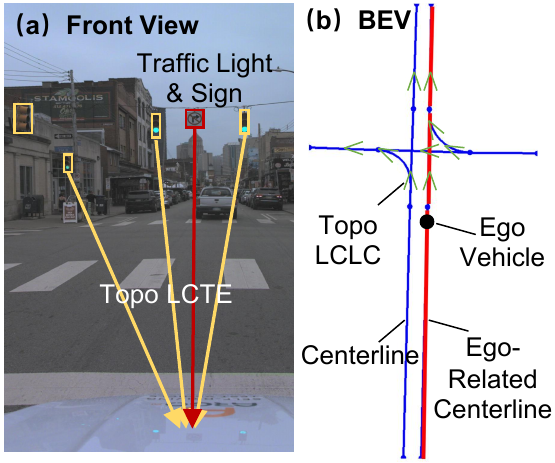}
    \vspace{-1.5em}
    \caption{\textbf{Definitions of driving topologies.} (a) LCTE topology mapping traffic elements (\textcolor{yellow!90!black}{light}/\textcolor{red}{sign}) to the ego-lane in FV. (b) BEV representation of \textcolor{blue}{centerlines}, \textcolor{red}{ego-related centerlines}, and \textcolor{green!60!black}{LCLC topology}.}
    \label{fig:topo definition}
    \vspace{-25pt}
\end{wrapfigure}

\subsection{Traffic-Aware Scene Understanding}
Prior work injects traffic-aware scene understanding into end-to-end planning via structured intermediates such as occupancy~\cite{tong2023scene, hu2023planning, paradrive}, HD map~\cite{hu2022stp3, jiang2023vad, zheng2024genad, wang2026meanfuser, zhang2026umpe, zhang2025delving}, 3D detection-tracking~\cite{sun2025sparsedrive, song2025momad, jia2025drivetransformer}, to enforce safety and rule compliance. VLM-based planners~\cite{zhou2025autovla, wang2025omnidrive, tian2024drivevlm} trained on specific VQA datasets~\cite{park2025nuplanqa, wu2025language, qian2024nuscenes-qa, nie2024reason2drive, sima2024drivelm, guo2026surds} provide higher-level semantic reasoning but at substantial runtime cost. 
Dedicated benchmarks like Openlane-V2~\cite{wang2023openlane, chang2025mapdr} evaluate these scene-understanding components, but most topology-predicted methods~\cite{wu2023topomlp, li2023toponet, lv2025t2sg, fu2024topologic, li2025reusing, li2023lanesegnet, zhang2025chameleon} primarily optimize benchmark scores without validating downstream planning impact.

We provide the first systematic evaluation of how such cues transfer to planning, and propose a lightweight alternative based on explicit traffic elements and lane topology that is more amenable to deployment.

\section{Preliminaries: OpenLane-V2 Scene Representation}
\subsubsection{Components.} Openlane-V2~\cite{wang2023openlane} has four components: 
\textbf{Centerlines:} A set of $M$ 3D lane centerlines (Fig.~\ref{fig:topo definition}b); \textbf{Traffic Elements (TE):} A set of $N$ entities (4 traffic light states and 9 traffic sign categories), parameterized as 2D bounding boxes in the image coordinate system (Fig.~\ref{fig:topo definition}a); \textbf{LCTE Topology:} The Lane-Centerline to Traffic-Element relationship, formulated as a binary adjacency matrix $\mathbf{R}_{\text{LCTE}} \in \{0, 1\}^{M \times N}$, where an entry is $1$ if the $j$-th TE directly governs the $i$-th centerline (Fig.~\ref{fig:topo definition}a); \textbf{LCLC Topology:} The Centerline to Centerline directed connectivity, formulated as an adjacency matrix $\mathbf{R}_{\text{LCLC}} \in \{0, 1\}^{M \times M}$, defines the valid navigable transitions between lanes (Fig.~\ref{fig:topo definition}b).

\subsubsection{Topology Matters.} Scene representation is crucial because it makes the driving constraints explicit. In this intersection (Fig.~\ref{fig:topo definition}), a \textcolor{red}{no-right-turn} sign prohibits right-turn maneuvers, the \textcolor{green!60!black}{LCLC topology} indicates that the ego lane has no lateral connectivity, and the \textcolor{yellow!90!black}{green traffic light} permits forward motion. Together, these cues constrain the feasible action space: the ego vehicle should proceed into the adjacent forward lane at an appropriate speed, rather than turning left or right—a failure mode that vision-only end-to-end planners always exhibit.
\vspace{-0.3cm}

\section{Method}
\vspace{-1cm}
\label{sec: method}
\begin{figure}
    \centering
    \includegraphics[width=0.85\linewidth]{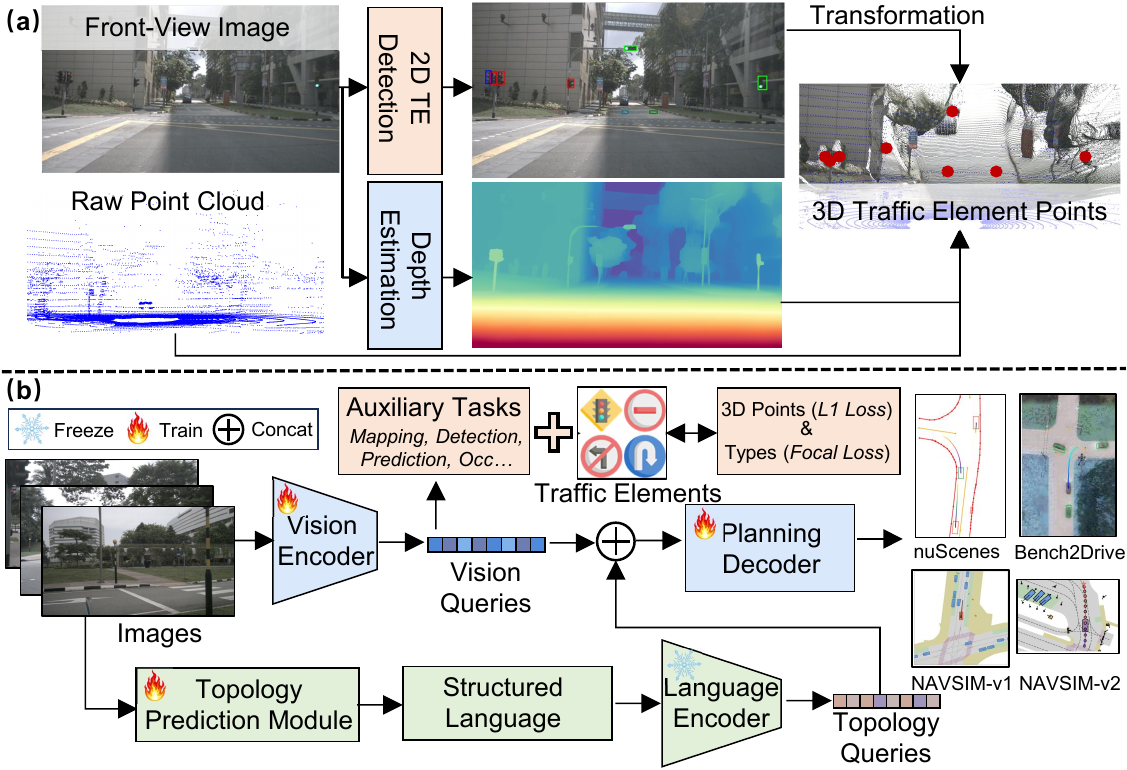}
    \caption{(a) \textbf{Pipeline for 3D traffic element extraction.} We first perform 2D traffic-element detection and monocular depth estimation on the FV image, then
    combine them and LiDAR geometry to localize each traffic element as a 3D center point (\textcolor{red}{$\bullet$}). (b) \textbf{Overview of our method.} Multi-view images are encoded into vision queries and enhanced by an auxiliary 3D traffic element detection task. In parallel, predicted topological adjacency matrices are formatted into structured language and processed by a frozen language encoder. The resulting topology queries are concatenated with the BEV queries to guide the planning decoder in generating the ego-vehicle trajectory.}
    \label{fig:method}
    \vspace{-1.2cm}
\end{figure}

\subsection{3D Dataset Construction} 
To effectively integrate traffic elements into the 3D planning space, we establish a robust pipeline to extract their 3D coordinates $(x, y, z)$ (Fig.~\ref{fig:method}a).
Given a front-view image, we process it through two parallel branches: a 2D TE detection model to obtain bounding boxes, and a monocular depth foundation model~\cite{piccinelli2025unidepthv2} to estimate dense depth maps. Using the camera intrinsic and extrinsic parameters, we project the 2D bounding boxes into the LiDAR coordinate system. The precise 3D center point $(x, y, z)$ of each TE is then jointly determined by aggregating the 2D box constraints, the estimated depth, and the corresponding LiDAR point cloud (details in the Appendix). We adapt this pipeline across different benchmarks based on their available annotations: \textbf{nuScenes~\cite{caesar2020nuscenes} \& Bench2Drive~\cite{jia2024bench2drive}:}  For nuScenes, we directly utilize the 2D TE bounding boxes and topology annotations provided by OpenLane-V2, and apply our extraction pipeline to acquire 3D centers. The Bench2Drive dataset natively provides detailed 3D TE coordinates and topological relationships. \textbf{NAVSIM (v1~\cite{dauner2024navsim} \& v2~\cite{cao2025navsimv2}):} Since these datasets lack TE annotations, we first train a highly accurate YOLO-based~\cite{sapkota2025yolo26} 2D TE detector on OpenLane-V2, adopting progressive training strategies (e.g., resampling and reweighting, pseudo labeling) from \cite{ge2021yolox, wu20231sttopology}. We then deploy this detector alongside our depth-LiDAR fusion pipeline to automatically construct the 3D TE pseudo-labels for NAVSIM.

\subsection{Auxiliary 3D Traffic Element Supervision}
In typical end-to-end autonomous driving frameworks, multi-view images are encoded~\cite{he2016resnet, dosovitskiy2020ViT, oquab2023dinov2, chen2024internvl, esser2021vqgan, wang2026emu3} into latent vision queries (e.g., BEV queries $\mathbf{Q}_\text{bev}$), which are then used for downstream planning (Fig.~\ref{fig:method}b). Building on the original auxiliary tasks of each planner (e.g., detection of 3D vehicle boxes, prediction, tracking, occupancy, etc), we introduce an additional 3D traffic element detection objective. Specifically, for each TE, we supervise its 3D center location with an $L_1$ loss and its category with a focal loss. This TE loss is aggregated into the overall training objective with a weight equivalent to other auxiliary tasks:
\begin{equation}
    \mathcal{L}_{\text{total}} = \mathcal{L}_{\text{plan}} + \sum_{k \in \mathcal{K}} \lambda_k \mathcal{L}_{\text{aux}}^{(k)} + \lambda_{\text{TE}} \big( \mathcal{L}_{\text{L1}}^{\text{loc}} + \mathcal{L}_{\text{focal}}^{\text{cls}} \big),
\end{equation}
where $\mathcal{L}_{\text{plan}}$ is the primary trajectory planning loss, $\mathcal{L}_{\text{aux}}$ includes all original auxiliary losses of the baseline planner, and $\lambda_{\text{TE}}$ is set identically to $\mathcal{L}_{\text{aux}}$.

\subsection{Language-Guided Topology Conditioning}
In parallel with TE-aware BEV learning, a topology prediction module estimates the global adjacency matrices for the entire driving scene, denoted as $\mathbf{R}_{\text{LCLC}}$ and $\mathbf{R}_{\text{LCTE}}$ (Fig.~\ref{fig:method}b). To prevent the planner from being distracted by redundant distant elements, we perform an \textbf{ego-centric topology filtering}. With the ego vehicle's spatial coordinates, we identify its lane ID. We then query $\mathbf{R}_{\text{LCLC}}$ to retrieve the ego vehicle-related centerlines (highlighted in \textcolor{red}{red}, Fig.~\ref{fig:topo definition}b) and query $\mathbf{R}_{\text{LCTE}}$ to isolate the traffic elements directly governing this active lane (Fig.~\ref{fig:topo definition}a). We convert this ego-centric topology graph into a structured language sequence, \textit{e.g.,}  {\small\textit{``There are a `green traffic\_light', a `green traffic\_light', a `green traffic\_light', a `no\_right\_turn road\_sign' ahead controlling the current ego-lane, which connects `1' `straight' centerline.''}} Here, parameters such as the traffic element categories, the number of connected centerlines (`1'), and their directional semantics (`straight', `left') are dynamically instantiated based on the retrieved local topology.

This text is encoded by a pretrained BERT-base~\cite{devlin2019bert} language encoder ($\sim$110M parameters) to obtain topology queries $\mathbf{Q}_\text{topo}$. Finally, we concatenate $\mathbf{Q}_\text{topo}$ with the TE-enhanced vision/BEV queries $\mathbf{Q}_\text{bev}$, and feed the joint queries $\mathbf{Q}_\text{plan} = [\mathbf{Q}_\text{bev};\mathbf{Q}_\text{topo}]$ into the planning decoder. This design injects lane connectivity and rule-control information into the planner in a compact form, enabling trajectory prediction to respect both geometric feasibility and traffic regulations.

We validate our method on \textbf{six} representative end-to-end methods across \textbf{four} benchmarks. Implementation details for integrating our TE supervision and topology conditioning into each specific backbone are provided in the Appendix.
\vspace{-0.7cm}
\section{Experiment}
\label{sec: experiment}
We first specify the experimental protocol (Sec.~\ref{sec: exp setup}). We then report quantitative (Sec.~\ref{sec: exp results}) and qualitative comparisons (Sec.~\ref{sec: exp vis}) to assess the planning performance of our approach across benchmarks. Finally, we conduct targeted ablation studies to isolate the contribution of each component and design choice, and to explain why the proposed mechanism yields consistent gains (Sec.~\ref{sec: ablation study}).

\vspace{-0.4cm}
\subsection{Experimental Setup}
\label{sec: exp setup}

\subsubsection{Datasets and Metrics.} 
We evaluate on \textbf{four datasets}: \textbf{three open-loop} benchmarks (nuScenes~\cite{caesar2020nuscenes}, NAVSIM-v1~\cite{dauner2024navsim}, and NAVSIM-v2~\cite{cao2025navsimv2}), and \textbf{one closed-loop} benchmark (Bench2Drive~\cite{jia2024bench2drive}). NuScenes provides large-scale real-world logs for open-loop ego-trajectory prediction. NAVSIM-v1 (\texttt{navtest}) provides a non-reactive, data-driven evaluation setting with standardized simulation-based scoring. NAVSIM-v2 (\texttt{navhard}) further introduces a two-stage pseudo closed-loop aggregation to better reflect compounding-error effects without requiring full closed-loop simulation. Bench2Drive evaluates end-to-end planners in closed loop in CARLA~\cite{dosovitskiy2017carla} across a large set of short, scenario-isolated routes.

For nuScenes, we report \textbf{L2 trajectory error} and \textbf{Collision Rate} under the standard open-loop protocol. For NAVSIM-v1, we use the official \textbf{PDMS} as the primary score. For NAVSIM-v2, we report the aggregate \textbf{EPDMS} along with the per-stage breakdown (Stage 1/Stage 2) over the core compliance/safety terms. For Bench2Drive, we report \textbf{Driving Score} as the main closed-loop indicators, and additionally include Success Rate, Efficiency, Comfortness, and the open-loop Avg. L2. Detailed metric definitions are provided in the Appendix.
\vspace{-1cm}

\subsubsection{Training Details.} We follow the original training schedules and hyperparameters of each baseline (details in the Appendix). On nuScenes, we train on 8×A100 with batch size 1. On NAVSIM, we train on 4×RTX 3090 with batch size 32. On Bench2Drive, we train on 8×A100 with batch size 1.
\vspace{-0.5cm}

\subsection{Quantitative Results}
\label{sec: exp results}

\subsubsection{NuScenes.}
\begin{table}[t]
    \caption{\textbf{Comparison of methods on the nuScenes dataset}. $\diamond$: Lidar-based methods. \textdagger: The ego status is utilized. FPS is measured on an NVIDIA A100 GPU.}
    \vspace{-0.9cm}
    \label{sample-table}
    \begin{center}
    \footnotesize
    \resizebox{\linewidth}{!}{
    \begin{tabular}{l l cccc cccc c}
        \toprule
        \multirow{2}{*}{Method} & \multirow{2}{*}{Auxiliary Task} & \multicolumn{4}{c}{L2 (m) $\downarrow$} & \multicolumn{4}{c}{Collision Rate (\%) $\downarrow$} & \multirow{2}{*}{FPS$\uparrow$} \\
        \cmidrule(lr){3-6} \cmidrule(lr){7-10}
        & & 1s & 2s & 3s & Avg. & 1s & 2s & 3s & Avg. & \\
        \midrule
        NMP$\diamond$~\cite{zeng2019nmp} & Det \& Motion  & 0.53 & 1.25 & 2.67 & 1.48 & 0.04 & 0.12 & 0.87 & 0.34 & - \\
        FF$\diamond$~\cite{hu2021ff} & FreeSpace  & 0.55 & 1.20 & 2.54 & 1.43 & 0.06 & 0.17 & 1.07 & 0.43 & - \\
       EO$\diamond$  \cite{khurana2022eo} & FreeSpace  & 0.67 & 1.36 & 2.78 & 1.60 & 0.04 & 0.09 & 0.88 & 0.33 & - \\
       \midrule
        
        ST-P3 \cite{hu2022stp3} & Det \& Map \& Depth & 1.33 & 2.11 & 2.90 & 2.11 & 0.23 & 0.62 & 1.27 & 0.71 & 1.6 \\
        UniAD \cite{hu2023planning} & \tiny{Det\&Track\&Map\&Motion\&Occ}  & 0.44 & 0.67 & 0.96 & 0.69 & 0.04 & 0.08 & 0.23 & 0.12 & 1.8 \\
       VAD-Tiny  \cite{jiang2023vad} & Det \& Map \& Motion & 0.46 & 0.76 & 1.12 & 0.78 & 0.21 & 0.35 & 0.58 & 0.38 & 16.8 \\
       BEV-Planner \cite{li2024bevplanner}  & None & {0.28} & {0.42} & {0.68} & {0.46} & {0.04} & {0.37} & {1.07} & {0.49} & - \\
        PARA-Drive \cite{paradrive}  & \tiny{Det\&Track\&Map\&Motion\&Occ} & {0.25} & {0.46} & {0.74} & {0.48} & {0.14} & {0.23} & {0.39} & {0.25} & 5.0 \\
        LAW \cite{li2024law}  & None & {0.26} & {0.57} & {1.01} & {0.61} & {0.14} & {0.21} & {0.54} & {0.30} & 19.5 \\
       GenAD \cite{zheng2024genad} & Det \& Map \& Motion & {0.28} & {0.49} & {0.78} & {0.52} & {0.08} & {0.14} & 0.34 & {0.19} & 6.7 \\
        SparseDrive \cite{sun2025sparsedrive} & \tiny{Det \& Track \& Map \& Motion} & {0.29} & {0.58} & {0.96} & {0.61} & {0.01} & {0.05} & 0.18 & {0.08} & 9.0 \\
        UAD \cite{guo2025uad} & Det & {0.28} & {0.41} & {0.65} & {0.45} & {0.01} & \textbf{0.03} & 0.14 & {0.06} & 7.2 \\
        MomAD~\cite{song2025momad}  & \tiny{Det \& Track \& Map \& Motion} & 0.31 & 0.57 & 0.91 & 0.60 & 0.01 & 0.05 & 0.22 &0.09 & 7.8\\
        \midrule
        VAD-Base \cite{jiang2023vad} & Det \& Map \& Motion & 0.41 & 0.70 & 1.05 & 0.72 & 0.07 & 0.17 & 0.41 & 0.22 & 5.7 \\
        \rowcolor{myblue}
        VAD + TE & Det \& Map \& Motion & 0.36 & 0.61 & 0.92 & 0.63 & 0.09 & 0.14 & 0.28 & \textbf{0.17} & 5.7 \\
        \rowcolor{myblue}
        VAD + Topo & \tiny{Det \& Map \& Motion \& Topo} & 0.34 & 0.59 & 0.92 & 0.62 & 0.05 & 0.15 & 0.29 &  0.16 & 5.4 \\
        \rowcolor{myblue}
        VAD + Ours  & \tiny{Det \& Map \& Motion \& Topo} & 0.34 & 0.56 & 0.90 & \textbf{0.60} & 0.04 & 0.21 & 0.26 & \textbf{0.17} & 5.4 \\
            \rowcolor[HTML]{EEE5F4}\hline
        \multicolumn{11}{c}{\textit{{VLM/VLA-based Method}}} \\
        Qwen-2.5-VL~\cite{Qwen2.5-VL} & Scene Understanding & 0.46 & 1.33 & 2.55 & 1.45 & -- & -- & -- & -- & 0.8\\
        Senna~\cite{jiang2024senna} & Det \& Motion & 0.37 & 0.54 &  0.86 & 0.59 & 0.09 & 0.12 & 0.33 & \textbf{0.18} & 1.6\\
        DriveVLM~\cite{tian2024drivevlm} & Scene Understanding & 0.18 & 0.34 & 0.68 & 0.40 & 0.10 & 0.22 & 0.45 & 0.27 & --\\
        OmniDrive~\cite{wang2025omnidrive} & Scene Understanding & 0.14 & 0.29 & 0.55 & 0.33 & 0.00 &  0.13 & 0.78 & 0.30 &  1.2\\
        EMMA~\cite{hwang2024emma} & Scene Understanding & 0.14 & 0.29 & 0.54 & 0.32 & -- & -- & -- & -- & --\\
        ImpromptuVLA~\cite{chi2025impromptu} & Scene Understanding & 0.13 & 0.27 & 0.53 & \textbf{0.30} & -- & -- & -- & -- & 0.9\\
        \midrule
        Orion\textdagger~\cite{fu2025orion} & Det \& Motion &  0.17 & 0.31 & 0.55 & 0.34 & 0.05 & 0.25 & 0.80 & 0.37 & 0.9\\
        \rowcolor{myblue}
        Orion\textdagger + TE & Det \& Motion &  0.14 & 0.27 & 0.47 & 0.29 & 0.06 & 0.20 & 0.55 & 0.27 & 0.9 \\
        \rowcolor{myblue}
        Orion\textdagger + Topo & Det \& Motion \& Topo &  0.12 & 0.25 & 0.45 & 0.27 & 0.04 & 0.21 & 0.50 & 0.25 & 0.8 \\
        \rowcolor{myblue}
        Orion\textdagger + Ours & Det \& Motion \& Topo &  0.11 & 0.25 & 0.43 & \textbf{0.26} & 0.04 & 0.19 & 0.47 & \textbf{0.23} & 0.8 \\
        
        \bottomrule
    \end{tabular}
    }
    \label{tab:nuscenes}
    \vspace{-1cm}
    \end{center}
\end{table}
We construct 3D traffic elements using the OpenLane-V2 subset annotated on nuScenes, and use the same detection-aligned perception range as prior work~\cite{jiang2023vad}: lateral $[-15,15]$ m and longitudinal $[0,30]$ m. Depth is estimated by UniDepthv2~\cite{piccinelli2025unidepthv2} with frozen weights. We encode topological cues by first predicting a topology adjacency matrix with a lightweight TopoMLP~\cite{wu2023topomlp}, and then mapping it to a language embedding via a frozen BERT encoder~\cite{devlin2019bert}. 

We compare against two representative families of end-to-end planners: the conventional perception-planning baseline VAD~\cite{jiang2023vad} and the VLM-based planner Orion~\cite{fu2025orion}. Tab.~\ref{tab:nuscenes} shows that introducing traffic elements as auxiliary supervision improves both L2 ($\downarrow$0.09; $\downarrow$0.05) and CR ($\downarrow$0.05\%; $\downarrow$0.10\%) over the baselines. We attribute this gain to the fact that the additional TE objective makes the learned BEV queries more \textbf{traffic-aware}: latent vision queries are typically dominated by dense scene geometry and dynamic agents (e.g., via standard occupancy or vehicle detection tasks), which inherently overshadow spatially sparse yet logically critical traffic elements. We force the network to allocate dedicated representational capacity to these vital regulatory signals. Consequently, this prevents traffic rules from being marginalized in the shared feature space.

Furthermore, injecting topological cues yields additional gains, indicating that high-level connectivity/route structure complements local rule signals. The best performance is obtained when both traffic elements and topology are incorporated. Crucially, these scene-understanding signals are predicted by lightweight networks, so the runtime impact is small (FPS drops only 0.3). Compared to VLM-based scene understanding methods~\cite{Qwen2.5-VL, chi2025impromptu, tian2024drivevlm}, this design delivers $>10\times$ higher throughput while still capturing the key cues that drive performance, making the approach substantially more amenable to real-time deployment.
\vspace{-0.5cm}

\subsubsection{NAVSIM.} 
\begin{table}[t]
\caption{\textbf{Camera-only methods on the NAVSIM-v1 benchmark (\texttt{navtest})}.
    }
    \vspace{-0.3cm}
    \centering
    \small
\setlength{\tabcolsep}{2pt}
    \resizebox{\columnwidth}{!}{
    \begin{tabular}{@{}l@{}cc|ccccc|c@{}}
    \toprule
    Method & Backbone & Venue & NC $\uparrow$ & DAC $\uparrow$ & TTC $\uparrow$ & Comf. $\uparrow$ & EP $\uparrow$ & \textbf{PDMS} $\uparrow$\\
    \midrule
    \rowcolor{gray!10}
    PDM‑Closed~\cite{dauner2023pdm}  &  --     & \scriptsize{PMLR'23} & 94.6 & 99.8 & 89.9 & 86.9 & 99.9 & 89.1 \\ 
    \rowcolor{gray!10}
    Human driver~\cite{dauner2024navsim}  &   --       & \scriptsize{NeurIPS'24} & 100 & 100&  100 & 99.9 & 87.5 & \textbf{94.8} \\ 
    \midrule
    Ego‑stat. MLP~\cite{dauner2024navsim} &   --     & \scriptsize{NeurIPS'24} & 93.0 & 77.3 & 83.6 & 100 & 62.8 & 65.6 \\ 
    UniAD~\cite{hu2023planning}   &     ResNet34           & \scriptsize{CVPR'23} & 97.8 & 91.9 & 92.9 & 100  & 78.8 & 83.4 \\ 
    VAD-v2~\cite{chen2024vadv2} & ResNet34         & \scriptsize{ICLR'26} & 98.1 & 94.8 & 94.3 & 100 & 80.6 & 86.2 \\ 
    ReCogDrive~\cite{li2025recogdrive}  &    InternViT   & \scriptsize{ICLR'26} & 97.9 & 97.3 & 94.9 & 100  & 87.3 & 90.8 \\ 
    Hydra-MDP~\cite{li2024hydra}   & ResNet34         & \scriptsize{arXiv'24} & 98.3 & 96.0 & 94.6 & 100  & 78.7 & 86.5 \\ 
    Centaur~\cite{sima2025centaur}      &   ResNet34        & \scriptsize{arXiv'25} & 99.5 & 98.9 & 98.0 & 100  & 85.9 & \textbf{92.6} \\ 

    DriveSuprim~\cite{yao2025drivesuprim}  & ResNet34  & \scriptsize{AAAI'26} &97.8 & 97.3 & 93.6 & 100 & 86.7 & 89.9 \\
    \midrule
         \rowcolor[HTML]{EEE5F4}\hline
     \multicolumn{9}{c}{\textit{{Regression-based Planner}}} \\
    LTF~\cite{chitta2022transfuser}& ResNet34          & \scriptsize{TPAMI'22} & 97.8 & 92.8 & 93.3 & 100  & 78.9 & 84.1 \\ 
    \rowcolor{myblue}
    LTF + Ours& ResNet34             & \scriptsize{ECCV'26} & 97.8 & 93.9 & 93.8 & 100  & 79.8 & 85.2 \textcolor[rgb]{0,0.3,0.6}{{\tiny +1.1}}\\ 
    \rowcolor{myblue}
    LTF \tiny{(+SimScale~\cite{tian2025simscale}) \hspace{+1mm}} & ResNet34             & \scriptsize{ECCV'26} & 98.3 & 95.6 & 94.6 & 100  & 81.3 & 87.3 \textcolor[rgb]{0,0.3,0.6}{{\tiny +3.2}}\\
    \rowcolor{myblue}
    LTF {\tiny(+SimScale)} + Ours \hspace{+1mm} & ResNet34     & \scriptsize{ECCV'26} & 98.2 & 95.9 & 94.5 & 100  & 82.0 & \textbf{87.6} \textcolor[rgb]{0,0.3,0.6}{{\tiny +\textbf{3.5}}}\\ 
    \midrule
         \rowcolor[HTML]{EEE5F4}\hline
     \multicolumn{9}{c}{\textit{{Diffusion-based Planner}}}\\
    DiffusionDrive~\cite{liao2025diffusiondrive} & ResNet34 & \scriptsize{CVPR'25} & 97.9 & 94.6 & 93.6 & 100  & 80.7 & 86.0 \\ 
    \rowcolor{myblue}
    DiffusionDrive + Ours & ResNet34 & \scriptsize{ECCV'26} & 98.1 & 96.0 & 94.2 & 100  & 82.3 & 87.7 \textcolor[rgb]{0,0.3,0.6}{{\tiny +1.7}}\\ 
    \rowcolor{myblue}
    DiffusionDrive \tiny{(+SimScale~\cite{tian2025simscale}) \hspace{+1mm}} & ResNet34             & \scriptsize{ECCV'26} & 98.5 & 97.0 & 94.7 & 100  & 83.1 & 88.9 \textcolor[rgb]{0,0.3,0.6}{{\tiny +2.9}}\\
    \rowcolor{myblue}
    DiffusionDrive {\tiny(+SimScale)} + Ours \hspace{+1mm} & ResNet34  & \scriptsize{ECCV'26} & 98.6 & 97.2 & 94.7 & 100  & 83.5 & \textbf{89.1} \textcolor[rgb]{0,0.3,0.6}{{\tiny +\textbf{3.1}}}\\ 
    \midrule
        \rowcolor[HTML]{EEE5F4}\hline
     \multicolumn{9}{c}{\textit{{Scoring-based Planner}}}\\
   DrivoR~\cite{kirby2026drivoR}   &ViT-S&\scriptsize{CVPR'26}& 98.9 & 98.3 & 96.2 & 100 & 89.1 &  93.1 \\

   \rowcolor{myblue}
   DrivoR + Ours   & ViT-S & \scriptsize{ECCV'26} & 99.0 & 98.7 & 96.9 & 100 & 90.8 &  94.4 \textcolor[rgb]{0,0.3,0.6}{{\tiny +1.3}}\\
    \rowcolor{myblue}
    \rowcolor{myblue}  
    DrivoR \tiny{(+SimScale~\cite{tian2025simscale}) \hspace{+1mm}}   & ViT-S & \scriptsize{ECCV'26} & 99.1 & 99.2 & 96.9 & 100 & 91.6 & 94.6 \textcolor[rgb]{0,0.3,0.6}{{\tiny +1.5}}     \\
    \rowcolor{myblue}
    DrivoR {\tiny(+SimScale)} + Ours \hspace{+1mm}   &ViT-S & \scriptsize{ECCV'26} & 99.7 & 99.7 & 98.1 & 100 & 92.2 & \textbf{95.1}  \textcolor[rgb]{0,0.3,0.6}{{\tiny +\textbf{2.0}}}   \\
    \bottomrule

    \end{tabular}
    }
    \label{tab:benchmark_navsim_v1}
    \vspace{-0.5cm}
\end{table}
Since NAVSIM does not provide traffic element annotations, we generate pseudo-labels by running our OpenLaneV2-trained TE detector on the full front-view image, and supervise a BEV TE heatmap head with focal loss ($\alpha{=}2$, $\beta{=}4$). We adopt the NAVSIM perception range lateral $[-32,32]$ m and longitudinal $[0,32]$ m, and estimate depth using UniDepthv2 as in nuScenes. Due to the different camera setup in NAVSIM compared to nuScenes, we do not train a topology predictor in this setting; instead, we focus on TE only and feed the predicted TE representation to the decoder to guide planning.

On NAVSIM-v1, we evaluate \textbf{three} representative end-to-end camera-only planning paradigms: the regression-based planner LTF~\cite{chitta2022transfuser}, the diffusion-based planner DiffusionDrive~\cite{liao2025diffusiondrive}, and scoring-based planners DrivoR~\cite{kirby2026drivoR}. Tab.~\ref{tab:benchmark_navsim_v1} shows that adding traffic elements as explicit supervision and conditioning the planner on the predicted TE representation yield consistent gains across all paradigms. The improvements are primarily driven by higher NC (No Collision) and DAC (Drivable Area Compliance)—two key safety/compliance factors that carry substantial weight in the overall PDMS—highlighting the importance of traffic-element understanding for reliable planning. To test \textbf{data scalability}, we further leverage SimScale~\cite{tian2025simscale} to generate simulated data and co-train using traffic-element cues extracted from both simulated and real logs. This co-training strategy leads to a further increase in PDMS, indicating that our supervision signal transfers effectively across domains and benefits from larger-scale training. Notably, with DrivoR, our approach surpasses both the rule-based method~\cite{dauner2023pdm} and the human-driver score~\cite{dauner2024navsim}, suggesting that \textbf{stronger backbones} and \textbf{more training data} can unlock additional gains. 

We further validate these findings on NAVSIM-v2 with same representative planners, and observe the same consistent trend (Tab.~\ref{tab:benchmark_navsim_v2}). Across methods, incorporating traffic elements improves the overall score by roughly \textbf{+10 EPDMS}. The gains are mainly attributed to stronger rule compliance and safety, reflected by improvements in the core terms NC, DAC, DDC, and TLC—suggesting that TE supervision helps the model internalize traffic rules. We also observe that adding SimScale data can noticeably reduce EC (Extended Comfort), likely because some counterfactual simulated scenarios introduce distribution shifts that affect ride quality. Importantly, combining SimScale with our TE-centric training mitigates this effect and brings EC back to a higher level. 
Overall, these results demonstrate that our method improves safety-critical behavior and exhibits \textbf{strong scalability} with \textbf{model} and \textbf{data} scale.

\begin{table*}[t]
 \caption{Comparison to existing methods on the \textbf{NAVSIM-v2} \textbf{\texttt{navhard-two-stage}} benchmark using EPDMS. \textdagger: Metrics reported prior to the official benchmark bug fix.
    }
    \vspace{-0.3cm}
    \centering
    \small
    \setlength{\tabcolsep}{1.8pt}
    \resizebox{\linewidth}{!}{%
    \begin{tabular}{@{}l@{}|ccccccccc|ccccccccc|c@{}}
    \toprule
            & \multicolumn{9}{c}{Stage 1} & \multicolumn{9}{c}{Stage 2} & \\
    Method  &NC & DAC & DDC & TLC & EP & TTC & LK & HC & EC & NC & DAC & DDC & TLC & EP & TTC & LK & HC & EC & \textbf{EPDMS} $\uparrow$\\
    \midrule
    ZTRS\textdagger~\cite{li2025ztrs} & 98.9 & 97.6 & 100 & 100 & 66.7& 98.9& 96.2& 96.7& 44.0 & 91.1 & 90.4& 95.8& 99.0 &63.6& 89.8 &60.4& 97.6& 66.1& 45.5 \\
    DiffVLA\textdagger~\cite{jiang2025diffvla} & 95.7 & 99.2 & 100 & 100 &  85.9 & 96.4 & 97.1 & 95 & 84.2 & 81.2 & 88.8 & 94.6 & 99.0 & 86.0 & 76.4 & 59.8 & 98.6 & 80.4 & 45.0 \\
    GTRS-Dense\textdagger~\cite{li2025generalized} &
     98.9  & 94.9  & 99.1 & 100 & 76.1 & 98.4 & 93.8  & 94.9 & 37.8 &
    89.9 & 90.5 &94.1 & 99.3 & 77.6  &  88.5 & 56.0 & 92.0  & 30.2 & 41.9 \\
    GuideFlow\textdagger~\cite{liu2025guideflow} & 97.8 & 97.1 & 100 & 100 & 81.4 & 98.5 & 91.4 & 92.8 & 34.2 & 87.3 & 92.3 & 98.0 & 96.9 & 75.8 & 85.5 & 59.3 & 95.4 & 53.5 & \textbf{46.7}\\ \midrule
    \rowcolor[HTML]{EEE5F4}\hline
     \multicolumn{20}{c}{\textit{{Regression-based Planner}}} \\
    LTF~\cite{chitta2022transfuser} & 96.2 & 79.5 & 99.1 & 99.5 & 84.1 & 95.1 & 94.2 & 97.5 & 79.1 & 77.7 & 70.2 & 84.2 & 98.0 & 85.1 & 75.6 & 45.4 & 95.7 &  75.9 & 25.1\\ 
    \rowcolor{myblue}
    LTF + Ours & 96.4 & 79.8 & 98.9 & 99.6 & 84.2 & 95.8 & 93.8 & 97.5 & 78.2 & 80.9 & 71.6 & 84.8 & 98.9 & 85.2 & 77.7 & 47.0 & 96.1 & 76.3 & 28.9 \textcolor[rgb]{0,0.3,0.6}{{\tiny $\uparrow$15\%}}\\
    \rowcolor{myblue}
    LTF{\tiny(+SimScale\cite{tian2025simscale})} & 96.1 & 85.3 & 99.4 & 99.3 & 84.7 & 94.7 & 93.5 & 97.5 & 77.3 & 85.5 & 66.9 & 91.5 & 99.1 & 93.0 & 81.1 & 58.2 & 95.1 & 42.9 & 33.6
 \textcolor[rgb]{0,0.3,0.6}{{\tiny $\uparrow$33\%}} \\
    \rowcolor{myblue}
    LTF{\tiny(+SimScale)}+Ours& 97.0 & 85.3 & 99.7 & 99.6 & 84.2 & 96.2 & 96.0 & 97.6 & 77.3 & 88.1 & 71.8 & 93.1 & 99.0 & 86.7 & 83.0 & 54.0 & 95.2 & 55.4 & \textbf{36.9} \textcolor[rgb]{0,0.3,0.6}{{\tiny $\uparrow$\textbf{47}\%}}\\
    \rowcolor[HTML]{EEE5F4}\hline
     \multicolumn{20}{c}{\textit{{Diffusion-based Planner}\quad \quad}} \\
    DiffusionDrive~\cite{liao2025diffusiondrive} & 96.7 & 86.7 & 98.7 & 99.3 & 84.3 & 94.9 & 95.3 & 97.6 & 77.8 & 78.9 & 72.4 & 84.2 & 98.2 & 87.1 & 74.9 & 47.3 & 96.2 & 71.0 & 29.4\\
    \rowcolor{myblue}
    DiffusionDrive + Ours & 97.1 & 86.9 & 98.8 & 99.6 & 84.2 & 95.1 & 95.6 & 97.6 & 79.5 & 80.2 & 74.0 & 85.0 & 98.1 & 86.3 & 77.2 & 48.4 & 96.6 & 74.4 & 32.7 \textcolor[rgb]{0,0.3,0.6}{{\tiny $\uparrow$11\%}}\\
    \rowcolor{myblue}
    DiffusionDrive{\tiny(+SimScale\cite{tian2025simscale})} & 97.2 & 88.0 & 99.2 & 99.3 & 82.8 & 96.7 & 98 &97.5 & 58.2 & 86.7 & 72.1 & 93.0 & 98.8 & 92.1 & 80.6 & 61.1 & 95.3 & 33.1 & 35.8 \textcolor[rgb]{0,0.3,0.6}{{\tiny $\uparrow$12\%}}\\
    \rowcolor{myblue}
    DiffusionDrive{\tiny(+SimScale)}+Ours & 97.1 & 88.4 & 99.3 & 99.3 & 84.1 & 95.6 & 98.2 & 97.6 & 72.9 & 83.7 & 76.3 & 92.5 & 98.9 & 90.3 & 78.9 & 57.7 & 94.4 & 56.1 & \textbf{37.9} \textcolor[rgb]{0,0.3,0.6}{{\tiny $\uparrow$\textbf{29}\%}}\\
       \rowcolor[HTML]{EEE5F4}\hline
     \multicolumn{20}{c}{\textit{{Scoring-based Planner}\quad \quad}} \\
    DrivoR~\cite{kirby2026drivoR} & 98.8 &95.1 &98.9 &100 &72.6 &98.7 &94.0 &97.6 &73.3 &90.2 &88.4 &91.9& 98.6 &70.0 &88.0 &50.1 &98.5&  76.2 & 48.3\\
    \rowcolor{myblue}
    DrivoR + Ours &
     98.9  & 94.9  & 99.1 & 100 & 75.1 & 98.9 & 93.9  & 97.5 & 74.4 &
    91.9 & 91.4 &96.1 & 99.3 & 74.6  &  87.5 & 56.0 & 97.0  & 78.5 & 51.8 \textcolor[rgb]{0,0.3,0.6}{{\tiny $\uparrow$7.2\%}}\\
    \rowcolor{myblue}
    DrivoR{\tiny(+SimScale\cite{tian2025simscale})} & 99.1 &98.2 &99.3 &99.8 &75.4 &98.7 &94.9 &97.6 &70.2 &92.3 &91.6 &97.3 &99.1 &75.7 &90.6 &56.1 &98.4 &44.7 & 54.6 \textcolor[rgb]{0,0.3,0.6}{{\tiny $\uparrow$13\%}}\\
    \rowcolor{myblue}
    DrivoR{\tiny(+SimScale)}+Ours &
     99.5  & 98.2 & 99.3 & 100 & 76.1 & 99.0 & 94.1  & 97.9 & 72.6 &
    93.5 & 91.8 &97.0 & 99.3 & 76.6  &  90.9 & 56.0 & 98.0  & 55.6 & \textbf{57.9} \textcolor[rgb]{0,0.3,0.6}{{\tiny $\uparrow$\textbf{20}\%}}\\

    \bottomrule
    \end{tabular}
    }
    \vspace{-0.3cm}
    \label{tab:benchmark_navsim_v2}
\end{table*}

\vspace{-0.6cm}
\subsubsection{Bench2Drive.}
\begin{table}[tb!]
\centering
\small
\caption{\textbf{Closed-Loop Planning Performance in Bench2Drive}. Avg. L2 is averaged over the predictions in 2 seconds. $\ast$: denotes expert feature distillation. All latency is measured by the averaged inference step-time on CARLA~\cite{dosovitskiy2017carla} evaluation in A6000.}
\vspace{-2mm}
\label{tab:b2d}
\resizebox{\linewidth}{!}{
\begin{tabular}{l|c|cccc|c}
\toprule
\multirow{2}{*}{\textbf{Method}} & \textbf{Open-loop Metric} & \multicolumn{4}{c|}{\textbf{Closed-loop Metric}} & \multirow{2}{*}{\textbf{Latency}} \\ \cmidrule{2-6} 
                                 & Avg. L2 $\downarrow$      &  Driving Score $\uparrow$  & Success Rate(\%) $\uparrow$ & Efficiency $\uparrow$ & Comfortness $\uparrow$ & \\ \midrule 

TCP$\ast$~\cite{wu2022trajectory}    & 1.70                & 40.70     & 15.00  & 54.26 & 47.80 & \textbf{86ms} \\ 
TCP-ctrl$\ast$ & -                 &  30.47    & 7.27  & 55.97 & \textbf{51.51} & \textbf{86ms} \\
TCP-traj$\ast$ & 1.70                &   59.90     & 30.00 & 76.54 & 18.08  & \textbf{86ms}  \\
TCP-traj w/o distillation                              & 1.96                &   49.30     & 20.45  & \textbf{78.78} & 22.96 & \textbf{86ms} \\
ThinkTwice$\ast$~\cite{jia2023thinktwice}                              & \textbf{0.95}               &   62.44     & 31.23  & 69.33 & 16.22 & 698ms  \\
DriveAdapter$\ast$~\cite{jia2023driveadapter}                             & 1.01                &  \textbf{64.22}     & \textbf{33.08} & 70.22 & 16.01 & 958ms   \\ \midrule
AD-MLP~\cite{zhai2023ADMLP}                            & 3.64              &  18.05     &  0.00  & 48.45 &   22.63 & \textbf{4.7ms}  \\ 
UniAD-Tiny~\cite{hu2023planning}                             &  0.80       &  40.73    & 13.18 & 123.92 & \textbf{47.04} & 400.3ms \\
UniAD-Base~\cite{hu2023planning}                 &  0.73          &  45.81     & 16.36 & \textbf{129.21} & 43.58 & 692.6ms  \\ 
\midrule
VAD~\cite{jiang2023vad}   &    0.91  &  42.3  & 15.00 & \textbf{157.94} & 46.01 & 278.3ms \\
\rowcolor{myblue}
VAD + Ours   &    \textbf{0.75} \textcolor[rgb]{0,0.3,0.6}{{\tiny -\textbf{0.16}}}  &  \textbf{56.4} \textcolor[rgb]{0,0.3,0.6}{{\tiny +\textbf{14.1}}}  & \textbf{21.30} & 125.64 & \textbf{48.02} & 282.5ms \\
DriveTransformer-Large~\cite{jia2025drivetransformer} & 0.62 &  63.46 &  35.01 & \textbf{100.64} & 20.78 & 211.7ms \\
\rowcolor{myblue}
DriveTransformer-Large + Ours & \textbf{0.57} \textcolor[rgb]{0,0.3,0.6}{{\tiny -\textbf{0.05}}} &  \textbf{68.29} \textcolor[rgb]{0,0.3,0.6}{{\tiny +\textbf{4.83}}} &  \textbf{39.61} & 82.45 & \textbf{23.58} & 216.5ms \\
\bottomrule 
\end{tabular}}
\label{tab: bench2drive}
\vspace{-0.5cm}
\end{table}
Bench2Drive provides instance-level traffic elements with 3D locations, as well as an HD-map lane topology that encodes lane-level topology and connectivity. We therefore supervise our traffic-element head directly using the provided 3D TE coordinates and encode map topology in the same manner as in nuScenes by mapping topology cues to a compact embedding that conditions the planner. Traffic elements used by the planner are obtained from our model predictions at inference time, rather than from reading annotation states. We run closed-loop planning using two baselines (VAD~\cite{jiang2023vad} and DriveTransformer~\cite{jia2025drivetransformer}) at 2 Hz. Tab.~\ref{tab: bench2drive} shows consistent improvements for both methods in L2 and the main closed-loop metrics (notably Driving Score), with a small change in latency. Interestingly, Efficiency decreases after adding our TE supervision and topology. We interpret this as a correction of over-aggressive behaviors encouraged by efficiency-dominated objectives: baseline planners can achieve high efficiency while under-modeling rule-critical traffic elements, whereas TE awareness biases the policy toward safer and more compliant maneuvering. 
\vspace{-0.3cm}

\subsection{Qualitative Results}
\label{sec: exp vis}

\begin{figure}
    \centering
    \includegraphics[width=1.0\linewidth]{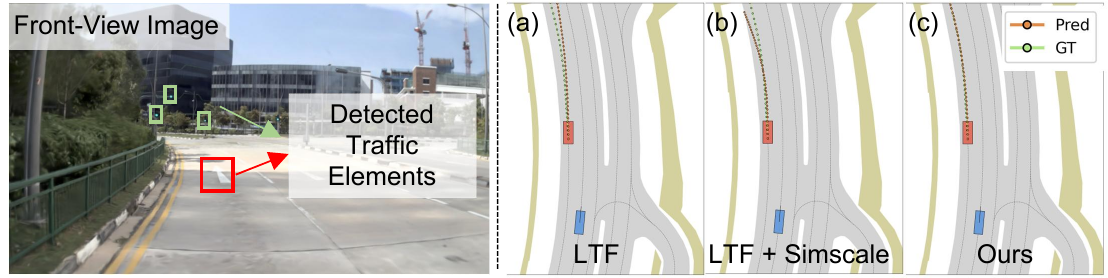}
    \caption{Qualitative comparison on \textbf{NAVSIM-v2}. \textbf{Left:} front-view image with detected traffic elements. \textbf{Right:} predicted trajectories of LTF, LTF+SimScale, and Ours.}
    \label{fig:navsimv2 vis}
\end{figure}

Fig.~\ref{fig:navsimv2 vis} shows a challenging NAVSIM-v2 intersection. Without traffic-element awareness, \textbf{LTF} and \textbf{LTF+SimScale} exhibit noticeable lateral drift (Fig.~\ref{fig:navsimv2 vis}a-b). In contrast, \textbf{Ours} detects the green traffic lights and the straight-ahead sign for the ego lane, and thus continues forward while staying lane-consistent (Fig.~\ref{fig:navsimv2 vis}c). This example highlights that traffic elements provide decision-critical cues that stabilize planning in complex junctions (see Appendix for more visualization).
\vspace{-0.5cm}

\vspace{-0.8cm}

\subsection{Ablation Study and Analysis}
\label{sec: ablation study}

\subsubsection{Ablation on Traffic-Element Representations for Planning.}
\begin{table*}[t]
 \caption{Ablation study of \textbf{traffic element (TE) representations} on the \textbf{NAVSIM-v2} using \textbf{LTF} (Latent Transfuser). \textbf{TE(2D)} denotes 2D traffic elements in the front-view (\textbf{FV}) image; \textbf{depth(FV)} uses the full FV depth; \textbf{LiDAR} uses the NAVSIM-v2 point-cloud input; \textbf{TL} denotes traffic lights (green/yellow/red light) only.
    }
    \vspace{-0.3cm}
    \centering
    \small
    \setlength{\tabcolsep}{1.8pt}
    \resizebox{\linewidth}{!}{%
    \begin{tabular}{@{}l@{}|ccccccccc|ccccccccc|c@{}}
    \toprule
            & \multicolumn{9}{c}{Stage 1} & \multicolumn{9}{c}{Stage 2} & \\
    Method  &NC & DAC & DDC & TLC & EP & TTC & LK & HC & EC & NC & DAC & DDC & TLC & EP & TTC & LK & HC & EC & \textbf{EPDMS} $\uparrow$\\
    \midrule
    LTF~\cite{chitta2022transfuser} & 96.2 & 79.5 & 99.1 & 99.5 & 84.1 & 95.1 & 94.2 & 97.5 & 79.1 & 77.7 & 70.2 & 84.2 & 98.0 & 85.1 & 75.6 & 45.4 & 95.7 &  75.9 & 25.1\\ 
    LTF+TE(2D) & 96.4 & 82.0 & 99.2 & 99.6 & 84.0 & 95.8 & 93.1 & 97.8 & 79.1& 81.6 & 68.4& 85.5&98.6& 85.4 & 78.4 & 46.9 & 96.2 & 76.5 & 28.1\textcolor[rgb]{0,0.3,0.6}{{\tiny +3.0}}\\
    
    LTF+depth(FV) & 96.9& 81.8& 99.0& 99.6& 84.2&95.8& 92.7& 97.8& 78.2& 81.3& 70.1& 84.3& 98.6& 85.7& 78.4& 45.9& 96.3& 76.5& 27.7\textcolor[rgb]{0,0.3,0.6}{{\tiny +2.6}}\\
    LTF+TE(Lidar) & 97.2 & 80.9 & 99.0 & 99.5 & 83.9 & 96.0 & 92.9 & 97.8 & 79.5 & 81.1 & 70.2 & 85.4 & 98.6 & 85.0 & 79.1 & 45.9  & 96.0 & 77.2  & 27.9\textcolor[rgb]{0,0.3,0.6}{{\tiny +2.8}} \\
    LTF+TE~\cite{lin2025depthanything3} & 96.7 & 81.8 & 99.1 & 99.3 & 84.1 & 95.6 & 92.4 & 97.6 & 78.2 & 81.2 & 69.4 & 84.9 & 98.6 & 85.6 & 78.3 & 46.6 & 96.2 & 76.0 & 27.8\textcolor[rgb]{0,0.3,0.6}{{\tiny +2.7}}\\
    LTF+TL~\cite{piccinelli2025unidepthv2} & 97.1 & 80.7 & 99.1 & 99.3 & 84.1 & 95.3 & 93.6 & 97.8 & 78.2 & 80.4 & 69.5 & 84.8 & 98.6 & 85.7 & 77.6 & 45.0 & 96.3 & 76.4 & 26.7\textcolor[rgb]{0,0.3,0.6}{{\tiny +1.6}}\\
    \rowcolor{myblue}
    LTF+Ours & 96.4 & 79.8 & 98.9 & 99.6 & 84.2 & 95.8 & 93.8 & 97.5 & 78.2 & 80.9 & 71.6 & 84.8 & 98.85 & 85.2 & 77.7 & 47.0 & 96.1 & 76.3 & \textbf{28.9}\textcolor[rgb]{0,0.3,0.6}{{\tiny \textbf{+3.8}}}\\

    \bottomrule
    \end{tabular}
    }
    \vspace{-7pt}
    \label{tab:dataset ablation}
\end{table*}

To dissect the efficacy of our proposed 3D traffic element (TE) representation, we conduct ablation studies on the NAVSIM-v2~\cite{cao2025navsimv2} using the LTF baseline~\cite{chitta2022transfuser} (Tab.~\ref{tab:dataset ablation}). 

(1) \textbf{Explicit 3D vs. 2D Representations.} While 2D traffic elements (LTF+TE(2D)) provide basic semantic context, they inherently lack the depth and spatial precision crucial for motion planning. In contrast, our method explicitly constructs these elements within a 3D ego-centric space (28.1 vs. 28.9).

(2) \textbf{Targeted Depth vs. Global Depth.} Incorporating spatial auxiliary tasks generally improves the LTF baseline (25.1 EPDMS). However, simply feeding the full front-view depth to the model (LTF+depth(FV), \textbf{+2.6}) yields sub-optimal gains compared to our explicit 3D TE formulation (LTF+Ours, \textbf{+3.8}). This indicates that unconstrained global depth may introduce redundant geometric noise, whereas selectively isolating and estimating the depth of specific traffic elements distills the most actionable spatial cues for the planner. 

(3) \textbf{Vision-based Depth Estimation vs. LiDAR Clustering.} We compare our method against a geometry-heuristic approach (LTF+TE(Lidar)), which directly projects 2D bounding boxes into the 3D LiDAR coordinate system and extracts element centers via point clustering. Although helpful (+\textbf{2.8}), it still underperforms our approach (27.9 vs. 28.9). This validates our hypothesis that LiDAR returns on certain traffic elements—particularly flat road surface signs—are often too sparse or severely affected by dynamic occlusions.

(4) \textbf{The Impact of Depth Quality.} The accuracy of the underlying depth prior bottlenecks planning performance. When we replace our autonomous-driving-aligned depth estimator (UniDepthV2~\cite{piccinelli2025unidepthv2}) with a general-purpose foundation model (Depth Anything 3~\cite{lin2025depthanything3}), the EPDMS drops to 27.8. 

(5) \textbf{The Crucial Role of Traffic Signs.} Modeling traffic lights only (LTF+TL) is insufficient. It lags behind our full TE setting (26.7 vs. 28.9), highlighting the often-overlooked importance of traffic signs in shaping planning decisions.
\vspace{-0.4cm}

\subsubsection{Design Choices for Traffic Element Integration.}
\newcommand{\cmark}{\textcolor{blue!80!black}{\ding{51}}}
\newcommand{\xmark}{\textcolor{red!70!black}{\ding{55}}}

\begin{table}[t]
\centering
\small
\caption{Ablation study of \textbf{traffic element integration} on the \textbf{NAVSIM-v2} using \textbf{LTF} (Latent Transfuser). We systematically investigate the design choices for the prediction head, loss function, spatial pooling strategy, and planning interaction method.}
\renewcommand{\arraystretch}{0.8}
\resizebox{0.8\linewidth}{!}{
\begin{tabular}{
                l | c c | c c | c c|  c c| c }
\toprule

\multirow{2}{*}{ID} &  \multicolumn{2}{c|}{TE Head} & \multicolumn{2}{c|}{Loss} & \multicolumn{2}{c|}{Pooling} & \multicolumn{2}{c|}{Interaction}  & \multirow{2}{*}{\textbf{EPDMS} $\uparrow$} \\

\cmidrule(lr){2-3}\cmidrule(lr){4-5} \cmidrule(lr){6-7} \cmidrule(lr){8-9}
 &   w/ BEV & Indep.  & Cross-Entropy  & Focal & Average & Max & C.A. & Concat& \\
\midrule
0 &  &  &   &  &   &     & &   &  25.1\\
1 & \cmark &  & \cmark  &  & \cmark  &     & &   &  24.3\textcolor[rgb]{0,0.3,0.6}{{\tiny -0.8}}\\
2 &  &  \cmark   &  &  \cmark   & \cmark &  &   & &26.1\textcolor[rgb]{0,0.3,0.6}{{\tiny +1.0}} \\
3 &  & \cmark &  \cmark &  & \cmark  &   &   &     & 25.4\textcolor[rgb]{0,0.3,0.6}{{\tiny +0.3}}\\
4 & \cmark &  &  \cmark  &  & \cmark  &   & \cmark  &     & 24.9\textcolor[rgb]{0,0.3,0.6}{{\tiny -0.2}}\\
5 &  &   \cmark  &  & \cmark &   & \cmark & \cmark &  & 27.5\textcolor[rgb]{0,0.3,0.6}{{\tiny +2.4}} \\
\rowcolor{myblue}
6 &  &  \cmark   &  &  \cmark   & & \cmark  &   & \cmark &\textbf{28.9}\textcolor[rgb]{0,0.3,0.6}{{\tiny \textbf{+3.8}}} \\
\bottomrule
\end{tabular}
}
\label{Tab:component ablation}
\end{table}
To validate the effectiveness of our proposed modules, we systematically ablate the design choices of the traffic element conditioning mechanism. Results are summarized in Tab.~\ref{Tab:component ablation}. 

(1) \textbf{Prediction Head and Loss Function.} Comparing ID 1 and 2, decoupling TE prediction into an independent head yields a significant performance boost (+\textbf{1.8} EPDMS). Traffic elements are extremely sparse; treating them as an additional BEV semantic class with standard cross-entropy (ID 1) even degrades performance (-0.8), as the TE gradients are dominated by the dense background. Moreover, replacing CE with focal loss under the independent head (ID 2 vs. ID 3) brings a further +\textbf{0.7} EPDMS, consistent with focal loss being better suited for the severe foreground–background imbalance of tiny TE targets.

(2) \textbf{Pooling and Interaction Mechanism.} Explicitly routing TE constraints to the planner via Cross-Attention (C.A.) improves the baseline (4 vs. 1, +\textbf{0.6}), proving the necessity of interactive planning. However, since critical signals occupy very few pixels, Average Pooling severely dilutes their activations during spatial downsampling. Upgrading to Adaptive Max Pooling (ID 5) guarantees these peak activations are perfectly preserved in the low-resolution grid. Our optimal configuration (ID 6) replaces C.A. by concatenating the max-pooled TE features directly into the BEV memory. This creates a strictly aligned "Dual Spatial Stream" that outperforms C.A. (+\textbf{1.4}). We attribute this to a more direct and spatially consistent conditioning signal, which helps the planner associate traffic rules with the correct ego-relevant lane context.

\begin{figure}[t] 
\centering 
\includegraphics[width=0.68\linewidth]{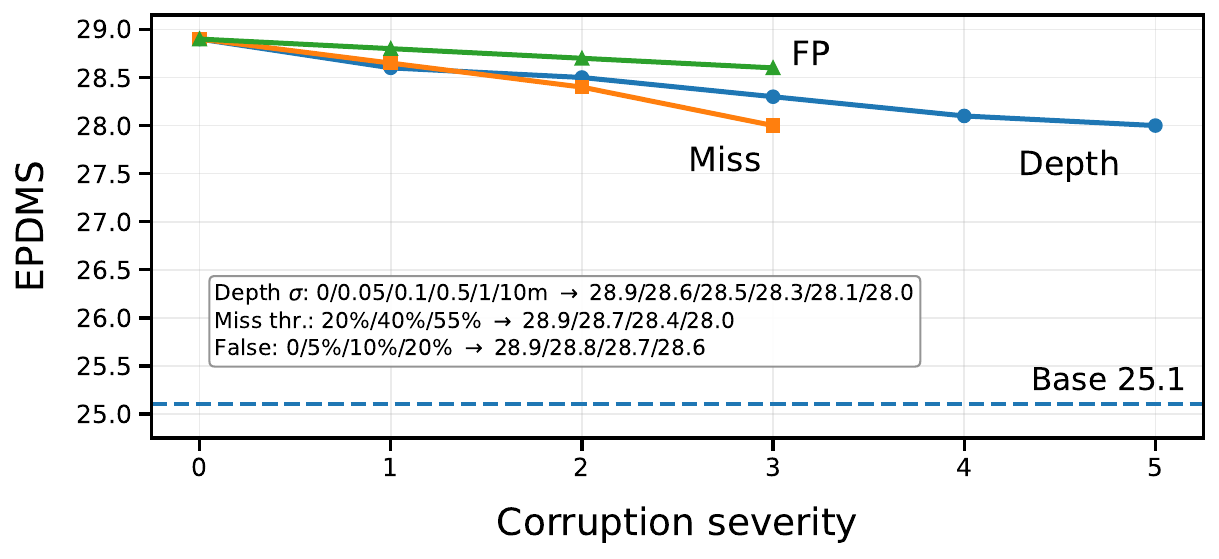} \caption{Robustness analysis on NAVSIM-v2 with LTF. We corrupt the predicted TE representation at inference time by adding TE depth noise, randomly dropping TE detections, or injecting false-positive TE predictions. The performance decreases smoothly under stronger corruption while remaining above the LTF baseline, indicating robustness to moderate upstream TE perception noise. } 
\label{fig:te_robustness} 
\end{figure}
\subsubsection{Robustness and semantic analysis.}
Since our framework uses predicted traffic elements at inference time, we further evaluate whether the planning gain is robust to upstream perception noise. We perform inference-stage sensitivity tests on NAVSIM-v2 with LTF by corrupting the predicted TE representation before it is fed to the planner. Specifically, we consider three common failure modes: noisy TE depth, missed TE detections, and false-positive TE predictions. As shown in Fig.~\ref{fig:te_robustness}, the planning score degrades smoothly as the corruption severity increases, but remains consistently above the original LTF baseline. This indicates that the proposed TE-aware planner does not rely on perfectly accurate TE predictions and is reasonably resilient to moderate upstream perception errors. We also verify that the improvement is not merely caused by generic multi-task regularization. To this end, we train a class-agnostic TE variant that only supervises TE localization with the L1 loss and removes the semantic classification loss. This variant reaches only 26.2 EPDMS on NAVSIM-v2, compared with 28.9 EPDMS for the full TE-aware model. The gap shows that regulatory semantics, rather than spatial localization alone, are critical for planning.


\subsubsection{Ablation on Topology Information Integration.} We systematically evaluate our topology integration strategies in Tab.~\ref{tab:topo ablation} to determine the most effective mechanism for routing topological priors into the end-to-end planner.

(1) \textbf{Topology Synergy.} Introducing either centerline-to-centerline connectivity (LCLC, ID 1) or centerline-to-traffic-element constraints (LCTE, ID 2) substantially improves planning metrics. Combining both (ID 3) provides the most comprehensive spatial-logical context, yielding better performance.

(2) \textbf{Encoding Strategy: Language vs. Graph.} When encoding the merged directed topology graph, language-based encoding via BERT~\cite{devlin2019bert} (ID 3) outperforms Graph Convolutional Networks (GCN~\cite{schlichtkrull2018gcn}, ID 4), particularly in reducing the Collision Rate. Driving topologies consist of highly heterogeneous nodes (e.g., continuous spatial centerlines vs. discrete logical traffic lights). While GCN message-passing tends to over-smooth these semantic features, language models naturally map these heterogeneous attributes into a unified semantic space. This prevents information loss and perfectly preserves long-range causal driving rules.

(3) \textbf{Interaction Scope: Ego vs. Global.} Restricting the topology to ego-relevant elements (ID 4) proves superior to utilizing the global topology (ID 5). Global structures introduce redundant noise from irrelevant lanes and unaffected traffic elements, which distract the planner's attention mechanism. Ego-centric filtering ensures the model focuses solely on actionable causal relationships.

(4) \textbf{Robustness to Predicted Topology.} Finally, replacing the ground-truth (GT) topology~\cite{wang2023openlane} with real-time TopoMLP~\cite{wu2023topomlp} predictions (ID 6, our final optimal setting) maintains performance comparable to the GT oracle (ID 3). This demonstrates the strong robustness of our interaction mechanism, proving it can effectively guide safe trajectory generation without relying on perfectly accurate perception priors.
\begin{table}[t]
    \caption{Ablation study of \textbf{topology information integration} on the \textbf{nuScenes} dataset. We ablate the topology types, encoding methods, topology source and interaction scope upon the \textbf{VAD} baseline.}
    \vspace{-0.8cm}
    \begin{center}
    \small
    \resizebox{\linewidth}{!}{
    \begin{tabular}{l | cc | cc | cc | cc | cccc  cccc }
        \toprule
        \multirow{2}{*}{ID} & \multicolumn{2}{c|}{Topology} & \multicolumn{2}{c|}{Encoder} & \multicolumn{2}{c|}{Source} & \multicolumn{2}{c|}{Scope} & \multicolumn{4}{c}{L2 (m) $\downarrow$} & \multicolumn{4}{c}{Collision Rate (\%) $\downarrow$} \\
        \cmidrule(lr){2-3} \cmidrule(lr){4-5} \cmidrule(lr){6-7} \cmidrule(lr){8-9} \cmidrule(lr){10-13} \cmidrule(lr){14-17}
        & LCTE & LCLC & BERT & GCN & GT~\cite{wang2023openlane} & Pred~\cite{wu2023topomlp} & Global & Ego & 1s & 2s & 3s & Avg. & 1s & 2s & 3s & Avg. \\ \midrule
        0 &  &  &  &  &  &  &  &  & 0.41 & 0.70 & 1.05 & 0.72 & 0.07 & 0.17 & 0.41 & 0.22  \\
        1 &  & \cmark & \cmark &  & \cmark &  &  & \cmark & 0.34 & 0.60 & 0.95 & 0.63 & 0.06 & 0.14 & 0.27 & \textbf{0.16}\\
        2 & \cmark &  & \cmark &  & \cmark &  &  & \cmark & 0.34 & 0.59 & 0.93 & \textbf{0.62} & 0.06 & 0.16 & 0.33 & 0.18 \\
        3 & \cmark & \cmark & \cmark &  & \cmark &  &  & \cmark & 0.33 & 0.58 & 0.94 & \textbf{0.62} & 0.04 & 0.16 & 0.28 & \textbf{0.16} \\
        4 & \cmark & \cmark &  & \cmark & \cmark &  &  & \cmark & 0.34 & 0.60 & 0.93 & \textbf{0.62} & 0.07 & 0.18 & 0.43 & 0.23 \\
        5 & \cmark & \cmark &  & \cmark & \cmark &  & \cmark &  & 0.34 & 0.61 & 0.96 & 0.64 & 0.33 & 0.50 & 0.68 &  0.50 \\    
        \rowcolor{myblue}
        6 & \cmark & \cmark & \cmark &  &  & \cmark &  & \cmark & 0.34 & 0.59 & 0.92 & \textbf{0.62} & 0.05 & 0.15 & 0.29 &  \textbf{0.16}\\
        \bottomrule
    \end{tabular}
    }
    \label{tab:topo ablation}
    \end{center}
\end{table}

\vspace{-0.3cm}

\begin{wrapfigure}{r}{0.5\textwidth}
    \vspace{-5pt}
    \centering
    \includegraphics[width=1\linewidth]{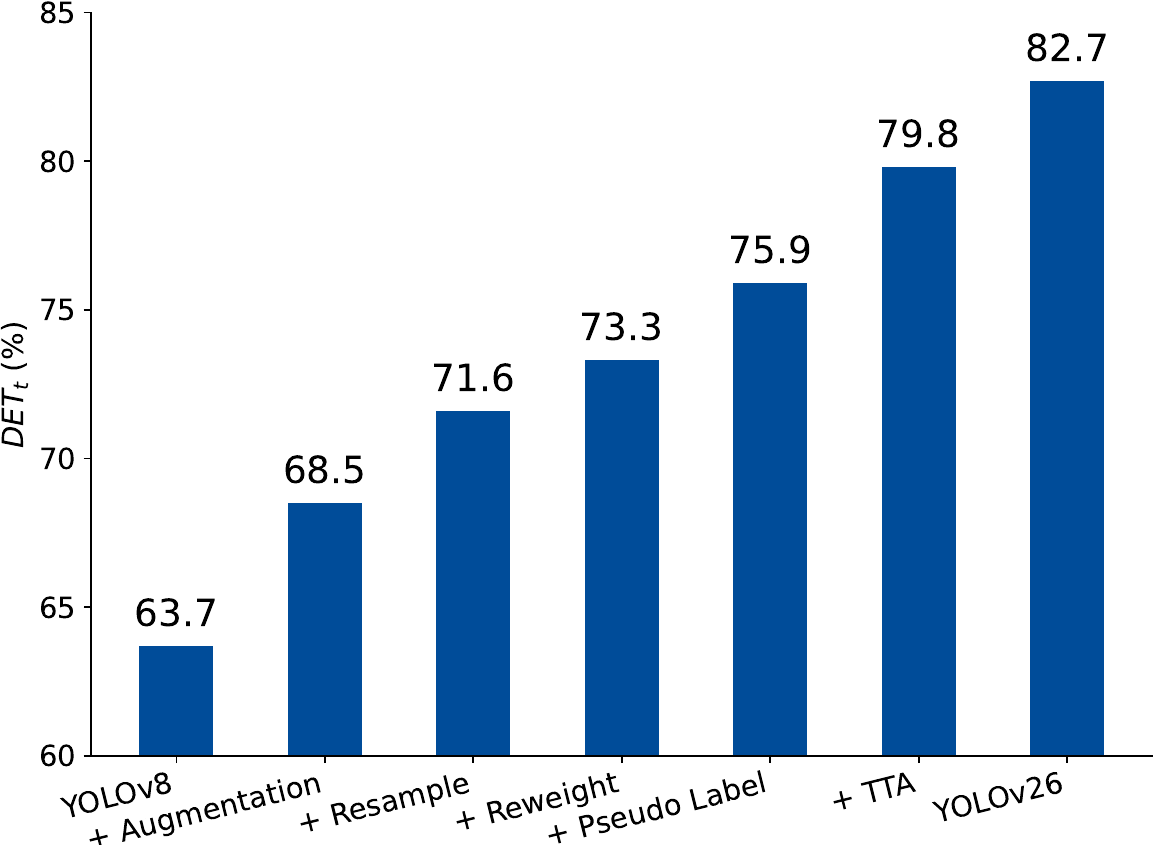}
    \vspace{-1.5em}
    \caption{\textbf{Stepwise performance improvements} of our traffic element detector on the OpenLane-V2 validation set, evaluated by the $\text{DET}_t$ metric.}
    \label{fig:detection_ablation}
    \vspace{-20pt}
\end{wrapfigure}
\subsubsection{Ablation on Traffic Element Detection Optimization.} To obtain a highly robust traffic element detector for datasets lacking TE coordinates annotations (e.g., NAVSIM), we progressively optimize a baseline YOLOv8~\cite{redmon2016yolo} detector using datasets with available ground truth. As illustrated in Fig.~\ref{fig:detection_ablation}, we progressively optimize the baseline detector into a robust feature extractor for the downstream planner by systematically applying strong data augmentation, class-balanced resampling, loss reweighting, pseudo-labeling, test-time augmentation (TTA), and an advanced YOLO backbone~\cite{sapkota2025yolo26}.

\section{Conclusion}
\label{sec: conclusion}
We propose a lightweight way to inject traffic elements (lights/signs) and lane topology into vision-based end-to-end planning. By constructing a unified 3D traffic-element representation, adding TE auxiliary supervision, and conditioning the planner with ego-centric topology encoded as structured language, we obtain consistent gains across four benchmarks (open-loop and closed-loop) and six representative planners, improving both trajectory accuracy and safety/compliance metrics with only minor runtime overhead. Importantly, our results demonstrate strong scalability: the gains persist with stronger backbones and further increase when training with more data, highlighting traffic-element reasoning as a previously under-emphasized but decision-critical factor in end-to-end driving.

%
%
\bibliographystyle{splncs04}
\bibliography{main}

@String(CVPR  = {IEEE Conf. Comput. Vis. Pattern Recog.})

@String(AAAI  = {AAAI})

@String(CVPR  = {CVPR})

@article{li2025ztrs,
  title={Ztrs: Zero-imitation end-to-end autonomous driving with trajectory scoring},
  author={Li, Zhenxin and Yao, Wenhao and Wang, Zi and Sun, Xinglong and Chen, Jingde and Chang, Nadine and Shen, Maying and Song, Jingyu and Wu, Zuxuan and Lan, Shiyi and others},
  journal={arXiv preprint arXiv:2510.24108},
  year={2025}
}

@article{jiang2025diffvla,
  title={Diffvla: Vision-language guided diffusion planning for autonomous driving},
  author={Jiang, Anqing and Gao, Yu and Sun, Zhigang and Wang, Yiru and Wang, Jijun and Chai, Jinghao and Cao, Qian and Heng, Yuweng and Jiang, Hao and Dong, Yunda and others},
  journal={arXiv preprint arXiv:2505.19381},
  year={2025}
}

@article{liu2025guideflow,
  title={GuideFlow: Constraint-guided flow matching for planning in end-to-end autonomous driving},
  author={Liu, Lin and Jia, Caiyan and Yu, Guanyi and Song, Ziying and Li, JunQiao and Jia, Feiyang and Wu, Peiliang and Hao, Xiaoshuai and Luo, Yadan},
  journal={arXiv preprint arXiv:2511.18729},
  year={2025}
}

@article{chitta2022transfuser,
  title={Transfuser: Imitation with transformer-based sensor fusion for autonomous driving},
  author={Chitta, Kashyap and Prakash, Aditya and Jaeger, Bernhard and Yu, Zehao and Renz, Katrin and Geiger, Andreas},
  journal={IEEE transactions on pattern analysis and machine intelligence},
  volume={45},
  number={11},
  pages={12878--12895},
  year={2022},
  publisher={IEEE}
}

@inproceedings{liao2025diffusiondrive,
  title={Diffusiondrive: Truncated diffusion model for end-to-end autonomous driving},
  author={Liao, Bencheng and Chen, Shaoyu and Yin, Haoran and Jiang, Bo and Wang, Cheng and Yan, Sixu and Zhang, Xinbang and Li, Xiangyu and Zhang, Ying and Zhang, Qian and others},
  booktitle={Proceedings of the Computer Vision and Pattern Recognition Conference},
  pages={12037--12047},
  year={2025}
}

@article{li2025generalized,
  title={Generalized trajectory scoring for end-to-end multimodal planning},
  author={Li, Zhenxin and Yao, Wenhao and Wang, Zi and Sun, Xinglong and Chen, Joshua and Chang, Nadine and Shen, Maying and Wu, Zuxuan and Lan, Shiyi and Alvarez, Jose M},
  journal={arXiv preprint arXiv:2506.06664},
  year={2025}
}

@article{wu2022trajectory,
  title={Trajectory-guided control prediction for end-to-end autonomous driving: A simple yet strong baseline},
  author={Wu, Penghao and Jia, Xiaosong and Chen, Li and Yan, Junchi and Li, Hongyang and Qiao, Yu},
  journal={Advances in Neural Information Processing Systems},
  volume={35},
  pages={6119--6132},
  year={2022}
}

@inproceedings{jia2023thinktwice,
  title={Think twice before driving: Towards scalable decoders for end-to-end autonomous driving},
  author={Jia, Xiaosong and Wu, Penghao and Chen, Li and Xie, Jiangwei and He, Conghui and Yan, Junchi and Li, Hongyang},
  booktitle={Proceedings of the IEEE/CVF Conference on Computer Vision and Pattern Recognition},
  pages={21983--21994},
  year={2023}
}

@inproceedings{jia2023driveadapter,
  title={Driveadapter: Breaking the coupling barrier of perception and planning in end-to-end autonomous driving},
  author={Jia, Xiaosong and Gao, Yulu and Chen, Li and Yan, Junchi and Liu, Patrick Langechuan and Li, Hongyang},
  booktitle={Proceedings of the IEEE/CVF International Conference on Computer Vision},
  pages={7953--7963},
  year={2023}
}

@article{zhai2023ADMLP,
  title={Rethinking the open-loop evaluation of end-to-end autonomous driving in nuscenes},
  author={Zhai, Jiang-Tian and Feng, Ze and Du, Jinhao and Mao, Yongqiang and Liu, Jiang-Jiang and Tan, Zichang and Zhang, Yifu and Ye, Xiaoqing and Wang, Jingdong},
  journal={arXiv preprint arXiv:2305.10430},
  year={2023}
}

@inproceedings{hu2023planning,
  title={Planning-oriented autonomous driving},
  author={Hu, Yihan and Yang, Jiazhi and Chen, Li and Li, Keyu and Sima, Chonghao and Zhu, Xizhou and Chai, Siqi and Du, Senyao and Lin, Tianwei and Wang, Wenhai and others},
  booktitle={Proceedings of the IEEE/CVF conference on computer vision and pattern recognition},
  pages={17853--17862},
  year={2023}
}

@inproceedings{jiang2023vad,
  title={Vad: Vectorized scene representation for efficient autonomous driving},
  author={Jiang, Bo and Chen, Shaoyu and Xu, Qing and Liao, Bencheng and Chen, Jiajie and Zhou, Helong and Zhang, Qian and Liu, Wenyu and Huang, Chang and Wang, Xinggang},
  booktitle={Proceedings of the IEEE/CVF International Conference on Computer Vision},
  pages={8340--8350},
  year={2023}
}

@article{jia2025drivetransformer,
  title={Drivetransformer: Unified transformer for scalable end-to-end autonomous driving},
  author={Jia, Xiaosong and You, Junqi and Zhang, Zhiyuan and Yan, Junchi},
  journal={arXiv preprint arXiv:2503.07656},
  year={2025}
}

@inproceedings{dauner2023pdm,
  title={Parting with misconceptions about learning-based vehicle motion planning},
  author={Dauner, Daniel and Hallgarten, Marcel and Geiger, Andreas and Chitta, Kashyap},
  booktitle={Conference on Robot Learning},
  pages={1268--1281},
  year={2023},
  organization={PMLR}
}

@article{dauner2024navsim,
  title={Navsim: Data-driven non-reactive autonomous vehicle simulation and benchmarking},
  author={Dauner, Daniel and Hallgarten, Marcel and Li, Tianyu and Weng, Xinshuo and Huang, Zhiyu and Yang, Zetong and Li, Hongyang and Gilitschenski, Igor and Ivanovic, Boris and Pavone, Marco and others},
  journal={Advances in Neural Information Processing Systems},
  volume={37},
  pages={28706--28719},
  year={2024}
}

@article{wang2025unified,
  title={Unified vision-language-action model},
  author={Wang, Yuqi and Li, Xinghang and Wang, Wenxuan and Zhang, Junbo and Li, Yingyan and Chen, Yuntao and Wang, Xinlong and Zhang, Zhaoxiang},
  journal={arXiv preprint arXiv:2506.19850},
  year={2025}
}

@inproceedings{chen2025drivinggpt,
  title={Drivinggpt: Unifying driving world modeling and planning with multi-modal autoregressive transformers},
  author={Chen, Yuntao and Wang, Yuqi and Zhang, Zhaoxiang},
  booktitle={Proceedings of the IEEE/CVF International Conference on Computer Vision},
  pages={26890--26900},
  year={2025}
}

@inproceedings{paradrive,
  title={Para-drive: Parallelized architecture for real-time autonomous driving},
  author={Weng, Xinshuo and Ivanovic, Boris and Wang, Yan and Wang, Yue and Pavone, Marco},
  booktitle={Proceedings of the IEEE/CVF Conference on Computer Vision and Pattern Recognition},
  pages={15449--15458},
  year={2024}
}

@inproceedings{shi2025drivex,
  title={Drivex: Omni scene modeling for learning generalizable world knowledge in autonomous driving},
  author={Shi, Chen and Shi, Shaoshuai and Sheng, Kehua and Zhang, Bo and Jiang, Li},
  booktitle={Proceedings of the IEEE/CVF International Conference on Computer Vision},
  pages={28599--28609},
  year={2025}
}

@inproceedings{zheng2025world4drive,
  title={World4drive: End-to-end autonomous driving via intention-aware physical latent world model},
  author={Zheng, Yupeng and Yang, Pengxuan and Xing, Zebin and Zhang, Qichao and Zheng, Yuhang and Gao, Yinfeng and Li, Pengfei and Zhang, Teng and Xia, Zhongpu and Jia, Peng and others},
  booktitle={Proceedings of the IEEE/CVF International Conference on Computer Vision},
  pages={28632--28642},
  year={2025}
}

@article{yuan2024drama,
  title={Drama: An efficient end-to-end motion planner for autonomous driving with mamba},
  author={Yuan, Chengran and Zhang, Zhanqi and Sun, Jiawei and Sun, Shuo and Huang, Zefan and Lee, Christina Dao Wen and Li, Dongen and Han, Yuhang and Wong, Anthony and Tee, Keng Peng and others},
  journal={arXiv preprint arXiv:2408.03601},
  year={2024}
}

@article{chen2024vadv2,
  title={Vadv2: End-to-end vectorized autonomous driving via probabilistic planning},
  author={Chen, Shaoyu and Jiang, Bo and Gao, Hao and Liao, Bencheng and Xu, Qing and Zhang, Qian and Huang, Chang and Liu, Wenyu and Wang, Xinggang},
  journal={arXiv preprint arXiv:2402.13243},
  year={2024}
}

@article{wozniak2025prix,
  title={PRIX: Learning to plan from raw pixels for end-to-end autonomous driving},
  author={Wozniak, Maciej K and Liu, Lianhang and Cai, Yixi and Jensfelt, Patric},
  journal={arXiv preprint arXiv:2507.17596},
  year={2025}
}

@article{song2025breaking,
  title={Breaking imitation bottlenecks: Reinforced diffusion powers diverse trajectory generation},
  author={Song, Ziying and Liu, Lin and Pan, Hongyu and Liao, Bencheng and Guo, Mingzhe and Yang, Lei and Zhang, Yongchang and Xu, Shaoqing and Jia, Caiyan and Luo, Yadan},
  journal={arXiv e-prints},
  pages={arXiv--2507},
  year={2025}
}

@article{zhou2025autovla,
  title={Autovla: A vision-language-action model for end-to-end autonomous driving with adaptive reasoning and reinforcement fine-tuning},
  author={Zhou, Zewei and Cai, Tianhui and Zhao, Seth Z and Zhang, Yun and Huang, Zhiyu and Zhou, Bolei and Ma, Jiaqi},
  journal={arXiv preprint arXiv:2506.13757},
  year={2025}
}

@article{li2025drivevla,
  title={DriveVLA-W0: World models amplify data scaling law in autonomous driving},
  author={Li, Yingyan and Shang, Shuyao and Liu, Weisong and Zhan, Bing and Wang, Haochen and Wang, Yuqi and Chen, Yuntao and Wang, Xiaoman and An, Yasong and Tang, Chufeng and others},
  journal={arXiv preprint arXiv:2510.12796},
  year={2025}
}

@article{li2025recogdrive,
  title={Recogdrive: A reinforced cognitive framework for end-to-end autonomous driving},
  author={Li, Yongkang and Xiong, Kaixin and Guo, Xiangyu and Li, Fang and Yan, Sixu and Xu, Gangwei and Zhou, Lijun and Chen, Long and Sun, Haiyang and Wang, Bing and others},
  journal={arXiv preprint arXiv:2506.08052},
  year={2025}
}

@article{li2024hydra,
  title={Hydra-mdp: End-to-end multimodal planning with multi-target hydra-distillation},
  author={Li, Zhenxin and Li, Kailin and Wang, Shihao and Lan, Shiyi and Yu, Zhiding and Ji, Yishen and Li, Zhiqi and Zhu, Ziyue and Kautz, Jan and Wu, Zuxuan and others},
  journal={arXiv preprint arXiv:2406.06978},
  year={2024}
}

@article{sima2025centaur,
  title={Centaur: Robust end-to-end autonomous driving with test-time training},
  author={Sima, Chonghao and Chitta, Kashyap and Yu, Zhiding and Lan, Shiyi and Luo, Ping and Geiger, Andreas and Li, Hongyang and Alvarez, Jose M},
  journal={arXiv preprint arXiv:2503.11650},
  year={2025}
}

@article{yao2025drivesuprim,
  title={Drivesuprim: Towards precise trajectory selection for end-to-end planning},
  author={Yao, Wenhao and Li, Zhenxin and Lan, Shiyi and Wang, Zi and Sun, Xinglong and Alvarez, Jose M and Wu, Zuxuan},
  journal={arXiv preprint arXiv:2506.06659},
  year={2025}
}

@article{kirby2026drivoR,
  title={Driving on Registers},
  author={Kirby, Ellington and Boulch, Alexandre and Xu, Yihong and Yin, Yuan and Puy, Gilles and Zablocki, {\'E}loi and Bursuc, Andrei and Gidaris, Spyros and Marlet, Renaud and Bartoccioni, Florent and others},
  journal={arXiv preprint arXiv:2601.05083},
  year={2026}
}

@article{tian2025simscale,
  title={Simscale: Learning to drive via real-world simulation at scale},
  author={Tian, Haochen and Li, Tianyu and Liu, Haochen and Yang, Jiazhi and Qiu, Yihang and Li, Guang and Wang, Junli and Gao, Yinfeng and Zhang, Zhang and Wang, Liang and others},
  journal={arXiv preprint arXiv:2511.23369},
  year={2025}
}

@inproceedings{zeng2019nmp,
  title={End-to-end interpretable neural motion planner},
  author={Zeng, Wenyuan and Luo, Wenjie and Suo, Simon and Sadat, Abbas and Yang, Bin and Casas, Sergio and Urtasun, Raquel},
  booktitle={Proceedings of the IEEE/CVF conference on computer vision and pattern recognition},
  pages={8660--8669},
  year={2019}
}

@inproceedings{hu2021ff,
  title={Safe local motion planning with self-supervised freespace forecasting},
  author={Hu, Peiyun and Huang, Aaron and Dolan, John and Held, David and Ramanan, Deva},
  booktitle={Proceedings of the IEEE/CVF Conference on Computer Vision and Pattern Recognition},
  pages={12732--12741},
  year={2021}
}

@inproceedings{khurana2022eo,
  title={Differentiable raycasting for self-supervised occupancy forecasting},
  author={Khurana, Tarasha and Hu, Peiyun and Dave, Achal and Ziglar, Jason and Held, David and Ramanan, Deva},
  booktitle={European Conference on Computer Vision},
  pages={353--369},
  year={2022},
  organization={Springer}
}

@inproceedings{hu2022stp3,
  title={St-p3: End-to-end vision-based autonomous driving via spatial-temporal feature learning},
  author={Hu, Shengchao and Chen, Li and Wu, Penghao and Li, Hongyang and Yan, Junchi and Tao, Dacheng},
  booktitle={European Conference on Computer Vision},
  pages={533--549},
  year={2022},
  organization={Springer}
}

@inproceedings{tong2023scene,
  title={Scene as occupancy},
  author={Tong, Wenwen and Sima, Chonghao and Wang, Tai and Chen, Li and Wu, Silei and Deng, Hanming and Gu, Yi and Lu, Lewei and Luo, Ping and Lin, Dahua and others},
  booktitle={Proceedings of the IEEE/CVF International Conference on Computer Vision},
  pages={8406--8415},
  year={2023}
}

@inproceedings{zheng2024genad,
  title={Genad: Generative end-to-end autonomous driving},
  author={Zheng, Wenzhao and Song, Ruiqi and Guo, Xianda and Zhang, Chenming and Chen, Long},
  booktitle={European Conference on Computer Vision},
  pages={87--104},
  year={2024},
  organization={Springer}
}

@article{guo2025uad,
  title={End-to-end autonomous driving without costly modularization and 3d manual annotation},
  author={Guo, Mingzhe and Zhang, Zhipeng and He, Yuan and Wang, Ke and Jing, Liping and Ling, Haibin},
  journal={IEEE Transactions on Pattern Analysis and Machine Intelligence},
  year={2025},
  publisher={IEEE}
}

@article{li2024ssr,
  title={Navigation-guided sparse scene representation for end-to-end autonomous driving},
  author={Li, Peidong and Cui, Dixiao},
  journal={arXiv preprint arXiv:2409.18341},
  year={2024}
}

@inproceedings{li2024bevplanner,
  title={Is ego status all you need for open-loop end-to-end autonomous driving?},
  author={Li, Zhiqi and Yu, Zhiding and Lan, Shiyi and Li, Jiahan and Kautz, Jan and Lu, Tong and Alvarez, Jose M},
  booktitle={Proceedings of the IEEE/CVF Conference on Computer Vision and Pattern Recognition},
  pages={14864--14873},
  year={2024}
}

@article{li2024law,
  title={Enhancing end-to-end autonomous driving with latent world model},
  author={Li, Yingyan and Fan, Lue and He, Jiawei and Wang, Yuqi and Chen, Yuntao and Zhang, Zhaoxiang and Tan, Tieniu},
  journal={arXiv preprint arXiv:2406.08481},
  year={2024}
}

@inproceedings{sun2025sparsedrive,
  title={Sparsedrive: End-to-end autonomous driving via sparse scene representation},
  author={Sun, Wenchao and Lin, Xuewu and Shi, Yining and Zhang, Chuang and Wu, Haoran and Zheng, Sifa},
  booktitle={2025 IEEE International Conference on Robotics and Automation (ICRA)},
  pages={8795--8801},
  year={2025},
  organization={IEEE}
}

@inproceedings{song2025momad,
  title={Don't shake the wheel: Momentum-aware planning in end-to-end autonomous driving},
  author={Song, Ziying and Jia, Caiyan and Liu, Lin and Pan, Hongyu and Zhang, Yongchang and Wang, Junming and Zhang, Xingyu and Xu, Shaoqing and Yang, Lei and Luo, Yadan},
  booktitle={Proceedings of the IEEE/CVF Conference on Computer Vision and Pattern Recognition},
  pages={22432--22441},
  year={2025}
}

@article{Qwen2.5-VL,
  title={Qwen2.5-VL Technical Report},
  author={Bai, Shuai and Chen, Keqin and Liu, Xuejing and Wang, Jialin and Ge, Wenbin and Song, Sibo and Dang, Kai and Wang, Peng and Wang, Shijie and Tang, Jun and Zhong, Humen and Zhu, Yuanzhi and Yang, Mingkun and Li, Zhaohai and Wan, Jianqiang and Wang, Pengfei and Ding, Wei and Fu, Zheren and Xu, Yiheng and Ye, Jiabo and Zhang, Xi and Xie, Tianbao and Cheng, Zesen and Zhang, Hang and Yang, Zhibo and Xu, Haiyang and Lin, Junyang},
  journal={arXiv preprint arXiv:2502.13923},
  year={2025}
}

@article{jiang2024senna,
  title={Senna: Bridging large vision-language models and end-to-end autonomous driving},
  author={Jiang, Bo and Chen, Shaoyu and Liao, Bencheng and Zhang, Xingyu and Yin, Wei and Zhang, Qian and Huang, Chang and Liu, Wenyu and Wang, Xinggang},
  journal={arXiv preprint arXiv:2410.22313},
  year={2024}
}

@article{tian2024drivevlm,
  title={Drivevlm: The convergence of autonomous driving and large vision-language models},
  author={Tian, Xiaoyu and Gu, Junru and Li, Bailin and Liu, Yicheng and Wang, Yang and Zhao, Zhiyong and Zhan, Kun and Jia, Peng and Lang, Xianpeng and Zhao, Hang},
  journal={arXiv preprint arXiv:2402.12289},
  year={2024}
}

@inproceedings{wang2025omnidrive,
  title={Omnidrive: A holistic vision-language dataset for autonomous driving with counterfactual reasoning},
  author={Wang, Shihao and Yu, Zhiding and Jiang, Xiaohui and Lan, Shiyi and Shi, Min and Chang, Nadine and Kautz, Jan and Li, Ying and Alvarez, Jose M},
  booktitle={Proceedings of the computer vision and pattern recognition conference},
  pages={22442--22452},
  year={2025}
}

@article{hwang2024emma,
  title={Emma: End-to-end multimodal model for autonomous driving},
  author={Hwang, Jyh-Jing and Xu, Runsheng and Lin, Hubert and Hung, Wei-Chih and Ji, Jingwei and Choi, Kristy and Huang, Di and He, Tong and Covington, Paul and Sapp, Benjamin and others},
  journal={arXiv preprint arXiv:2410.23262},
  year={2024}
}

@inproceedings{fu2025orion,
  title={Orion: A holistic end-to-end autonomous driving framework by vision-language instructed action generation},
  author={Fu, Haoyu and Zhang, Diankun and Zhao, Zongchuang and Cui, Jianfeng and Liang, Dingkang and Zhang, Chong and Zhang, Dingyuan and Xie, Hongwei and Wang, Bing and Bai, Xiang},
  booktitle={Proceedings of the IEEE/CVF International Conference on Computer Vision},
  pages={24823--24834},
  year={2025}
}

@article{chi2025impromptu,
  title={Impromptu vla: Open weights and open data for driving vision-language-action models},
  author={Chi, Haohan and Gao, Huan-ang and Liu, Ziming and Liu, Jianing and Liu, Chenyu and Li, Jinwei and Yang, Kaisen and Yu, Yangcheng and Wang, Zeda and Li, Wenyi and others},
  journal={arXiv preprint arXiv:2505.23757},
  year={2025}
}

@inproceedings{he2016resnet,
  title={Deep residual learning for image recognition},
  author={He, Kaiming and Zhang, Xiangyu and Ren, Shaoqing and Sun, Jian},
  booktitle={Proceedings of the IEEE conference on computer vision and pattern recognition},
  pages={770--778},
  year={2016}
}

@article{wu20231sttopology,
  title={The 1st-place solution for cvpr 2023 openlane topology in autonomous driving challenge},
  author={Wu, Dongming and Jia, Fan and Chang, Jiahao and Li, Zhuoling and Sun, Jianjian and Han, Chunrui and Li, Shuailin and Liu, Yingfei and Ge, Zheng and Wang, Tiancai},
  journal={arXiv preprint arXiv:2306.09590},
  year={2023}
}

@article{wang2023openlane,
  title={Openlane-v2: A topology reasoning benchmark for unified 3d hd mapping},
  author={Wang, Huijie and Li, Tianyu and Li, Yang and Chen, Li and Sima, Chonghao and Liu, Zhenbo and Wang, Bangjun and Jia, Peijin and Wang, Yuting and Jiang, Shengyin and others},
  journal={Advances in Neural Information Processing Systems},
  volume={36},
  pages={18873--18884},
  year={2023}
}

@inproceedings{caesar2020nuscenes,
  title={nuscenes: A multimodal dataset for autonomous driving},
  author={Caesar, Holger and Bankiti, Varun and Lang, Alex H and Vora, Sourabh and Liong, Venice Erin and Xu, Qiang and Krishnan, Anush and Pan, Yu and Baldan, Giancarlo and Beijbom, Oscar},
  booktitle={Proceedings of the IEEE/CVF conference on computer vision and pattern recognition},
  pages={11621--11631},
  year={2020}
}

@article{jia2024bench2drive,
  title={Bench2drive: Towards multi-ability benchmarking of closed-loop end-to-end autonomous driving},
  author={Jia, Xiaosong and Yang, Zhenjie and Li, Qifeng and Zhang, Zhiyuan and Yan, Junchi},
  journal={Advances in Neural Information Processing Systems},
  volume={37},
  pages={819--844},
  year={2024}
}

@inproceedings{dosovitskiy2017carla,
  title={CARLA: An open urban driving simulator},
  author={Dosovitskiy, Alexey and Ros, German and Codevilla, Felipe and Lopez, Antonio and Koltun, Vladlen},
  booktitle={Conference on robot learning},
  pages={1--16},
  year={2017},
  organization={PMLR}
}

@article{cao2025navsimv2,
  title={Pseudo-simulation for autonomous driving},
  author={Cao, Wei and Hallgarten, Marcel and Li, Tianyu and Dauner, Daniel and Gu, Xunjiang and Wang, Caojun and Miron, Yakov and Aiello, Marco and Li, Hongyang and Gilitschenski, Igor and others},
  journal={arXiv preprint arXiv:2506.04218},
  year={2025}
}

@article{wu2023topomlp,
  title={Topomlp: A simple yet strong pipeline for driving topology reasoning},
  author={Wu, Dongming and Chang, Jiahao and Jia, Fan and Liu, Yingfei and Wang, Tiancai and Shen, Jianbing},
  journal={arXiv preprint arXiv:2310.06753},
  year={2023}
}

@article{li2023toponet,
  title={Graph-based topology reasoning for driving scenes},
  author={Li, Tianyu and Chen, Li and Wang, Huijie and Li, Yang and Yang, Jiazhi and Geng, Xiangwei and Jiang, Shengyin and Wang, Yuting and Xu, Hang and Xu, Chunjing and others},
  journal={arXiv preprint arXiv:2304.05277},
  year={2023}
}

@inproceedings{lv2025t2sg,
  title={T2sg: Traffic topology scene graph for topology reasoning in autonomous driving},
  author={Lv, Changsheng and Qi, Mengshi and Liu, Liang and Ma, Huadong},
  booktitle={Proceedings of the Computer Vision and Pattern Recognition Conference},
  pages={17197--17206},
  year={2025}
}

@article{fu2024topologic,
  title={Topologic: An interpretable pipeline for lane topology reasoning on driving scenes},
  author={Fu, Yanping and Liao, Wenbin and Liu, Xinyuan and Xu, Hang and Ma, Yike and Zhang, Yucheng and Dai, Feng},
  journal={Advances in Neural Information Processing Systems},
  volume={37},
  pages={61658--61676},
  year={2024}
}

@inproceedings{li2025reusing,
  title={Reusing attention for one-stage lane topology understanding},
  author={Li, Yang and Zhang, Zongzheng and Qiu, Xuchong and Li, Xinrun and Liu, Ziming and Wang, Leichen and Li, Ruikai and Zhu, Zhenxin and Gao, Huan-ang and Lin, Xiaojian and others},
  booktitle={2025 IEEE/RSJ International Conference on Intelligent Robots and Systems (IROS)},
  pages={16977--16984},
  year={2025},
  organization={IEEE}
}

@article{li2023lanesegnet,
  title={Lanesegnet: Map learning with lane segment perception for autonomous driving},
  author={Li, Tianyu and Jia, Peijin and Wang, Bangjun and Chen, Li and Jiang, Kun and Yan, Junchi and Li, Hongyang},
  journal={arXiv preprint arXiv:2312.16108},
  year={2023}
}

@inproceedings{chang2025mapdr,
  title={Driving by the rules: A benchmark for integrating traffic sign regulations into vectorized hd map},
  author={Chang, Xinyuan and Xue, Maixuan and Liu, Xinran and Pan, Zheng and Wei, Xing},
  booktitle={Proceedings of the Computer Vision and Pattern Recognition Conference},
  pages={6823--6833},
  year={2025}
}

@inproceedings{devlin2019bert,
  title={Bert: Pre-training of deep bidirectional transformers for language understanding},
  author={Devlin, Jacob and Chang, Ming-Wei and Lee, Kenton and Toutanova, Kristina},
  booktitle={Proceedings of the 2019 conference of the North American chapter of the association for computational linguistics: human language technologies, volume 1 (long and short papers)},
  pages={4171--4186},
  year={2019}
}

@article{lin2025depthanything3,
  title={Depth anything 3: Recovering the visual space from any views},
  author={Lin, Haotong and Chen, Sili and Liew, Junhao and Chen, Donny Y and Li, Zhenyu and Shi, Guang and Feng, Jiashi and Kang, Bingyi},
  journal={arXiv preprint arXiv:2511.10647},
  year={2025}
}

@article{piccinelli2025unidepthv2,
  title={Unidepthv2: Universal monocular metric depth estimation made simpler},
  author={Piccinelli, Luigi and Sakaridis, Christos and Yang, Yung-Hsu and Segu, Mattia and Li, Siyuan and Abbeloos, Wim and Van Gool, Luc},
  journal={IEEE Transactions on Pattern Analysis and Machine Intelligence},
  year={2025},
  publisher={IEEE}
}

@article{sapkota2025yolo26,
  title={YOLO26: key architectural enhancements and performance benchmarking for real-time object detection},
  author={Sapkota, Ranjan and Cheppally, Rahul Harsha and Sharda, Ajay and Karkee, Manoj},
  journal={arXiv preprint arXiv:2509.25164},
  year={2025}
}

@article{ge2021yolox,
  title={Yolox: Exceeding yolo series in 2021},
  author={Ge, Zheng and Liu, Songtao and Wang, Feng and Li, Zeming and Sun, Jian},
  journal={arXiv preprint arXiv:2107.08430},
  year={2021}
}

@inproceedings{redmon2016yolo,
  title={You only look once: Unified, real-time object detection},
  author={Redmon, Joseph and Divvala, Santosh and Girshick, Ross and Farhadi, Ali},
  booktitle={Proceedings of the IEEE conference on computer vision and pattern recognition},
  pages={779--788},
  year={2016}
}

@inproceedings{schlichtkrull2018gcn,
  title={Modeling relational data with graph convolutional networks},
  author={Schlichtkrull, Michael and Kipf, Thomas N and Bloem, Peter and Van Den Berg, Rianne and Titov, Ivan and Welling, Max},
  booktitle={European Semantic Web Conference},
  pages={593--607},
  year={2018},
  organization={Springer}
}

@article{zou2025diffusiondrivev2,
  title={DiffusionDriveV2: Reinforcement learning-constrained truncated diffusion modeling in end-to-end autonomous driving},
  author={Zou, Jialv and Chen, Shaoyu and Liao, Bencheng and Zheng, Zhiyu and Song, Yuehao and Zhang, Lefei and Zhang, Qian and Liu, Wenyu and Wang, Xinggang},
  journal={arXiv preprint arXiv:2512.07745},
  year={2025}
}

@inproceedings{xing2025goalflow,
  title={Goalflow: Goal-driven flow matching for multimodal trajectories generation in end-to-end autonomous driving},
  author={Xing, Zebin and Zhang, Xingyu and Hu, Yang and Jiang, Bo and He, Tong and Zhang, Qian and Long, Xiaoxiao and Yin, Wei},
  booktitle={Proceedings of the Computer Vision and Pattern Recognition Conference},
  pages={1602--1611},
  year={2025}
}

@article{liu2025takead,
  title={TakeAD: Preference-Based Post-Optimization for End-to-End Autonomous Driving With Expert Takeover Data},
  author={Liu, Deqing and Gao, Yinfeng and Qian, Deheng and Zhang, Qichao and Ye, Xiaoqing and Han, Junyu and Zheng, Yupeng and Liu, Xueyi and Xia, Zhongpu and Ding, Dawei and others},
  journal={IEEE Robotics and Automation Letters},
  volume={11},
  number={2},
  pages={1738--1745},
  year={2025},
  publisher={IEEE}
}

@article{wang2026meanfuser,
  title={MeanFuser: Fast One-Step Multi-Modal Trajectory Generation and Adaptive Reconstruction via MeanFlow for End-to-End Autonomous Driving},
  author={Wang, Junli and Liu, Xueyi and Zheng, Yinan and Xing, Zebing and Li, Pengfei and Li, Guang and Ma, Kun and Chen, Guang and Ye, Hangjun and Xia, Zhongpu and others},
  journal={arXiv preprint arXiv:2602.20060},
  year={2026}
}

@article{liu2025bridgedrive,
  title={BridgeDrive: Diffusion Bridge Policy for Closed-Loop Trajectory Planning in Autonomous Driving},
  author={Liu, Shu and Chen, Wenlin and Li, Weihao and Wang, Zheng and Yang, Lijin and Huang, Jianing and Zhang, Yipin and Huang, Zhongzhan and Cheng, Ze and Yang, Hao},
  journal={arXiv preprint arXiv:2509.23589},
  year={2025}
}

@article{dosovitskiy2020ViT,
  title={An image is worth 16x16 words: Transformers for image recognition at scale},
  author={Dosovitskiy, Alexey and Beyer, Lucas and Kolesnikov, Alexander and Weissenborn, Dirk and Zhai, Xiaohua and Unterthiner, Thomas and Dehghani, Mostafa and Minderer, Matthias and Heigold, Georg and Gelly, Sylvain and others},
  journal={arXiv preprint arXiv:2010.11929},
  year={2020}
}

@article{oquab2023dinov2,
  title={Dinov2: Learning robust visual features without supervision},
  author={Oquab, Maxime and Darcet, Timoth{\'e}e and Moutakanni, Th{\'e}o and Vo, Huy and Szafraniec, Marc and Khalidov, Vasil and Fernandez, Pierre and Haziza, Daniel and Massa, Francisco and El-Nouby, Alaaeldin and others},
  journal={arXiv preprint arXiv:2304.07193},
  year={2023}
}

@inproceedings{chen2024internvl,
  title={Internvl: Scaling up vision foundation models and aligning for generic visual-linguistic tasks},
  author={Chen, Zhe and Wu, Jiannan and Wang, Wenhai and Su, Weijie and Chen, Guo and Xing, Sen and Zhong, Muyan and Zhang, Qinglong and Zhu, Xizhou and Lu, Lewei and others},
  booktitle={Proceedings of the IEEE/CVF conference on computer vision and pattern recognition},
  pages={24185--24198},
  year={2024}
}

@inproceedings{esser2021vqgan,
  title={Taming transformers for high-resolution image synthesis},
  author={Esser, Patrick and Rombach, Robin and Ommer, Bjorn},
  booktitle={Proceedings of the IEEE/CVF conference on computer vision and pattern recognition},
  pages={12873--12883},
  year={2021}
}

@article{wang2026emu3,
  title={Multimodal learning with next-token prediction for large multimodal models},
  author={Wang, Xinlong and Cui, Yufeng and Wang, Jinsheng and Zhang, Fan and Wang, Yueze and Zhang, Xiaosong and Luo, Zhengxiong and Sun, Quan and Li, Zhen and Wang, Yuqi and others},
  journal={Nature},
  pages={1--7},
  year={2026},
  publisher={Nature Publishing Group UK London}
}

@article{park2025nuplanqa,
  title={NuPlanQA: A Large-Scale Dataset and Benchmark for Multi-View Driving Scene Understanding in Multi-Modal Large Language Models},
  author={Park, Sung-Yeon and Cui, Can and Ma, Yunsheng and Moradipari, Ahmadreza and Gupta, Rohit and Han, Kyungtae and Wang, Ziran},
  journal={arXiv preprint arXiv:2503.12772},
  year={2025}
}

@inproceedings{wu2025language,
  title={Language prompt for autonomous driving},
  author={Wu, Dongming and Han, Wencheng and Liu, Yingfei and Wang, Tiancai and Xu, Cheng-zhong and Zhang, Xiangyu and Shen, Jianbing},
  booktitle={Proceedings of the AAAI Conference on Artificial Intelligence},
  volume={39},
  number={8},
  pages={8359--8367},
  year={2025}
}

@inproceedings{qian2024nuscenes-qa,
  title={Nuscenes-qa: A multi-modal visual question answering benchmark for autonomous driving scenario},
  author={Qian, Tianwen and Chen, Jingjing and Zhuo, Linhai and Jiao, Yang and Jiang, Yu-Gang},
  booktitle={Proceedings of the AAAI Conference on Artificial Intelligence},
  volume={38},
  number={5},
  pages={4542--4550},
  year={2024}
}

@inproceedings{nie2024reason2drive,
  title={Reason2drive: Towards interpretable and chain-based reasoning for autonomous driving},
  author={Nie, Ming and Peng, Renyuan and Wang, Chunwei and Cai, Xinyue and Han, Jianhua and Xu, Hang and Zhang, Li},
  booktitle={European Conference on Computer Vision},
  pages={292--308},
  year={2024},
  organization={Springer}
}

@inproceedings{sima2024drivelm,
  title={Drivelm: Driving with graph visual question answering},
  author={Sima, Chonghao and Renz, Katrin and Chitta, Kashyap and Chen, Li and Zhang, Hanxue and Xie, Chengen and Bei{\ss}wenger, Jens and Luo, Ping and Geiger, Andreas and Li, Hongyang},
  booktitle={European Conference on Computer Vision},
  pages={256--274},
  year={2024},
  organization={Springer}
}

@inproceedings{vitelli2022safetynet,
  title={Safetynet: Safe planning for real-world self-driving vehicles using machine-learned policies},
  author={Vitelli, Matt and Chang, Yan and Ye, Yawei and Ferreira, Ana and Wo{\l}czyk, Maciej and Osi{\'n}ski, B{\l}a{\.z}ej and Niendorf, Moritz and Grimmett, Hugo and Huang, Qiangui and Jain, Ashesh and others},
  booktitle={2022 International Conference on Robotics and Automation (ICRA)},
  pages={897--904},
  year={2022},
  organization={IEEE}
}

@article{bansal2018chauffeurnet,
  title={Chauffeurnet: Learning to drive by imitating the best and synthesizing the worst},
  author={Bansal, Mayank and Krizhevsky, Alex and Ogale, Abhijit},
  journal={arXiv preprint arXiv:1812.03079},
  year={2018}
}

@inproceedings{chen2020learning,
  title={Learning by cheating},
  author={Chen, Dian and Zhou, Brady and Koltun, Vladlen and Kr{\"a}henb{\"u}hl, Philipp},
  booktitle={Conference on robot learning},
  pages={66--75},
  year={2020},
  organization={PMLR}
}

@inproceedings{sauer2018conditional,
  title={Conditional affordance learning for driving in urban environments},
  author={Sauer, Axel and Savinov, Nikolay and Geiger, Andreas},
  booktitle={Conference on robot learning},
  pages={237--252},
  year={2018},
  organization={PMLR}
}

@inproceedings{carion2020end,
  title={End-to-end object detection with transformers},
  author={Carion, Nicolas and Massa, Francisco and Synnaeve, Gabriel and Usunier, Nicolas and Kirillov, Alexander and Zagoruyko, Sergey},
  booktitle={European conference on computer vision},
  pages={213--229},
  year={2020},
  organization={Springer}
}

@article{fang2024eva,
  title={Eva-02: A visual representation for neon genesis},
  author={Fang, Yuxin and Sun, Quan and Wang, Xinggang and Huang, Tiejun and Wang, Xinlong and Cao, Yue},
  journal={Image and Vision Computing},
  volume={149},
  pages={105171},
  year={2024},
  publisher={Elsevier}
}

@article{zheng2023judging,
  title={Judging llm-as-a-judge with mt-bench and chatbot arena},
  author={Zheng, Lianmin and Chiang, Wei-Lin and Sheng, Ying and Zhuang, Siyuan and Wu, Zhanghao and Zhuang, Yonghao and Lin, Zi and Li, Zhuohan and Li, Dacheng and Xing, Eric and others},
  journal={Advances in neural information processing systems},
  volume={36},
  pages={46595--46623},
  year={2023}
}

@inproceedings{bhattacharyya2024ssl,
  title={SSL-interactions: Pretext tasks for interactive trajectory prediction},
  author={Bhattacharyya, Prarthana and Huang, Chengjie and Czarnecki, Krzysztof},
  booktitle={2024 IEEE Intelligent Vehicles Symposium (IV)},
  pages={1450--1457},
  year={2024},
  organization={IEEE}
}

@article{sanh2019distilbert,
  title={DistilBERT, a distilled version of BERT: smaller, faster, cheaper and lighter},
  author={Sanh, Victor and Debut, Lysandre and Chaumond, Julien and Wolf, Thomas},
  journal={arXiv preprint arXiv:1910.01108},
  year={2019}
}

@article{dwivedi2020graphtransformer,
  title={A generalization of transformer networks to graphs},
  author={Dwivedi, Vijay Prakash and Bresson, Xavier},
  journal={arXiv preprint arXiv:2012.09699},
  year={2020}
}

@article{ding2024hintad,
  title={Hint-ad: Holistically aligned interpretability in end-to-end autonomous driving},
  author={Ding, Kairui and Chen, Boyuan and Su, Yuchen and Gao, Huan-ang and Jin, Bu and Sima, Chonghao and Zhang, Wuqiang and Li, Xiaohui and Barsch, Paul and Li, Hongyang and others},
  journal={arXiv preprint arXiv:2409.06702},
  year={2024}
}

@inproceedings{zheng2024large,
  title={Large language models powered context-aware motion prediction in autonomous driving},
  author={Zheng, Xiaoji and Wu, Lixiu and Yan, Zhijie and Tang, Yuanrong and Zhao, Hao and Zhong, Chen and Chen, Bokui and Gong, Jiangtao},
  booktitle={2024 IEEE/RSJ International Conference on Intelligent Robots and Systems (IROS)},
  pages={980--985},
  year={2024},
  organization={IEEE}
}

@inproceedings{tian2023unsupervised,
  title={Unsupervised road anomaly detection with language anchors},
  author={Tian, Beiwen and Liu, Mingdao and Gao, Huan-ang and Li, Pengfei and Zhao, Hao and Zhou, Guyue},
  booktitle={2023 IEEE international conference on robotics and automation (ICRA)},
  pages={7778--7785},
  year={2023},
  organization={IEEE}
}

@article{guo2026surds,
  title={Surds: Benchmarking spatial understanding and reasoning in driving scenarios with vision language models},
  author={Guo, Xianda and Zhang, Ruijun and Duan, Yiqun and He, Yuhang and Nie, Dujun and Huang, Wenke and Zhang, Chenming and Liu, Shuai and Zhao, Hao and Chen, Long},
  journal={Advances in Neural Information Processing Systems},
  volume={38},
  year={2026}
}

@inproceedings{zhang2025chameleon,
  title={Chameleon: Fast-slow neuro-symbolic lane topology extraction},
  author={Zhang, Zongzheng and Li, Xinrun and Zou, Sizhe and Chi, Guoxuan and Li, Siqi and Qiu, Xuchong and Wang, Guoliang and Zheng, Guantian and Wang, Leichen and Zhao, Hang and others},
  booktitle={2025 IEEE International Conference on Robotics and Automation (ICRA)},
  pages={3752--3758},
  year={2025},
  organization={IEEE}
}

@article{gao2026uniuncer,
  title={UniUncer: Unified Dynamic Static Uncertainty for End to End Driving},
  author={Gao, Yu and Wang, Jijun and Zhang, Zongzheng and Jiang, Anqing and Wang, Yiru and Heng, Yuwen and Wang, Shuo and Sun, Hao and Hu, Zhangfeng and Zhao, Hao},
  journal={arXiv preprint arXiv:2603.07686},
  year={2026}
}

@article{zhang2026umpe,
  title={Unified Map Prior Encoder for Mapping and Planning},
  author={Zhang, Zongzheng and Zou, Sizhe and Zheng, Guantian and Zhu, Zhenxin and Gao, Yu and Chi, Guoxuan and Wang, Shuo and Heng, Yuwen and Sun, Zhigang and Wang, Yiru and others},
  journal={arXiv preprint arXiv:2605.02762},
  year={2026}
}

@inproceedings{zhang2025delving,
  title={Delving into Mapping Uncertainty for Mapless Trajectory Prediction},
  author={Zhang, Zongzheng and Qiu, Xuchong and Zhang, Boran and Zheng, Guantian and Gu, Xunjiang and Chi, Guoxuan and Gao, Huan-ang and Wang, Leichen and Liu, Ziming and Li, Xinrun and others},
  booktitle={2025 IEEE/RSJ International Conference on Intelligent Robots and Systems (IROS)},
  pages={16969--16976},
  year={2025},
  organization={IEEE}
}

\clearpage
\appendix

\setcounter{figure}{0}
\setcounter{equation}{0}
\setcounter{table}{0}
\renewcommand{\thefigure}{A.\arabic{figure}}
\renewcommand{\theequation}{A.\arabic{equation}}
\renewcommand{\thetable}{A.\arabic{table}}

\section*{Appendix}
\label{sec:app}

This appendix provides supplementary technical details, extended discussions, and additional experimental evidence to support the main findings of the paper. The content is organized as follows:

\begin{itemize} 
\item \textbf{App.~\ref{apd:extended related work}: Extended Related Work.} We provide a more comprehensive review of representative end-to-end autonomous driving paradigms, as well as prior efforts that use scene understanding to improve downstream planning.

\item \textbf{App.~\ref{apd:3D dataset construction details}: 3D Dataset Construction Details.} We describe the full pipeline for constructing 3D traffic elements across different datasets. We also detail the prediction and processing of topological relationships, concluding with an analysis of potential failure cases during the construction process.

\item \textbf{App.~\ref{apd: datasets and metric details}: Datasets \& Metrics Details.} We introduce the four evaluation benchmarks in detail and summarize the definitions, computation, and interpretation of the metrics used in our experiments.

\item \textbf{App.~\ref{apd:model details}: Model Details.} We present architecture-specific integration details, training protocols, and implementation settings for applying our method to different end-to-end planning backbones.

\item \textbf{App.~\ref{apd:additional visualization}: Additional Visualization.} We provide more qualitative examples and analysis to illustrate the scenarios in which our method is more beneficial.

\item \textbf{App.~\ref{apd:ablation study details}: Ablation Study Details.} We detail the structural configurations, modifications, and exact implementation specifics of the comparative variants utilized in our ablation experiments.

\item \textbf{App.~\ref{apd:addtional nuscens metrics}: Limitations of nuScenes Metrics.} We discuss the limitations of the standard nuScenes~\cite{caesar2020nuscenes} evaluation protocol and report supplementary metrics for a more comprehensive assessment.

\item \textbf{App.~\ref{apd:limitations & future work}: Limitations \& Future Work:} We discuss current system constraints and outlines directions for future research.
\end{itemize}

\section{Extended Related Work}
\label{apd:extended related work}
\subsection{End-to-End Autonomous Driving}
Recent studies on end-to-end autonomous driving have explored several paradigms for integrating perception, reasoning, and planning within a unified framework. Early studies on end-to-end autonomous driving mainly follow either a perception-planning paradigm or regression-based formulations. Regression-based methods directly map sensor observations to driving actions or trajectories. For example, TCP~\cite{wu2022trajectory} proposes a simple baseline that predicts control commands with trajectory supervision. TransFuser~\cite{chitta2022transfuser} introduces transformer-based sensor fusion for imitation learning, while BEV-Planner~\cite{li2024bevplanner} predicts driving trajectories from BEV representations using ego status as key inputs. The perception-planning paradigm learns structured scene representations before performing planning. ST-P3~\cite{hu2022stp3} learns spatial-temporal scene representations for joint perception and planning. UniAD~\cite{hu2023planning} proposes a planning-oriented framework that unifies perception, prediction, and planning tasks. VAD~\cite{jiang2023vad} introduces vectorized scene representations for efficient planning, while SparseDrive~\cite{sun2025sparsedrive} adopts sparse scene representations for scalable end-to-end driving. VADv2~\cite{chen2024vadv2} further extends this paradigm with probabilistic planning for multimodal decision-making.

Recent studies have explored generative planning to capture the inherent multi-modality of driving behaviors by modeling a distribution over future trajectories rather than predicting a single deterministic plan. GenAD~\cite{zheng2024genad} adopts variational autoencoder (VAE) formulations to model trajectory uncertainty through latent variables. Methods such as DiffusionDrive~\cite{liao2025diffusiondrive}, DiffusionDriveV2~\cite{zou2025diffusiondrivev2}, BridgeDrive~\cite{liu2025bridgedrive}, and DIVER~\cite{song2025breaking} formulate trajectory generation as a diffusion process that iteratively refines noisy trajectory samples toward feasible driving behaviors. Approaches including GoalFlow~\cite{xing2025goalflow}, GuideFlow~\cite{liu2025guideflow}, and MeanFuser~\cite{wang2026meanfuser} investigate flow-based generative models for more efficient trajectory synthesis. These generative approaches provide strong expressiveness for modeling complex motion distributions and diverse driving behaviors.

Another line of work formulates end-to-end driving as a trajectory scoring problem, where the model evaluates multiple candidate trajectories and selects the most suitable one. Hydra-MDP~\cite{li2024hydra} introduces a multi-target hydra distillation framework for multimodal trajectory evaluation. DriveSuprim~\cite{yao2025drivesuprim} improves trajectory selection by learning a more precise scoring mechanism. GTRS~\cite{li2025generalized} further studies a unified scoring formulation for multimodal planning. Centaur~\cite{sima2025centaur} enhances robustness of scoring-based planners through test-time training under distribution shifts. ZTRS~\cite{li2025ztrs} explores trajectory scoring without imitation learning by directly optimizing policies from environment feedback.

More recently, researchers have begun to integrate large-scale foundation models into autonomous driving through vision-language models (VLMs) and vision-language-action (VLA) frameworks, enabling semantic reasoning and language grounding for driving decisions. Works such as DrivingGPT~\cite{chen2025drivinggpt}, Senna~\cite{jiang2024senna}, and OmniDrive~\cite{wang2025omnidrive} introduce VLMs into autonomous driving to enhance scene interpretation and reasoning. Other works focus on end-to-end VLA driving architectures that directly map multimodal observations and language inputs to driving behaviors. EMMA~\cite{hwang2024emma} and UniVLA~\cite{wang2025unified} propose unified multimodal frameworks that integrate perception, language grounding, and driving policy learning. AutoVLA~\cite{zhou2025autovla} enhances this paradigm through adaptive reasoning and reinforcement fine-tuning, while DriveVLA-W0~\cite{li2025drivevla} investigates the scaling properties of VLA driving systems. Impromptu-VLA~\cite{chi2025impromptu} promotes open-weight driving foundation models trained with large-scale multimodal data. DiffVLA~\cite{jiang2025diffvla} and LLM-augmented motion prediction~\cite{zheng2024large}
incorporate language-based reasoning into motion prediction and planning, while Orion~\cite{fu2025orion} explores instruction-conditioned driving policies within a unified end-to-end framework.

Researchers have also explored learning-based world models to capture environment dynamics and improve long-horizon decision-making in autonomous driving. SSR~\cite{li2024ssr} introduces navigation-guided sparse scene representations to model structured driving environments for end-to-end planning. LAW~\cite{li2024law} incorporates a latent world model to capture temporal dependencies and environment dynamics. World4Drive~\cite{zheng2025world4drive} further proposes an intention-aware physical latent world model that explicitly models interactions between agents and the environment. DriveX~\cite{shi2025drivex} extends this idea by learning omni-scene representations to capture generalizable world knowledge for autonomous driving.

To validate the effectiveness of our approach, we select representative methods from different end-to-end driving paradigms and augment them with traffic element information, aiming to enhance their planning performance.

\subsection{Traffic-Aware Scene Understanding}
Prior work has injected rule-critical scene cues into driving in several different ways. 
Conditional Affordance Learning (CAL)~\cite{sauer2018conditional} predicts compact driving affordances, including traffic-light and sign-related signals, and uses them to guide downstream control. Learning by Cheating~\cite{chen2020learning} transfers privileged bird's-eye-view supervision, allowing the policy to exploit traffic-rule-relevant scene structure during training. ChauffeurNet~\cite{bansal2018chauffeurnet} rasterizes roadmap and traffic-light states as planner inputs, so that rule information is injected through a top-down representation. SafetyNet~\cite{vitelli2022safetynet} instead places traffic-rule reasoning after the learned planner, using a rule-based fallback layer to override unsafe outputs when necessary. More recently, VADv2~\cite{chen2024vadv2} introduces explicit traffic-element tokens for traffic lights and stop signs, supervising both signal state and whether the signal affects the ego vehicle. These works confirm that traffic lights/signs are decision-critical, but they are typically explored either as affordances, privileged/raster cues, post-hoc safety constraints, or within a single planner architecture.

A second related direction focuses on topology and traffic regulation as structured scene understanding problems. OpenLane-V2~\cite{wang2023openlane} is the first benchmark to jointly annotate lanes, traffic elements, and their relations, explicitly arguing that lane--traffic associations facilitate downstream decision-making. Follow-up methods such as TopoNet~\cite{li2023toponet}, TopoMLP~\cite{wu2023topomlp}, LaneSegNet~\cite{li2023lanesegnet}, T2SG~\cite{lv2025t2sg} improve lane--lane and lane--traffic reasoning, while MapDR~\cite{chang2025mapdr} further emphasizes the traffic regulation layer by associating traffic-sign rules with vectorized HD maps. However, these methods are primarily evaluated on topology or online-mapping benchmarks, with limited validation of how such cues transfer to planning. 

Another route is to use heavy VLM/VQA-style scene understanding for driving. Methods such as Orion~\cite{fu2025orion}, OmniDrive~\cite{wang2025omnidrive}, and AutoVLA~\cite{zhou2025autovla} reason about traffic lights, signs, and other rule-relevant semantics in language space, often enabled by dedicated driving VQA datasets~\cite{park2025nuplanqa, wu2025language, qian2024nuscenes-qa, nie2024reason2drive, sima2024drivelm, guo2026surds, tian2023unsupervised}, but they typically incur substantially higher runtime cost. In contrast, our work targets the same rule-critical cues with a lightweight alternative based on explicit 3D traffic elements and ego-centric lane topology. More importantly, we systematically evaluate how these cues transfer to planning across multiple representative end-to-end backbones and both open-loop and closed-loop benchmarks, which is largely missing in prior traffic-aware and topology-aware studies. 

\textbf{Scope of traffic elements.} 
In this work, traffic elements refer specifically to regulatory traffic lights and traffic signs, following the definition used in OpenLane-V2. We do not include vehicles or pedestrians in this category, since they are already modeled as dynamic agents in most end-to-end driving systems. Our focus is instead on the rule-critical but relatively underexplored regulatory cues that directly constrain legal driving maneuvers, such as stop/go decisions, turning permissions, and lane-level access constraints. Recent work has also explored map- and lane-aware representations for trajectory prediction. For example, SSL-Interactions~\cite{bhattacharyya2024ssl} uses HD-map and lane context together with self-supervised pretext tasks to improve agent--agent interaction modeling. This line of work is complementary to ours. Rather than proposing a new lane-topology representation itself, our goal is to evaluate whether regulatory traffic elements and ego-relevant topology provide a general downstream planning signal across end-to-end driving planners and benchmarks. In particular, our experiments focus on how these rule-critical cues affect planning performance, rather than only measuring their standalone perception or topology-prediction quality.

\section{3D Dataset Construction Details}
\label{apd:3D dataset construction details}
This section details how we construct the two key supervision signals used in this work when they are not directly available from existing benchmarks: the 3D spatial coordinates of traffic elements and the ego-relevant topology. Specifically, we describe how 2D traffic elements are detected and lifted into 3D space, how depth and LiDAR cues are combined to recover reliable TE center points, and how ego-related lane--traffic, lane--lane topology is obtained across different datasets.

\subsection{3D Traffic Element Construction}

\subsubsection{2D Traffic Element Detection.}
The 3D location of each traffic element is derived from 2D detection and depth estimation.
Given an input image $I \in \mathbb{R}^{H \times W \times 3}$, a 2D detector~\cite{redmon2016yolo} predicts a set of bounding boxes $B=\{b_i\}_{i=1}^{N}$. Each bounding box is defined as \(b_i=(x_i^{\text{min}}, y_i^{\text{min}}, x_i^{\text{max}}, y_i^{\text{max}})\).
The center of the $i$-th bounding box is computed as:

\begin{equation}
(u_i, v_i) =
\left(
\frac{x_i^\text{{min}} + x_i^\text{{max}}}{2},
\frac{y_i^\text{{min}} + y_i^\text{{max}}}{2}
\right),
\end{equation}

\noindent where $(u_i, v_i)$ denotes the pixel coordinate of the traffic element center.

To construct reliable 3D traffic elements on datasets without TE annotations, we first build a strong 2D detector on OpenLane-V2 and then use it to generate pseudo labels for downstream 3D TE extraction. Starting from a YOLO-style detector, we progressively strengthen the model with a sequence of training and inference refinements, including selective augmentation, class-balanced resampling, foreground reweighting, pseudo-label bootstrapping, test-time augmentation, and finally a stronger detector backbone. As summarized in Fig.~\ref{fig:detection_ablation}, this stepwise optimization substantially improves TE detection quality and yields a detector robust enough for automatic annotation on datasets such as NAVSIM. The progressive enhancements applied are detailed as follows:

\begin{itemize}
    \item \textbf{Selective Strong Augmentation:} While aggressive data augmentation is standard practice for robust 2D object detection, we selectively tailor these techniques. We adopt spatial mixing strategies, specifically MixUp and Mosaic augmentations, inspired by the YOLOX paradigm~\cite{ge2021yolox}. However, we strictly prohibit the use of color gamut augmentation and horizontal flipping. Modifying the HSV color space severely impairs the model's capacity to correctly recognize the semantic states of traffic lights (e.g., distinguishing between red and green). Similarly, applying horizontal flips reverse the semantic meaning of directional traffic signs, leading to critical misclassifications~\cite{wu20231sttopology}.
    
    \item \textbf{Class-Balanced Resampling:} The natural distribution of traffic elements in driving datasets exhibits a severe long-tail phenomenon. Statistical analysis reveals that the "unknown" state of traffic lights constitutes nearly 50\% of the annotations in the frontal view. However, these "unknown" instances provide negligible actionable information for the downstream motion planning task. Conversely, critical directive traffic signs often account for an exceptionally small fraction of the overall data. To mitigate this extreme class imbalance, we implement a resampling strategy that down-samples the over-represented, uninformative classes while aggressively over-sampling the rare but critical traffic signs, thereby synthesizing a uniformly distributed training manifold.
    
    \item \textbf{Foreground Loss Reweighting:} A major source of error is not coarse localization, but fine-grained category confusion among visually similar traffic signs, such as \texttt{turn\_left}, \texttt{no\_left\_turn}, and \texttt{slight\_left}. Since these categories are both infrequent and semantically important, we place extra emphasis on their classification by increasing the foreground classification weight. This encourages the detector to allocate more capacity to difficult sign recognition rather than being dominated by easier or more frequent classes. In our implementation, the foreground classification term is upweighted, while the localization branch remains unchanged.
    
    \item \textbf{Pseudo-Labeling for Sparse Annotations:} In typical driving logs, distant traffic elements are frequently omitted from ground-truth annotations when they initially enter the camera's field of view due to their diminutive pixel footprint. These missing annotations inherently act as noisy negative samples, which confuses the model and prevents optimal convergence during training. To rectify this, we leverage an intermediate, high-confidence detector to infer pseudo-labels across the unannotated frames. By identifying and labeling these distant objects, we provide the network with a denser, more consistent supervision signal, which significantly enhances subsequent training phases.
    
    \item \textbf{Test-Time Augmentation (TTA):} During the inference phase, we apply Test-Time Augmentation to systematically enhance detection robustness and stability. We strictly limit our TTA approach to multi-scale testing (utilizing scale factors ranging from 0.7x to 1.4x), as introducing more complex spatial transformations often degrades performance rather than improving it. Upscaling the input resolution specifically aids in recalling distant, small-scale traffic lights, whereas downscaling proves advantageous for successfully capturing large, ego-adjacent road markings.
    
    \item \textbf{Advanced Architecture Upgrade:} Finally, to push the representational capacity of our feature extractor to its upper limit, we transition from the baseline architecture to the state-of-the-art YOLO26~\cite{sapkota2025yolo26}. 
\end{itemize}

\subsubsection{Cross-dataset reliability of pseudo traffic-element labels.} 
Since NAVSIM does not provide native traffic-element annotations, our NAVSIM experiments rely on pseudo labels generated by the traffic-element detector trained on OpenLane-V2. To verify that these pseudo labels are reliable rather than artifacts of dataset bias, we manually annotate a held-out NAVSIM subset containing 100 frames sampled from 20 scenes, with 1,126 traffic-element instances in total. We then evaluate the OpenLane-V2-trained detector directly on this manually annotated NAVSIM subset without any dataset-specific fine-tuning. As shown in Tab.~\ref{tab:navsim_manual_te_eval}, the detector achieves strong cross-dataset performance, with 0.908 mAP$_{50}$, 0.725 mAP$_{50:95}$, 0.915 recall, and 0.918 precision. This suggests that the automatically constructed NAVSIM traffic-element labels provide meaningful supervision for downstream planning. The detector produces 92 false positives and 96 false negatives on this subset. Most errors come from small or distant traffic elements, which are also difficult to annotate consistently. Nevertheless, the high precision and recall indicate that the pseudo labels are sufficiently reliable for training traffic-element-aware planning models.
\begin{table}[t] 
\centering 
\caption{Cross-dataset validation of the OpenLane-V2-trained traffic-element detector on a manually annotated NAVSIM subset. The subset contains 100 frames from 20 scenes and 1,126 traffic-element instances.} \label{tab:navsim_manual_te_eval} 
\resizebox{0.78\linewidth}{!}
{ 
\begin{tabular}
{lcccccc} 
\toprule 
Dataset & Frames & TE Inst. & mAP$_{50}$ & mAP$_{50:95}$ & Recall & Precision \\ 
\midrule NAVSIM manual subset & 100 & 1,126 & 0.908 & 0.725 & 0.915 & 0.918 \\ 
\bottomrule \end{tabular} } 
\end{table}

\subsubsection{Depth Estimation and Fusion.}

A depth estimation network~\cite{piccinelli2025unidepthv2} predicts a dense depth map $D \in \mathbb{R}^{H \times W}$. The depth value corresponding to the \(i\)-th traffic element is obtained from the bounding box center as \(z_i^{\text{depth}} = D(u_i, v_i)\).
When LiDAR measurements are available, LiDAR points are projected onto the image plane using the calibration parameters between the LiDAR and camera sensors. 
The set of LiDAR points falling inside the bounding box $b_i$ is denoted as:
$\mathcal{P}_i = \{ \mathbf{p}_j^{\text{lidar}} \}.$

Since the LiDAR points inside the bounding box may correspond to multiple objects and therefore contain multiple depth values, 
the depth predicted by the monocular depth estimation network is used as a reference to select the LiDAR depth corresponding to the traffic element center.

Given the selected depth value $z_i$, the corresponding 3D point in the camera coordinate system is obtained by back-projecting the pixel coordinate:

\begin{equation}
    \mathbf{p}_i^{\text{cam}}
=
z_i K^{-1}
\begin{bmatrix}
u_i \\
v_i \\
1
\end{bmatrix},
\end{equation}

\noindent where $K$ denotes the camera intrinsic matrix.

Finally, the 3D center of the traffic element in the world coordinate system is obtained through a rigid transformation:

\begin{equation}
    \mathbf{p}_i^{\text{LiDAR}}
=
R_\text{cl}\mathbf{p}_i^{\text{cam}} + t_\text{cl},
\end{equation}
\noindent where $R_\text{cl} \in \mathbb{R}^{3\times3}$ and $t_\text{cl} \in \mathbb{R}^{3}$ denote the rotation matrix and translation vector between the camera and LiDAR coordinate systems.

\subsubsection{Failure Case.} 

\begin{figure}
    \centering
    \includegraphics[width=1.0\linewidth]{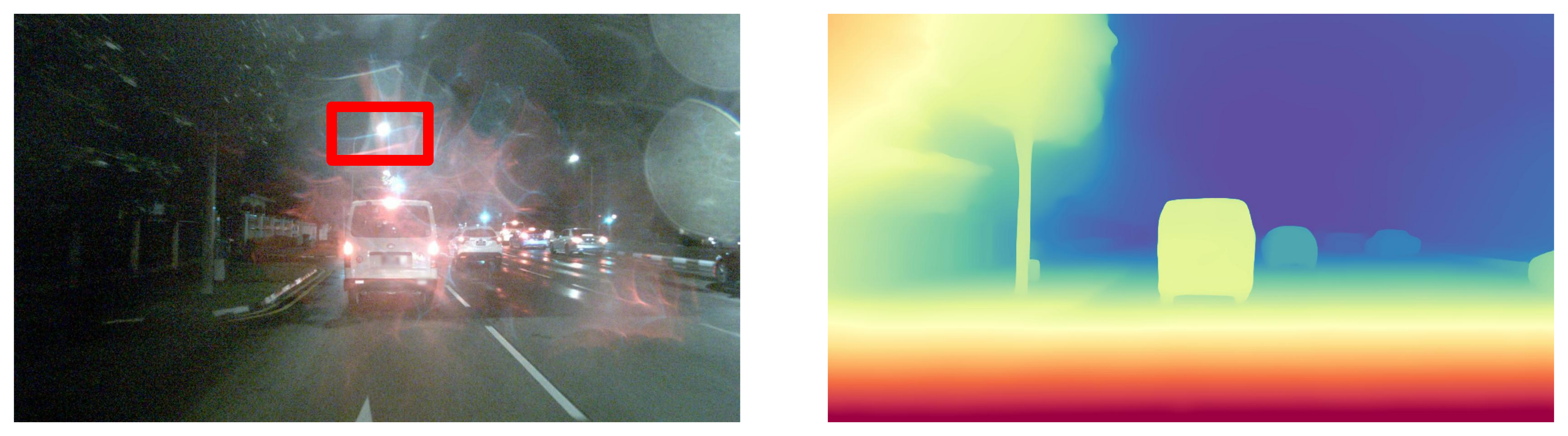}
    \caption{Failure case of depth estimation. The \textbf{left} image shows the front-view camera image, where the green traffic light is highlighted by the red bounding box. The \textbf{right} image shows the corresponding predicted depth map.}
    \label{fig:app-depth_failure}
\end{figure}

We present a failure case of the proposed pipeline. In challenging environmental conditions, the depth estimation network may produce unreliable predictions. In the example shown in Fig.~\ref{fig:app-depth_failure}, rain-induced image blur degrades the visual quality of the input image, causing the depth estimation network to incorrectly estimate the depth of the green traffic light indicated by the red bounding box. As a result, the reconstructed 3D position of the traffic element becomes inaccurate.

\begin{figure}
    \centering
    \includegraphics[width=1.0\linewidth]{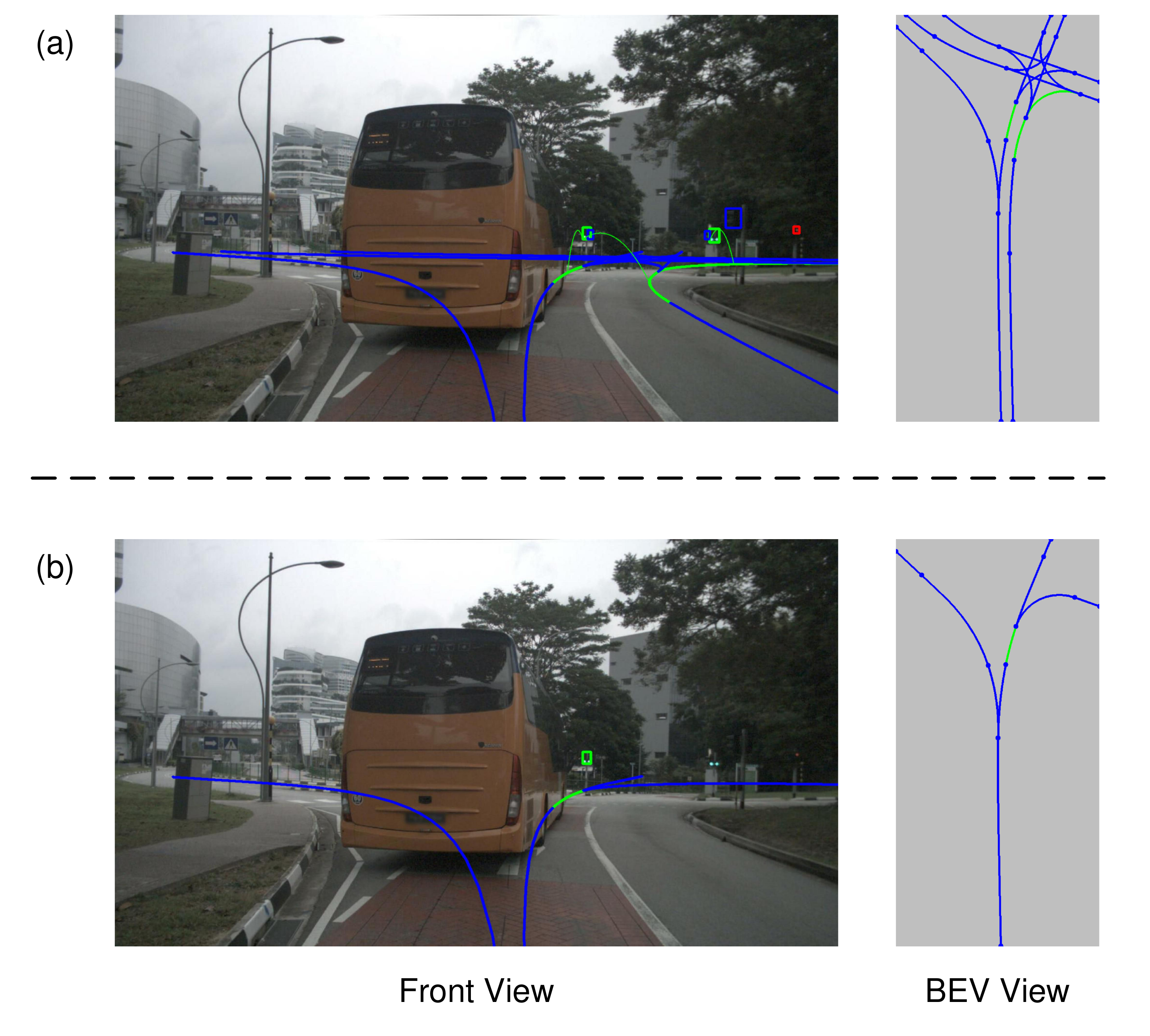}
    \caption{Illustration of topology extraction. Blue lines denote centerlines, while green lines indicate the topology relations between green traffic lights and the corresponding centerlines.}
    \label{fig:app-topo_extraction}
\end{figure}

\subsubsection{Depth quality under challenging lighting.}
We further analyze how depth-estimation quality changes under different lighting conditions and how this correlates with planning performance. We split the nuScenes validation samples into daytime and nighttime subsets and evaluate the VAD-based traffic-element model using standard depth metrics and planning metrics. As reported in Tab.~\ref{tab:day_night_depth_planning}, nighttime scenes have noticeably worse depth estimates than daytime scenes, with AbsRel increasing from 0.088 to 0.160 and RMSE$_{\log}$ increasing from 0.193 to 0.279. This degradation is accompanied by worse planning accuracy, where L2@3s increases from 0.88 to 1.33. These results confirm that challenging lighting can affect the quality of the reconstructed 3D traffic-element representation and therefore weaken downstream planning performance.

\begin{table}[t] \centering \caption{ Day/night breakdown on nuScenes with the VAD-based traffic-element model. Nighttime scenes exhibit worse depth quality, which correlates with degraded planning accuracy. } \label{tab:day_night_depth_planning} \resizebox{\linewidth}{!}{ \begin{tabular}{lccccccccc} \toprule \multirow{2}{*}{Subset} & \multicolumn{3}{c}{Depth} & \multicolumn{6}{c}{Planning} \\ \cmidrule(lr){2-4} \cmidrule(lr){5-10} & AbsRel$\downarrow$ & RMSE$_{\log}\downarrow$ & $\delta_1\uparrow$ & L2@1s$\downarrow$ & L2@2s$\downarrow$ & L2@3s$\downarrow$ & Col@1s$\downarrow$ & Col@2s$\downarrow$ & Col@3s$\downarrow$ \\ \midrule All & 0.095 & 0.201 & 0.920 & 0.36 & 0.61 & 0.92 & 0.09 & 0.14 & 0.28 \\ Night & 0.160 & 0.279 & 0.793 & 0.49 & 0.85 & 1.33 & 0.00 & 0.15 & 0.55 \\ Day & 0.088 & 0.193 & 0.934 & 0.34 & 0.58 & 0.88 & 0.10 & 0.13 & 0.25 \\ \bottomrule \end{tabular} } \end{table} 

This analysis also explains the failure case shown in Fig.~\ref{fig:depth_failure}. When image quality is degraded by adverse weather or low illumination, depth estimation can become unreliable and produce inaccurate 3D traffic-element locations. Our current pipeline is therefore robust to moderate noise, but still depends on the quality of upstream detection and depth estimation in extremely challenging visual conditions.

\subsection{Topology Prediction and Extraction}

\subsubsection{Prediction Network Architecture.}
For topology prediction, we adopt the \textbf{TopoMLP} architecture~\cite{wu2023topomlp}, a query-based \emph{first-detect-then-reason} pipeline for driving topology reasoning. Given multi-view images, TopoMLP first detects \textbf{3D lane centerlines} and \textbf{2D traffic elements} with two dedicated detection branches, and then predicts both \textbf{lane--lane} and \textbf{lane--traffic} topology using lightweight \textbf{MLP heads} applied to pairwise query embeddings. Concretely, the query features of candidate centerlines and traffic elements are encoded with positional information, concatenated in pairs, and classified by small MLPs to obtain the corresponding adjacency relations. We use~\cite{wu2023topomlp} in this work as an efficient topology provider, and then convert the predicted ego-relevant topology into the compact conditioning representation described in the main paper.

\subsubsection{Ego-Related Topology Extraction.}
Given the predicted or annotated lane centerlines and their topology relations, 
we extract a local topology subgraph centered around the ego vehicle. Let $\mathcal{L}=\{l_i\}$ denote the set of lane centerlines and $\mathcal{T}=\{t_j\}$ denote the set of traffic elements. We define the ego position in the local coordinate system as $\mathbf{p}_{\text{ego}}=(0, 0)$. 
The centerline closest to the ego vehicle is denoted as $l_{\text{ego}}$. 
Starting from $l_{\text{ego}}$, we traverse the centerline graph using the matrix $\mathbf{R}_{\text{LCLC}}$ to obtain the set of centerlines that are topologically connected to it, denoted as $\mathcal{L}_{\text{conn}}$. Next, using the matrix $\mathbf{R}_{\text{LCTE}}$, we collect the traffic elements that have topology relations with the centerlines in $\{l_{\text{ego}}\}\cup\mathcal{L}_{\text{conn}}$, 
which form the traffic element set $\mathcal{T}_{\text{topo}}$. The resulting topology subgraph therefore consists of the centerline set $\{l_{\text{ego}}\}\cup\mathcal{L}_{\text{conn}}$ and the associated traffic element set $\mathcal{T}_{\text{topo}}$. As illustrated in Fig.~\ref{fig:app-topo_extraction}, the topology extraction process converts the global topology graph shown in Fig.~\ref{fig:app-topo_extraction}(a) into a local topology subgraph centered around the ego vehicle, as depicted in Fig.~\ref{fig:app-topo_extraction}(b).

\subsubsection{Language Topology Description Examples.}

\vspace{0.4em}






\begin{tcolorbox}[
    enhanced,
    breakable,
    colback=gray!6,
    colframe=gray!60,
    boxrule=0.5pt,
    arc=2mm,
    left=1mm,
    right=1mm,
    top=1mm,
    bottom=1mm,
    title=Example 1,
    fonttitle=\bfseries,
    fontupper=\ttfamily\small
]
There are a `green` traffic\_light, a `green` traffic\_light ahead controlling the current ego-lane, which connects `1` `straight` centerline.

\begin{center}
\includegraphics[width=0.7\linewidth]{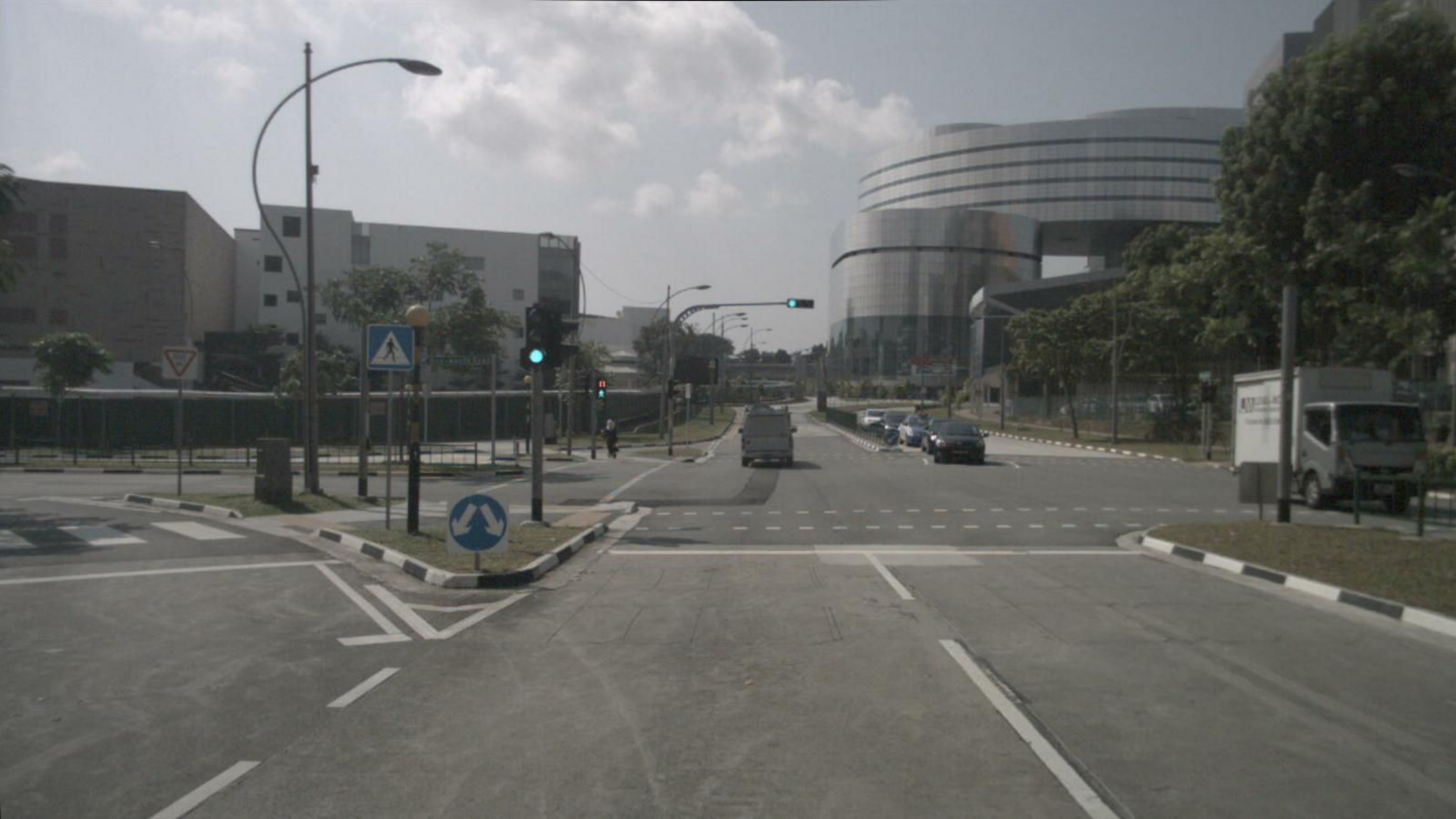}
\captionof{figure}{Illustration of the topology description in Example 1.}
\end{center}

\end{tcolorbox}



\vspace{0.6em}

\begin{tcolorbox}[
    enhanced,
    breakable,
    colback=gray!6,
    colframe=gray!60,
    boxrule=0.5pt,
    arc=2mm,
    left=1mm,
    right=1mm,
    top=1mm,
    bottom=1mm,
    title=Example 2,
    fonttitle=\bfseries,
    fontupper=\ttfamily\small
]
There are a `red` traffic\_light, a `red` traffic\_light, a `red` traffic\_light ahead controlling the current ego-lane, which connects `1` `straight` centerline.

\begin{center}
\includegraphics[width=0.7\linewidth]{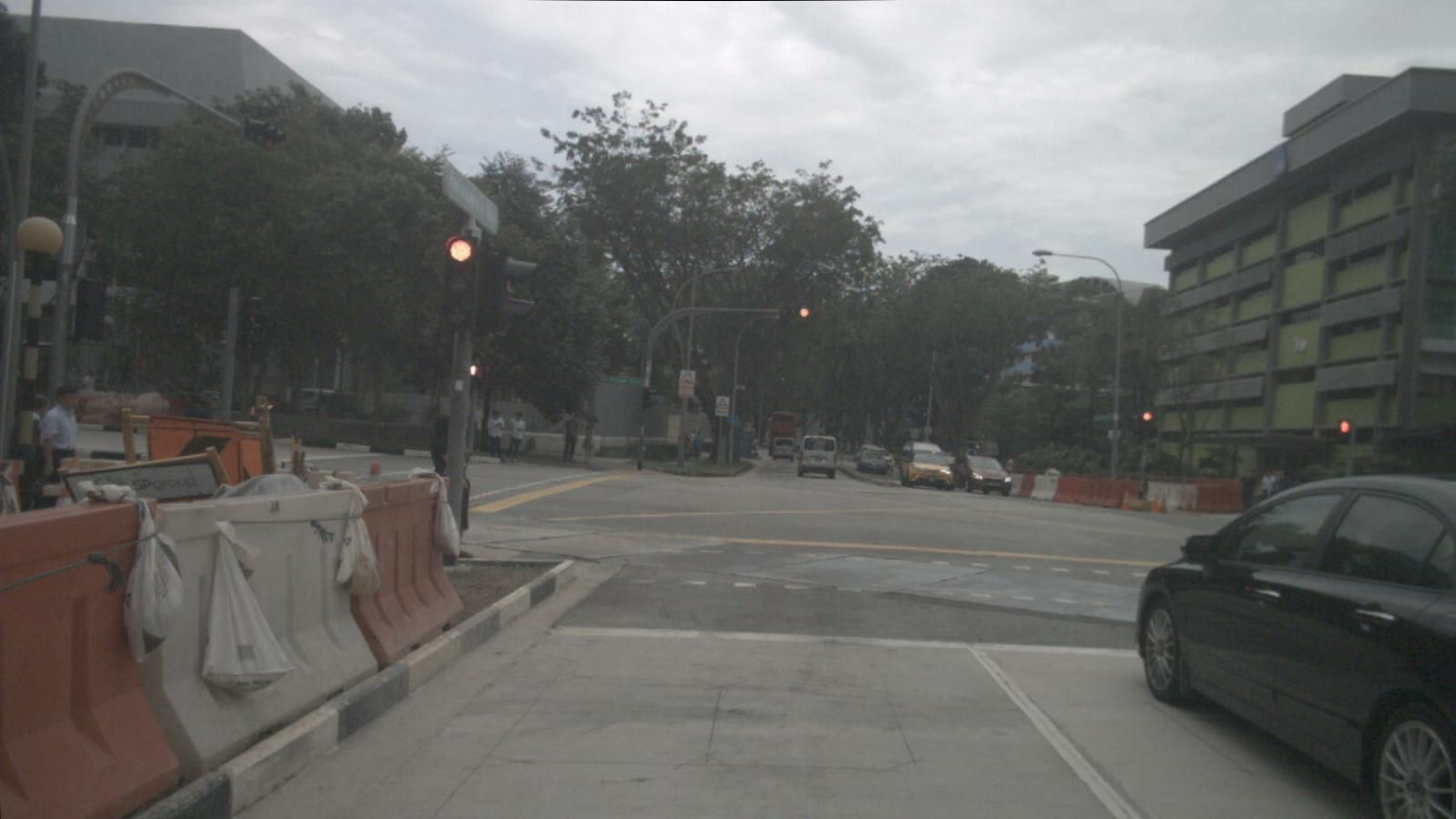}
\captionof{figure}{Illustration of the topology description in Example 2.}
\end{center}
\end{tcolorbox}

\vspace{0.4em}


\section{Datasets \& Metrics Details}
\label{apd: datasets and metric details}
\subsection{nuScenes}

The nuScenes~\cite{caesar2020nuscenes} dataset is a large-scale multimodal autonomous driving dataset containing 1, 000 driving scenes collected in urban environments. 
Each scene spans approximately 20 seconds and provides synchronized sensor data from multiple modalities, including six surround-view cameras, LiDAR, radar, and vehicle state measurements. 
For planning-oriented evaluation, the driving policy receives the current observation and predicts a future trajectory over a fixed horizon. 
The predicted trajectory is evaluated in an open-loop manner against the expert trajectory recorded by a human driver. 
Following prior works~\cite{hu2023planning,jiang2023vad}, we adopt the L2 trajectory error and collision rate as evaluation metrics.

Although the nuScenes HD map provides 3D annotations for traffic lights, their spatial locations are not sufficiently accurate. Therefore, we adopt 2D traffic element annotations from OpenLane-V2~\cite{wang2023openlane} to obtain 2D bounding boxes in the image plane. The corresponding 3D positions of traffic elements are obtained following the procedure described in App.~\ref{apd:3D dataset construction details}.






\subsection{NAVSIM-v1}

NAVSIM-v1~\cite{dauner2024navsim} evaluates sensor-based driving policies using non-reactive simulation on real-world data, containing approximately 120 hours of urban driving data recorded at 2 Hz. 
Each observation includes eight surround-view cameras with a resolution of $1920 \times 1080$ pixels and a fused LiDAR point cloud from five sensors. 
The driving policy receives the current frame and optionally several historical frames as input and predicts a future trajectory over a horizon of four seconds. 
The predicted trajectory is executed in a simplified bird's-eye-view simulation where surrounding agents replay their recorded trajectories while the ego vehicle is propagated using a kinematic bicycle model controlled at 10 Hz. 
Performance is measured using the Predictive Driver Model Score (PDMS):

\begin{equation}
\text{PDMS}=
\left(\prod_{m\in\{\text{NC,DAC}\}}\text{score}_m\right)
\left(
\frac{\sum_{{\omega} \in\{\text{EP,TTC,C}\}}w_{\omega}\cdot \text{score}_{\omega}}
{\sum_{{\omega} \in\{\text{EP,TTC,C}\}}w_{\omega}}
\right).
\end{equation}

The penalty terms include no collisions (NC) and  drivable area compliance (DAC), ensuring safety and map adherence, while the weighted metrics evaluate ego progress (EP), time-to-collision (TTC), and driving comfort (C). 

Since the NAVSIM map annotations only provide the state information of traffic elements without precise spatial locations, we construct the corresponding 3D pseudo-labels of traffic elements following the procedure described in App.~\ref{apd:3D dataset construction details}.

\subsection{NAVSIM-v2}

NAVSIM-v2~\cite{cao2025navsimv2} extends NAVSIM-v1 through a pseudo-simulation framework designed to better approximate closed-loop evaluation while maintaining scalability. 
The dataset is derived from the same large-scale driving logs but introduces additional synthetic observations generated using neural scene reconstruction techniques. 
Evaluation is conducted in two stages. 
In Stage 1, the policy predicts a trajectory from the real-world observation and the trajectory is simulated to obtain an initial score and endpoint. 
In Stage 2, multiple synthetic observations are generated around the predicted endpoint to approximate possible future states of the environment, and the policy is evaluated again on these observations to measure robustness. 
The resulting metric, called the Extended Predictive Driver Model Score (EPDMS), aggregates safety penalties and weighted driving performance metrics:

\begin{equation}
\mathrm{EPDMS}
=
\prod_{m \in \mathcal{M}_{\mathrm{pen}}}
\mathrm{filter}_m(\mathrm{agent}, \mathrm{human})
\cdot
\frac{
\sum_{m \in \mathcal{M}_{\mathrm{avg}}}
w_m \cdot \mathrm{filter}_m(\mathrm{agent}, \mathrm{human})
}{
\sum_{m \in \mathcal{M}_{\mathrm{avg}}} w_m
},
\end{equation}
where the penalty terms are
\[
\mathcal{M}_{\mathrm{pen}}=\{\mathrm{NC}, \mathrm{DAC}, \mathrm{DDC}, \mathrm{TLC}\},
\]
and the weighted-average terms are
\[
\mathcal{M}_{\mathrm{avg}}=\{\mathrm{TTC}, \mathrm{EP}, \mathrm{HC}, \mathrm{LK}, \mathrm{EC}\}.
\]
Here, \textbf{NC} denotes No at-fault Collision, \textbf{DAC} denotes Drivable Area Compliance, \textbf{DDC} denotes Driving Direction Compliance, and \textbf{TLC} denotes Traffic Light Compliance. The weighted-average terms include \textbf{EP} (Ego Progress), \textbf{TTC} (Time-to-Collision), \textbf{HC} (History Comfort), \textbf{LK} (Lane Keeping), and \textbf{EC} (Extended Comfort), with official weights $w_{\mathrm{EP}}=5$, $w_{\mathrm{TTC}}=5$, $w_{\mathrm{HC}}=2$, $w_{\mathrm{LK}}=2$, and $w_{\mathrm{EC}}=2$.

A distinctive component of EPDMS is the filtering function $\mathrm{filter}_m(\mathrm{agent},\mathrm{human})$. If the same rule violation is also committed by the human expert in the corresponding log, the associated penalty is ignored. This design reduces the sensitivity of the metric to annotation noise and to contextually justified maneuvers, such as temporarily entering the opposite lane to bypass a static obstacle. In the main paper, we report the overall EPDMS together with the Stage 1 / Stage 2 breakdown of the core subscores.

\subsection{Bench2Drive}
Bench2Drive~\cite{jia2024bench2drive} is a large-scale closed-loop benchmark built on CARLA for evaluating end-to-end driving under interactive scenarios. Its official training set contains 2 million fully annotated frames collected from 10,000 short clips (approximately 150 m each), covering 44 interactive scenarios, 23 weather conditions, and 12 towns. The sensor configuration is similar to nuScenes and includes 1 LiDAR, 6 RGB cameras, 5 radars, IMU/GNSS, and an HD map. Importantly for this work, the benchmark also provides HD-map lanes, centerlines, topology, dynamic traffic-light states, and trigger areas for traffic lights and stop signs.

Bench2Drive reports both open-loop and closed-loop metrics. Following the official protocol, we mainly use \textbf{Driving Score (DS)} and \textbf{Success Rate (SR)} as the primary closed-loop metrics, and additionally report \textbf{Efficiency}, \textbf{Comfortness}, and open-loop \textbf{Avg. L2}. Here, SR measures the proportion of routes completed without infractions, DS combines route completion with infraction penalties, Efficiency evaluates relative speed with respect to nearby traffic, and Comfortness follows the smoothness protocol based on accelerations, yaw dynamics, and jerk.

In this work, we directly use the benchmark's instance-level traffic-element annotations during training. These annotations provide each element's 3D location and semantic type; traffic lights additionally provide signal \texttt{state}, while lights and some signs also include \texttt{trigger\_volume} information. We further use the official \textbf{HD-map topology}, which contains lane centerlines, lane adjacency/topology, left/right lane relations, and trigger-volume structures for stop signs and traffic lights. In our pipeline, these annotations are used to supervise 3D traffic elements and ego-relevant topology during training.

\section{Model Details}
\label{apd:model details}
\subsection{Perception-Planning: VAD}

\begin{figure}
    \centering
    \includegraphics[width=1.0\linewidth]{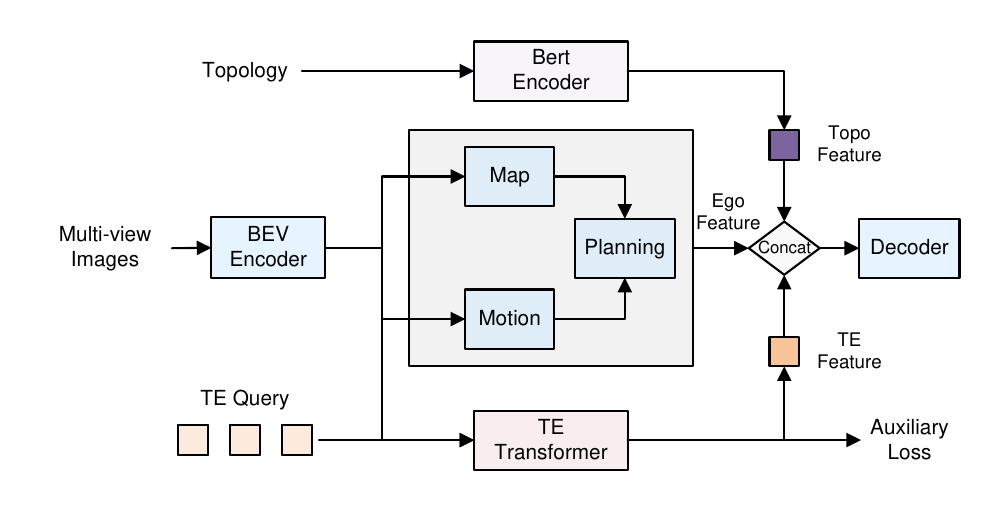}
    \caption{Overview of the VAD-based architecture with the proposed traffic element and topology branches.}
    \label{fig:app-VAD_ours}
\end{figure}

In this section, we describe how the proposed components are integrated into the VAD~\cite{jiang2023vad} framework. 
As illustrated in Fig.~\ref{fig:app-VAD_ours}, we follow the original Motion-Map-Planning architecture of VAD~\cite{jiang2023vad} and extend it with additional traffic element and topology modules. We introduce a traffic element (TE) branch to explicitly model traffic elements in the scene. Specifically, a TE Transformer is employed to predict traffic elements from the intermediate feature representations. The architecture of the TE Transformer is identical to the agent detection module used in VAD~\cite{jiang2023vad}. The predicted traffic elements are supervised using an additional auxiliary loss during training. The intermediate decoder features of the TE Transformer are extracted and used as the TE feature. To incorporate structural road information, we encode the topology graph using a BERT~\cite{devlin2019bert} encoder. The encoded topology representation is obtained from the final \text{[CLS]} token of the BERT output and is used as the topology feature. Finally, the ego feature produced by the VAD planning branch is concatenated with the traffic element feature and the topology feature along the channel dimension. 
The fused feature is then fed into the planning decoder to generate the final predicted trajectory.

For the training details, we initialize the model with the pretrained weights of VAD~\cite{jiang2023vad} and fine-tune it on the nuScenes dataset. 
The model is trained for 20 epochs using the AdamW optimizer with a learning rate of \(2\times10^{-5}\). In addition to the original training objectives, an extra supervision term for traffic elements is introduced. 
Specifically, the regression branch is optimized with the L1 loss with a weight of \(0.25\), while the classification branch adopts the Focal Loss with a weight of \(2.0\).

\subsection{LLM-Based: Orion}
We integrate our method into the VLM-based planner Orion~\cite{fu2025orion}, which follows a vision--reasoning--action pipeline. Orion first encodes multi-view images with a vision encoder and a query-based visual compressor (QT-Former / Q-Former), then uses an LLM to reason over the scene and produce a \emph{planning token}, which finally conditions a generative planner for trajectory prediction. In our nuScenes setting, we follow the official open-loop variant of Orion, which replaces the original QT-Former with the Q-Former from OmniDrive~\cite{wang2025omnidrive} and removes the explicit ego-status input in the generative planner. 

To incorporate \textbf{traffic elements}, we attach a lightweight TE prediction head to the perception queries produced by the visual compressor. This head predicts the 3D TE center and TE category, supervised by an additional \(L_1\) localization loss and focal classification loss, respectively. In this way, the perception queries are explicitly encouraged to encode rule-critical traffic cues such as traffic lights and traffic signs, rather than relying only on implicit visual reasoning.

To incorporate \textbf{topology}, we follow the language-centered design of Orion and convert the predicted ego-relevant topology into a compact structured text prompt, which is concatenated with the original textual input to the LLM. Concretely, the prompt summarizes the ego-lane-related centerlines, lane connectivity, and associated traffic elements. The LLM then reasons over both the visual queries and this topology-aware textual context to produce a more informative planning token. Compared with introducing a separate graph encoder, this design is more consistent with Orion's native reasoning space and allows topology and TE cues to be fused directly through language reasoning.

For training, we keep the official Orion settings unchanged and only add the TE supervision term. Specifically, we follow the original model configuration with EVA-02-L~\cite{fang2024eva} as the vision encoder and Vicuna v1.5~\cite{zheng2023judging} fine-tuned with LoRA (rank \(=16\), alpha \(=16\)); input images are resized to \(640\times640\), and the original multi-stage training schedule is preserved. The additional TE branch is optimized jointly with the original Orion objectives using the same \(L_1\) and focal losses as in our VAD setting. 

\subsection{Regression-Based: LTF}
\label{app:Regression-Based: LTF}
\begin{figure}
    \centering
    \includegraphics[width=1.0\linewidth]{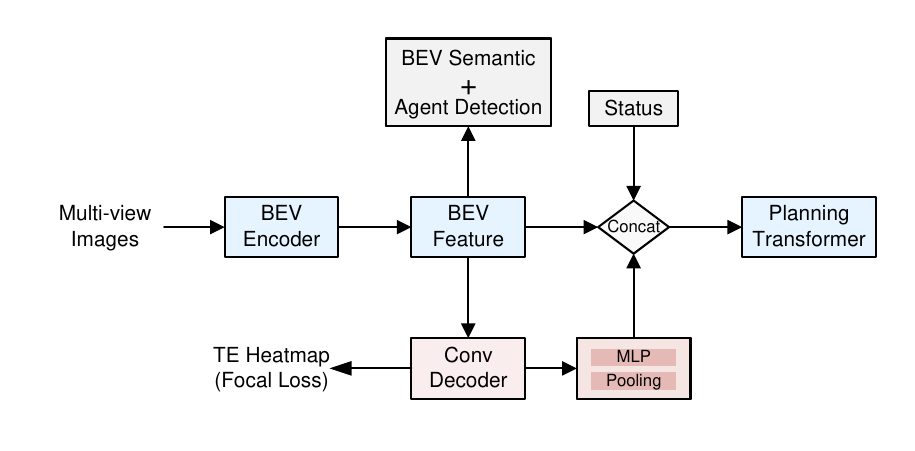}
    \caption{Overview of the LTF-based architecture with the proposed traffic element prediction branch.}
    \label{fig:app-LTF_ours}
\end{figure}

In this section, we describe how the proposed components are integrated into the LTF~\cite{chitta2022transfuser} baseline. As illustrated in Fig.~\ref{fig:app-LTF_ours}, we follow the original framework to generate BEV features from multi-view observations. In addition to the original auxiliary tasks in LTF~\cite{chitta2022transfuser}, including BEV segmentation and agent detection, we further introduce a traffic element prediction branch. Specifically, a convolution-based decoder is used to predict a traffic element heatmap from the intermediate features. The predicted heatmap represents the spatial distribution of traffic elements in the BEV space. To incorporate traffic element information into the planning module, the predicted heatmap is first pooled to match the spatial resolution of the BEV feature map. An MLP is then applied to adjust the channel dimension of the traffic element feature. The processed traffic element feature is concatenated with the BEV feature and the status feature along the channel dimension. Finally, the fused feature is fed into a Transformer-based decoder to generate the predicted future trajectory.

For the training details, we initialize the model with the pretrained weights of LTF~\cite{chitta2022transfuser} and fine-tune it on the navsim dataset. The model is trained for 20 epochs using the Adam optimizer with a learning rate of $1\times10^{-5}$. For traffic element prediction, the sparse center-point annotations on the BEV plane are converted into continuous supervision heatmaps using a two-dimensional Gaussian kernel with a radius of $r=2$. These heatmaps are used as the training targets for the traffic element prediction branch. The predicted heatmaps are supervised using the Focal Loss with a weight of \(1.0\).

\subsection{Diffusion-Based: DiffusionDrive}

In this section, we describe how the proposed components are integrated into the DiffusionDrive~\cite{liao2025diffusiondrive} baseline. Since DiffusionDrive~\cite{liao2025diffusiondrive} adopts the same BEV feature generation pipeline as LTF~\cite{chitta2022transfuser}, the traffic element branch is introduced in the same manner as described in the Sec.~\ref{app:Regression-Based: LTF}. The only difference lies in the trajectory prediction module, where the Transformer-based planning decoder is replaced by a diffusion-based planning decoder.

For the training details, we follow the design of LTF~\cite{chitta2022transfuser} and replace the LiDAR input with a learnable embedding. 
The model is first trained from scratch on the navsim dataset for 100 epochs using the AdamW optimizer with a learning rate of $6\times10^{-4}$. Based on the obtained weight, we further fine-tune the model for 20 epochs with a learning rate of $5\times10^{-5}$. The supervision loss and hyperparameter settings for traffic element prediction follow those described in Sec.~\ref{app:Regression-Based: LTF}.

\subsection{Scoring-Based: DrivoR}
We integrate our method into the scoring-based planner DrivoR~\cite{kirby2026drivoR}, which uses a \emph{perception encoder} to compress multi-camera features into compact scene tokens, followed by a \emph{trajectory decoder} that proposes candidate trajectories and a \emph{scoring decoder} that ranks them. 

In our implementation, we attach a lightweight \textbf{traffic-element (TE) prediction head} to the scene tokens produced by the perception encoder. Similar to the LTF setting, this head predicts a sparse BEV TE heatmap supervised by Gaussian-rendered TE centers with focal loss. The predicted TE feature is then spatially pooled and projected with an MLP, and the resulting TE embedding is concatenated to the scene-token memory. In this way, both the trajectory decoder and the downstream scoring decoder can attend to TE-aware scene tokens when generating and ranking candidate trajectories. Since our DrivoR experiments are conducted on NAVSIM, where we do not train a topology predictor, only the TE branch is added in this setting.

For training, we follow the official DrivoR configuration and keep the original optimization settings unchanged. For the standard NAVSIM-v1 model, we use the released recipe of 25 epochs, batch size 16, and AdamW with base learning rate $2\times10^{-4}$ on 4 GPUs; for NAVSIM-v2, we use the corresponding official 10-epoch recipe with the same optimizer and batch size. 
When using the SimScale mixed-training setting, we follow the official 30-epoch training schedule. 
The additional TE head is optimized with the same supervision as in Sec.~\ref{app:Regression-Based: LTF}, i.e., Gaussian heatmap targets and focal loss, and is added on top of the original DrivoR losses without modifying the base architecture or training protocol.

\subsection{Unified Transformer: DriveTransformer}
We integrate our method into \textbf{DriveTransformer}~\cite{jia2025drivetransformer}, a unified sparse-token framework in which \emph{agent}, \emph{map}, and \emph{planning} task tokens interact through task self-attention, sensor cross-attention, and temporal cross-attention, and jointly support detection, prediction, online mapping, and planning. In the Bench2Drive setting, the benchmark further provides instance-level \textbf{traffic-light} and \textbf{traffic-sign} annotations together with lane-level \textbf{HD-map topology}, including lane adjacency, left/right lane relations, and lane identifiers.

Since DriveTransformer already contains traffic-aware detection / mapping supervision, we keep its original traffic-element modeling unchanged and only add \textbf{topology conditioning}. Concretely, we first extract the ego-relevant topology from the lane graph and the lane--traffic relations, and convert it into a compact structured-language description. This text is encoded by a frozen BERT-base encoder into a topology token, which is appended to the original task-token set. The resulting token sequence is then processed by the original DriveTransformer blocks, so that the topology token can interact with the ego/planning token as well as the map and agent tokens through task self-attention. This design preserves the native sparse-token architecture and injects lane connectivity and rule constraints with minimal architectural changes.

For training, we keep the official DriveTransformer optimization settings and all original losses unchanged. The BERT encoder is frozen, and the topology-conditioning path introduces no additional modification to the base detection, prediction, online mapping, or planning heads. When predicted topology is used, it is obtained from a separately trained topology predictor and treated as an external conditioning signal during DriveTransformer training and inference.

\section{Additional Visualization}
\label{apd:additional visualization}

\begin{figure}
    \centering
    \includegraphics[width=1.0\linewidth]{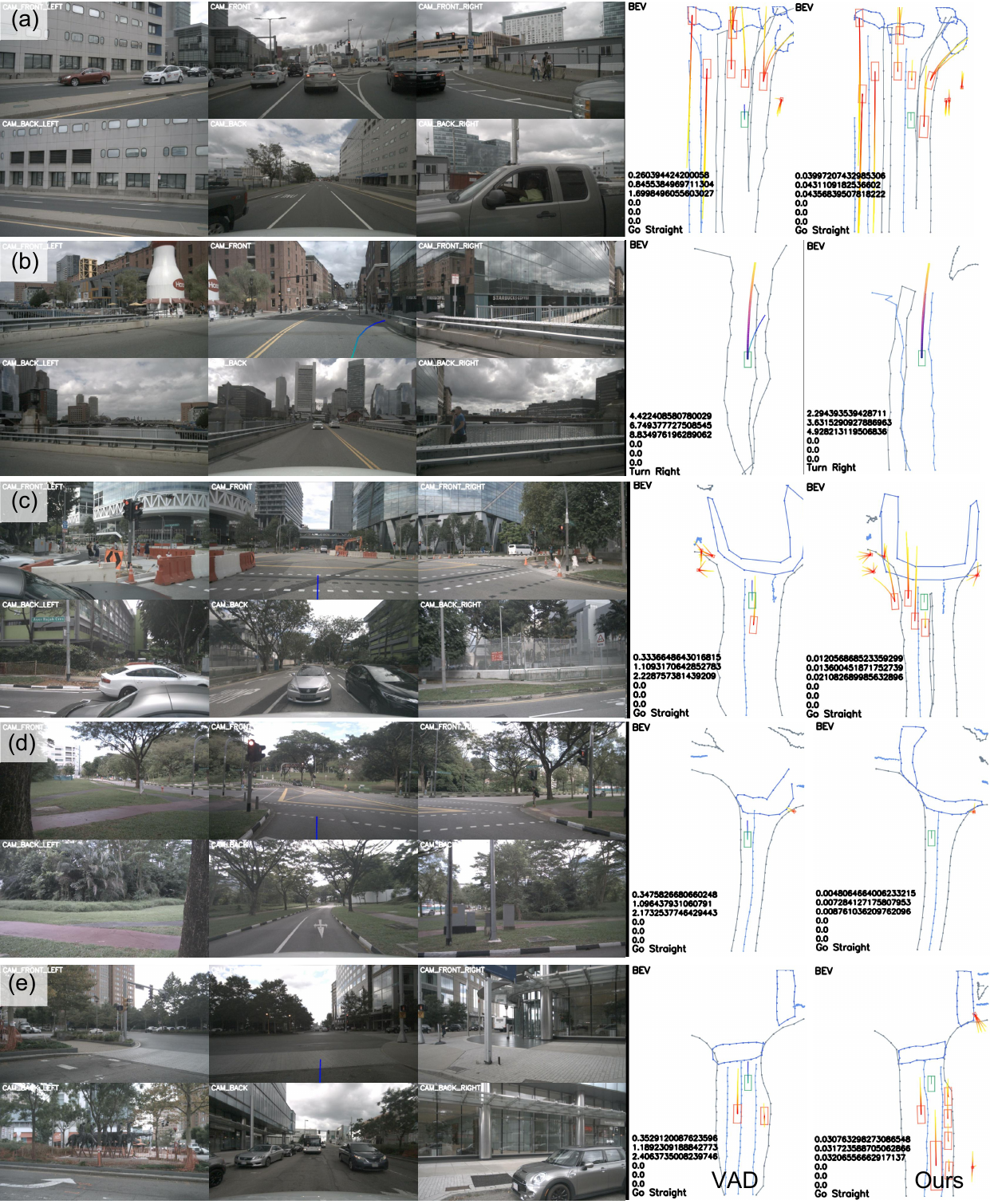}
    \caption{Additional qualitative comparison between VAD and Ours on five traffic-element-rich scenes (a)-(e) from the \textbf{nuScenes} dataset. For each case, the \textbf{left} shows the multi-view camera images, and the \textbf{right} shows the corresponding BEV planning visualizations of VAD and Ours. From top to bottom, the overlaid numbers denote the per-frame \textbf{L2 error} at 1s/2s/3s, \textbf{collision rate} at 1s/2s/3s, and the \textbf{driving command}, respectively.}
    \label{fig:app-nuscenes VAD vis}
\end{figure}

We first provide additional qualitative comparisons between \textbf{VAD} and \textbf{Ours} on five traffic-element-rich scenes from \textbf{nuScenes} (Fig.~\ref{fig:app-nuscenes VAD vis}). In case (a), the ego vehicle approaches a \emph{red traffic light}. Although the correct behavior is to stop, VAD still predicts a forward motion, while our method remains stationary. In case (b), the scene contains a \emph{yellow light} and a right-turn maneuver. VAD turns too aggressively and crosses the map boundary, whereas our prediction performs only a slight right turn and remains well aligned with the ground-truth trajectory. In cases (c)--(e), the scene again contains a \emph{red light}; similar to (a), our method consistently exhibits braking/stopping behavior even when the high-level command is \emph{go straight}. 

\begin{figure}
    \centering
    \includegraphics[width=1.0\linewidth]{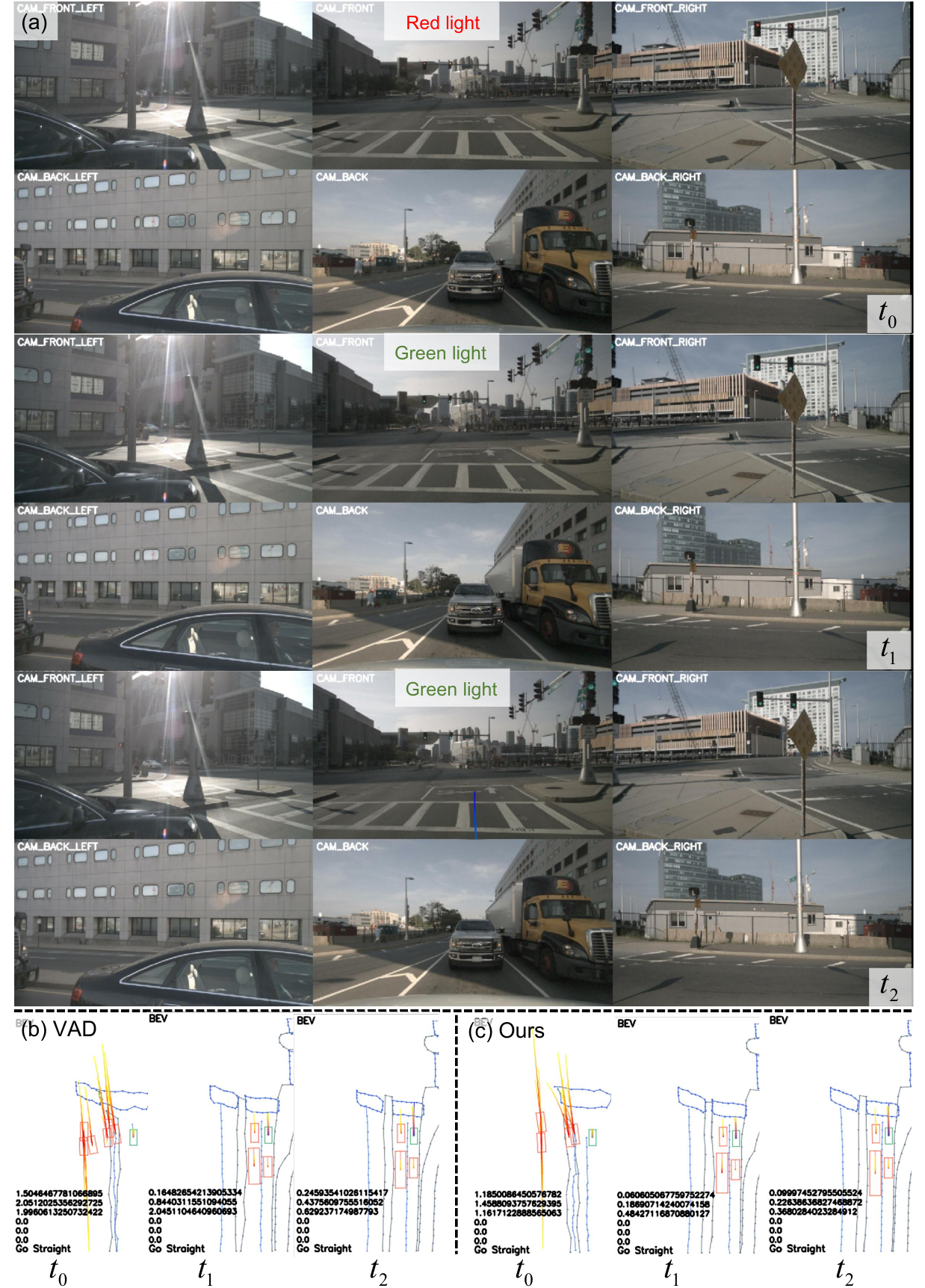}
    \caption{Temporal qualitative comparison on \textbf{nuScenes}. (a) Multi-view images at three consecutive time steps $t_0$, $t_1$, $t_2$, where the forward traffic light changes from \textbf{red to green}. (b) BEV planning trajectories predicted by VAD over the same three frames. (c) Corresponding BEV planning trajectories predicted by Ours.}
    \label{fig:app-nusenes VAD time series vis}
\end{figure}

We further visualize a temporal nuScenes intersection case in Fig.~\ref{fig:app-nusenes VAD time series vis}, where the traffic light changes from \emph{red} at \(t_0\) to \emph{green} at \(t_1\) and \(t_2\), while the high-level command remains \emph{go straight} throughout. In (b), VAD continues to predict forward motion even at \(t_0\) under the red light, and its initial velocity is also inconsistent with the ground-truth stop-to-go behavior. In contrast, (c) shows that our method, benefiting from TE-aware planning, produces substantially lower trajectory error and exhibits a more realistic motion profile: it stays closer to the stopped state when the signal is red, and then gradually increases speed after the light turns green. This example suggests that explicit traffic-element awareness helps the planner better capture both traffic-rule compliance and the underlying vehicle kinematics in temporally evolving scenes.

These examples highlight a key limitation of the baseline: without explicit traffic-element awareness, VAD may ignore rule-critical cues and produce rule-inconsistent trajectories, which also leads to larger trajectory error. In contrast, by explicitly supervising traffic elements and injecting the resulting information into the planning transformer, our method learns a more rule-aware intermediate representation and produces safer, more compliant plans.

\begin{figure}
    \centering
    \includegraphics[width=1.0\linewidth]{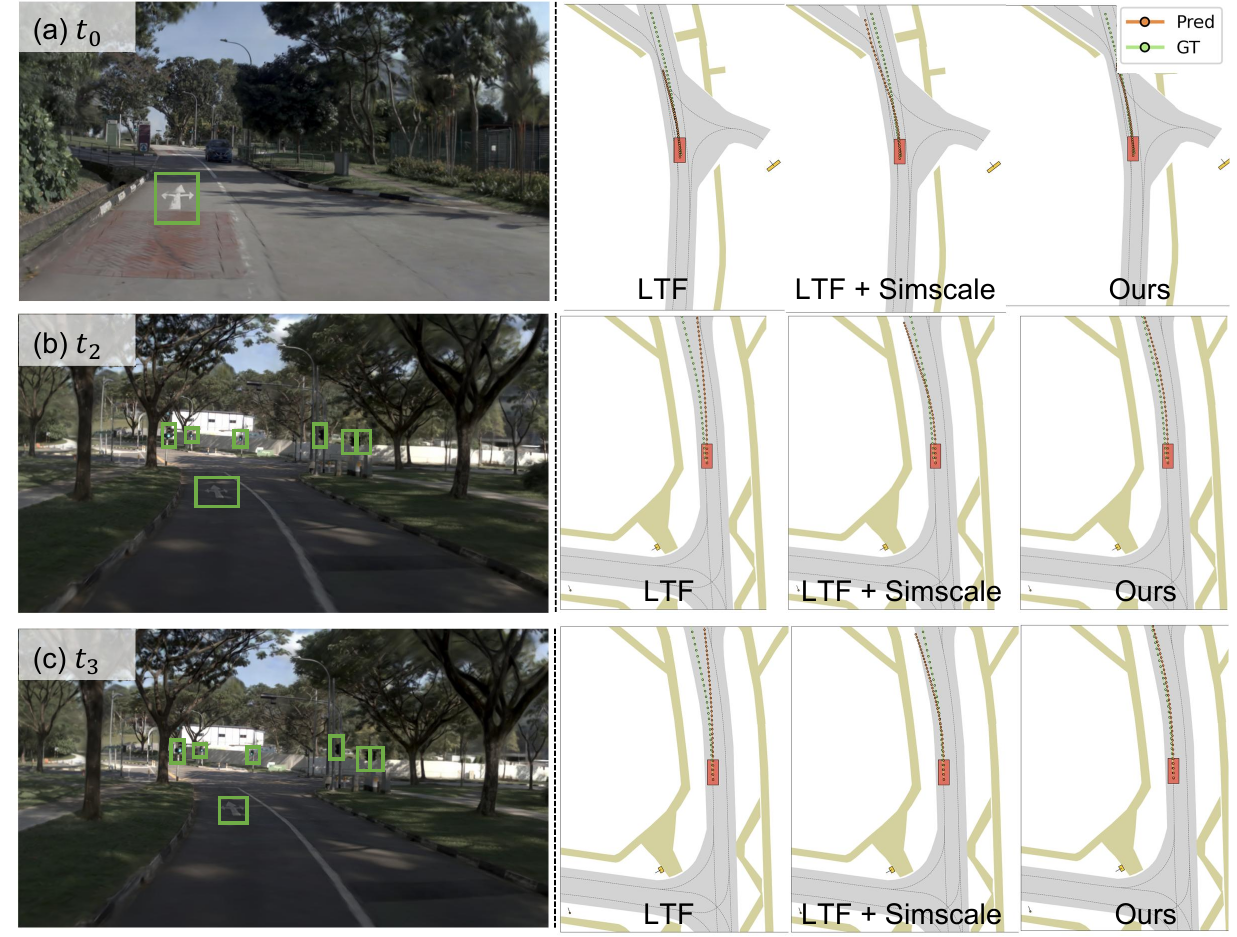}
    \caption{Temporal qualitative comparison on \textbf{NAVSIM}. The \textbf{left} shows front-view images at three consecutive time steps, where the \textbf{green boxes} indicate the detected traffic elements. The \textbf{right} compares the corresponding planning trajectories of LTF, LTF+SimScale, and Ours.}
    \label{fig:app-navsim LTF vis}
\end{figure}

We next show a temporal \textbf{NAVSIM} example in Fig.~\ref{fig:app-navsim LTF vis}, where the ego vehicle drives along a slightly curved road. From the front-view images, the scene contains clear rule-critical cues, including a \emph{straight-ahead road sign} and traffic lights. Benefiting from explicit TE detection, \textbf{Ours} follows the road geometry and maintains a stable, lane-centered trajectory across all three frames. In contrast, both \textbf{LTF} and \textbf{LTF+SimScale} gradually drift away from the lane center, eventually deviating toward an incorrect branch or crossing the road boundary. This example illustrates that TE-aware planning provides useful local constraints for maintaining lane-consistent behavior even in seemingly simple but geometrically ambiguous road segments.
 
\begin{figure}
    \centering
    \includegraphics[width=1.0\linewidth]{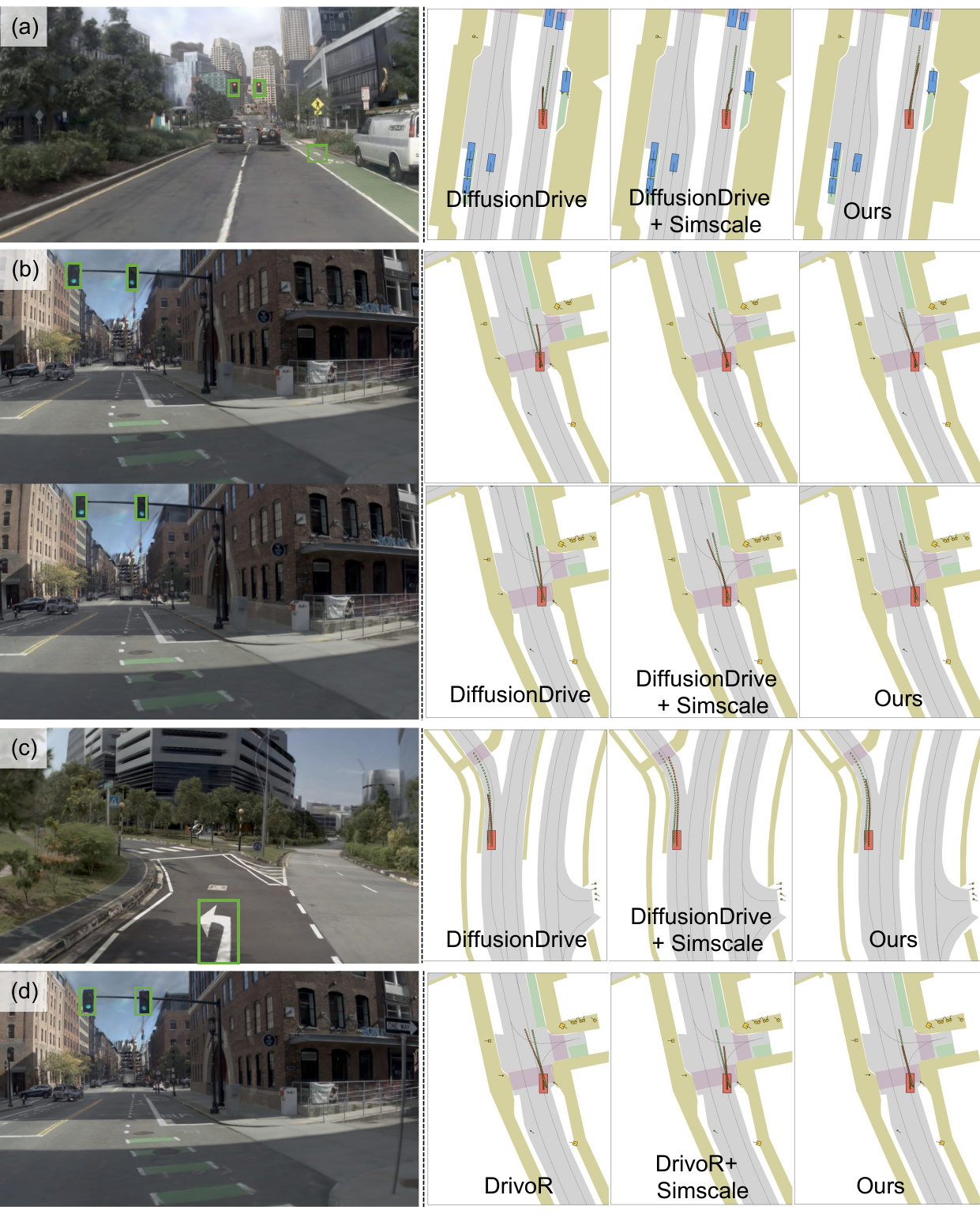}
    \caption{Additional qualitative comparison on \textbf{synthetic NAVSIM-v2 scenes}. The \textbf{left} shows front-view images with detected traffic elements (green boxes), and the \textbf{right} compares the corresponding planning trajectories. Cases (a)–(c) compare DiffusionDrive, DiffusionDrive+SimScale, and Ours, while (d) compares DrivoR, DrivoR+SimScale, and Ours.}
    \label{fig:app-navsim diffusiondrive drivor vis}
\end{figure}

Fig.~\ref{fig:app-navsim diffusiondrive drivor vis} further demonstrates the advantage of our method on two additional backbones, \textbf{DiffusionDrive}~\cite{liao2025diffusiondrive} and \textbf{DrivoR}~\cite{kirby2026drivoR}, in traffic-element-rich synthetic NAVSIM-v2 scenes. By leveraging explicit traffic-element cues to condition planning, our method produces trajectories with more appropriate heading and speed, leading to safer and more rule-consistent behavior than the corresponding baselines.

\section{Ablation Study Details}
\label{apd:ablation study details}
This section provides the detailed experimental settings and implementation choices for each ablation study, clarifying the exact design of every compared variant.
\subsection{Ablation on Traffic-Element Representations for Planning Details}

(1) \textbf{TE(2D).} We introduce a 2D detection branch on top of the baseline as an auxiliary task. Specifically, the model predicts the 2D center locations of traffic elements in the image plane. 
The detection branch follows a DETR-style~\cite{carion2020end} formulation, where a set of learnable queries are used to predict the 2D coordinates of traffic elements directly from image features. Each query outputs a predicted center point along with its classification score.

\noindent (2) \textbf{Depth(FV).} We introduce a depth prediction branch as an auxiliary task. 
Specifically, we use the depth values predicted by a pretrained depth estimation model as pseudo labels for supervision. Based on the baseline model, a convolutional prediction head is applied to the image features to estimate the depth value for each pixel in the image plane. The predicted depth map is supervised using the L1 loss against the pseudo depth labels.

\noindent (3) \textbf{LiDAR.} We estimate the 3D positions of traffic elements directly from LiDAR point clouds. Specifically, the LiDAR points are first projected onto the image plane using the projection matrix. For each traffic element, we collect the LiDAR points that fall inside its corresponding 2D bounding box. Then we apply a clustering algorithm to the collected LiDAR points to separate different point groups. Among the resulting clusters, the cluster that is closest to the ego vehicle is selected. The centroid of this cluster is then computed and used as the 3D center of the corresponding traffic element.

\noindent (4) \textbf{TE}. We utilize DepthAnythingV3~\cite{lin2025depthanything3} as the depth estimation model to predict dense depth maps from input images. The predicted depth values are used to estimate the 3D positions of traffic elements based on their corresponding image locations.

\noindent (5) \textbf{TL}. The traffic element prediction branch is trained using only traffic light annotations. Specifically, the supervision includes three traffic light states: red, yellow, and green.

\subsubsection{Yellow-light subset.}
We further verify that yellow lights are explicitly modeled as traffic-element states rather than being ignored or collapsed into a binary red/green setting. Using OpenLane-V2 ground-truth traffic-element annotations on nuScenes, we extract a small yellow-light subset and compare VAD with our traffic-element-aware model. As shown in Tab.~\ref{tab:yellow_light_subset}, our method improves the trajectory error over VAD on this subset, reducing L2@3s from 1.25 to 0.98. This indicates that the traffic-element branch can use the yellow-light state as a distinct regulatory cue for planning.

\begin{table}[t] \centering \caption{ Yellow-light subset on nuScenes. We report open-loop planning metrics on samples containing yellow-light traffic elements. } \label{tab:yellow_light_subset} \resizebox{0.72\linewidth}{!}{ \begin{tabular}{lcccccc} \toprule Method & L2@1s$\downarrow$ & L2@2s$\downarrow$ & L2@3s$\downarrow$ & Col@1s$\downarrow$ & Col@2s$\downarrow$ & Col@3s$\downarrow$ \\ \midrule VAD & 0.49 & 0.85 & 1.25 & 0.00 & 0.00 & 0.00 \\ Ours & 0.37 & 0.63 & 0.98 & 0.00 & 0.00 & 0.00 \\ \bottomrule \end{tabular} } \end{table}

We note that this evaluation is limited by the small number of yellow-light samples in the dataset. Moreover, our current traffic-element module reasons mainly from the current traffic-light state and does not explicitly track temporal state transitions. Modeling temporal light-state evolution is an important direction for future work.

\subsection{Design Choices for Traffic Element Integration Details}

(1) \textbf{Prediction Head and Loss Function.} Without using an independent branch for traffic element prediction, the traffic element category is incorporated into the original BEV segmentation task. Specifically, traffic elements are treated as additional semantic classes within the BEV segmentation head, and the number of predicted classes is increased from 7 to 20 (7+13), where the additional 13 classes correspond to traffic elements. Using the CE loss treats traffic element prediction as a multi-class classification problem. 
Specifically, traffic element categories are predicted using a multi-class cross-entropy loss. Considering the class imbalance between foreground and background categories, we also explore the use of the Focal Loss for supervision.

\noindent (2) \textbf{Pooling and  Interaction Mechanism.} We investigate different pooling strategies when downsampling the predicted traffic element heatmap. Specifically, the predicted traffic element heatmap is downsampled to match the spatial resolution of the BEV feature map using either max pooling or average pooling. Using cross-attention for interaction between predicted traffic elements and planning means that the predicted traffic element features are used as the keys and values. The trajectory queries attend to the traffic element features through the cross-attention mechanism, and the resulting features are used to generate the final trajectory predictions.

\subsection{Rule-based and Robustness Analyses}
\subsubsection{Rule-based traffic-element baselines.}
We compare our learned traffic-element integration with two simple rule-based post-processing baselines on nuScenes using VAD. The first baseline stops for any red light within 30 meters, without checking whether the light governs the ego lane. The second baseline uses topology to stop only for ego-lane-associated red lights. As shown in Tab.~\ref{tab:rule_based_te}, the naive rule without topology hurts trajectory accuracy and increases long-horizon collision rate, because many traffic lights in the scene are irrelevant to the ego vehicle. Adding topology improves the rule-based baseline by filtering out irrelevant lights, but it still underperforms our learned TE/topology integration. This confirms that the gain of our method does not come from a trivial stop rule; instead, it comes from learning a general traffic-element-aware planning signal that can handle more complex situations than fixed thresholding.

\begin{table}[t] \centering \caption{ Rule-based traffic-element baselines on nuScenes with VAD. ``Rule w/o Topo'' stops for any red light within 30 meters, while ``Rule w/ Topo'' stops only for ego-lane-associated red lights. } \label{tab:rule_based_te} \resizebox{0.78\linewidth}{!}{ \begin{tabular}{lcccccc} \toprule Method & L2@1s$\downarrow$ & L2@2s$\downarrow$ & L2@3s$\downarrow$ & Col@1s$\downarrow$ & Col@2s$\downarrow$ & Col@3s$\downarrow$ \\ \midrule VAD & 0.41 & 0.70 & 1.05 & 0.07 & 0.17 & 0.41 \\ Rule w/o Topo & 0.57 & 0.97 & 1.43 & 0.02 & 0.38 & 0.76 \\ Rule w/ Topo & 0.42 & 0.71 & 1.07 & 0.02 & 0.24 & 0.35 \\ Ours & 0.34 & 0.56 & 0.92 & 0.04 & 0.21 & 0.26 \\ \bottomrule \end{tabular} } \end{table}

\subsubsection{Robustness corruption protocol.}
To analyze robustness to upstream traffic-element perception noise, we conduct inference-stage sensitivity tests on NAVSIM-v2 with LTF. Starting from the predicted traffic-element representation, we apply three types of corruptions before feeding the representation to the planner: (i) depth noise, which perturbs the estimated traffic-element depth; (ii) missed detections, which randomly remove a portion of predicted traffic elements; and (iii) false positives, which inject additional spurious traffic elements into the BEV representation. We vary a corruption severity parameter for each failure mode and report the resulting EPDMS in Fig.~\ref{fig:te_robustness}. The robustness curves show that our method degrades gracefully under common traffic-element perception failures. This is important because, on datasets without native traffic-element annotations, the planner uses automatically detected and lifted traffic elements rather than perfect ground-truth inputs. Together with the cross-dataset detector validation in Tab.~\ref{tab:navsim_manual_te_eval}, this analysis indicates that our planning improvements are not overly dependent on oracle-quality traffic-element perception.

\subsection{Ablation on Topology Information Integration Details}

(1) \textbf{Topology Synergy.} When only $\mathbf{R}_{\text{LCTE}}$ is used, the model retrieves traffic elements associated with the centerline that is closest to the ego vehicle, without exploring the topology relations between neighboring centerlines and traffic elements. When only $\mathbf{R}_{\text{LCLC}}$ is used, the model considers only the connectivity between centerlines and retrieves neighboring centerlines through the centerline graph, while traffic element relations are not utilized.

\noindent (2) \textbf{Encoding Strategy.} To encode the topology subgraph, we adopt a GCN. 
The topology graph consists of two types of nodes: centerlines and traffic elements. 
For centerline nodes, the feature representation includes seven components: the mean $(x, y)$ coordinates on the BEV plane, the orientation, the length, the average curvature, a binary indicator of whether the centerline belongs to an intersection, and a binary indicator of whether the lane is the closest centerline to the ego vehicle. For traffic element nodes, the feature representation is constructed using one-hot encodings of their categories and attributes. The centerline features and traffic element features are used as the initial node embeddings and are fed into the RGCN~\cite{schlichtkrull2018gcn} to encode the topology relations within the subgraph.

\noindent (3) \textbf{Interaction Scope.} When using GCN to encode the topology subgraph, we explore two different strategies to obtain the final graph representation. For the \text{global} setting, the node features produced by the GCN are aggregated using average pooling over all nodes to obtain a global graph representation. For the \text{ego} setting, we directly use the feature of the centerline node that is closest to the ego vehicle as the output representation of the topology graph.

\noindent (4) \textbf{Robustness to Predicted Topology.} We replace the ground-truth topology with the topology predicted by the TopoMLP~\cite{wu2023topomlp} model. Specifically, the centerline-centerline and centerline-traffic-element relations are obtained from the predictions instead of the ground-truth topology annotations. The predicted topology is then used as the global topology graph for subsequent topology extraction and encoding.

\subsubsection{Topology encoder variants.}
In the main experiments, we encode ego-relevant topology with a frozen BERT encoder. To test whether the benefit depends on using a heavy language model, we compare BERT with a lighter DistilBERT encoder~\cite{sanh2019distilbert} and a Graph Transformer encoder~\cite{dwivedi2020graphtransformer} on nuScenes using the VAD backbone. As shown in Tab.~\ref{tab:topology_encoder_variants}, DistilBERT achieves comparable planning performance to BERT while slightly improving throughput. In contrast, the Graph Transformer performs worse in both L2 error and collision rate. This suggests that the benefit of our topology conditioning is not tied to a specific large BERT encoder; rather, language-style encoding provides a compact and effective way to represent heterogeneous lane--traffic-element relations.

\begin{table}[t] \centering \caption{Topology encoder variants on nuScenes with the VAD backbone. DistilBERT achieves performance comparable to BERT at slightly lower cost, while Graph Transformer is less effective. } \label{tab:topology_encoder_variants} \resizebox{0.84\linewidth}{!}{ \begin{tabular}{lccccccc} \toprule Encoder & L2@1s$\downarrow$ & L2@2s$\downarrow$ & L2@3s$\downarrow$ & Col@1s$\downarrow$ & Col@2s$\downarrow$ & Col@3s$\downarrow$ & FPS$\uparrow$ \\ \midrule BERT & 0.34 & 0.59 & 0.92 & 0.04 & 0.21 & 0.26 & 5.4 \\ DistilBERT & 0.35 & 0.61 & 0.94 & 0.05 & 0.15 & 0.26 & 5.6 \\ Graph Transformer & 0.39 & 0.65 & 0.99 & 0.11 & 0.19 & 0.34 & 5.5 \\ \bottomrule \end{tabular} } \end{table}

The weaker performance of the Graph Transformer may come from the heterogeneous nature of the topology graph. Centerline nodes encode continuous geometry, while traffic-element nodes encode discrete regulatory states and attributes. Language-style encoders naturally serialize these heterogeneous attributes into a unified semantic sequence, whereas graph message passing can over-smooth node representations or dilute the discrete rule-control signals.

\section{Limitations of nuScenes Metrics}
\label{apd:addtional nuscens metrics}
\begin{figure}
    \centering
    \includegraphics[width=1.0\linewidth]{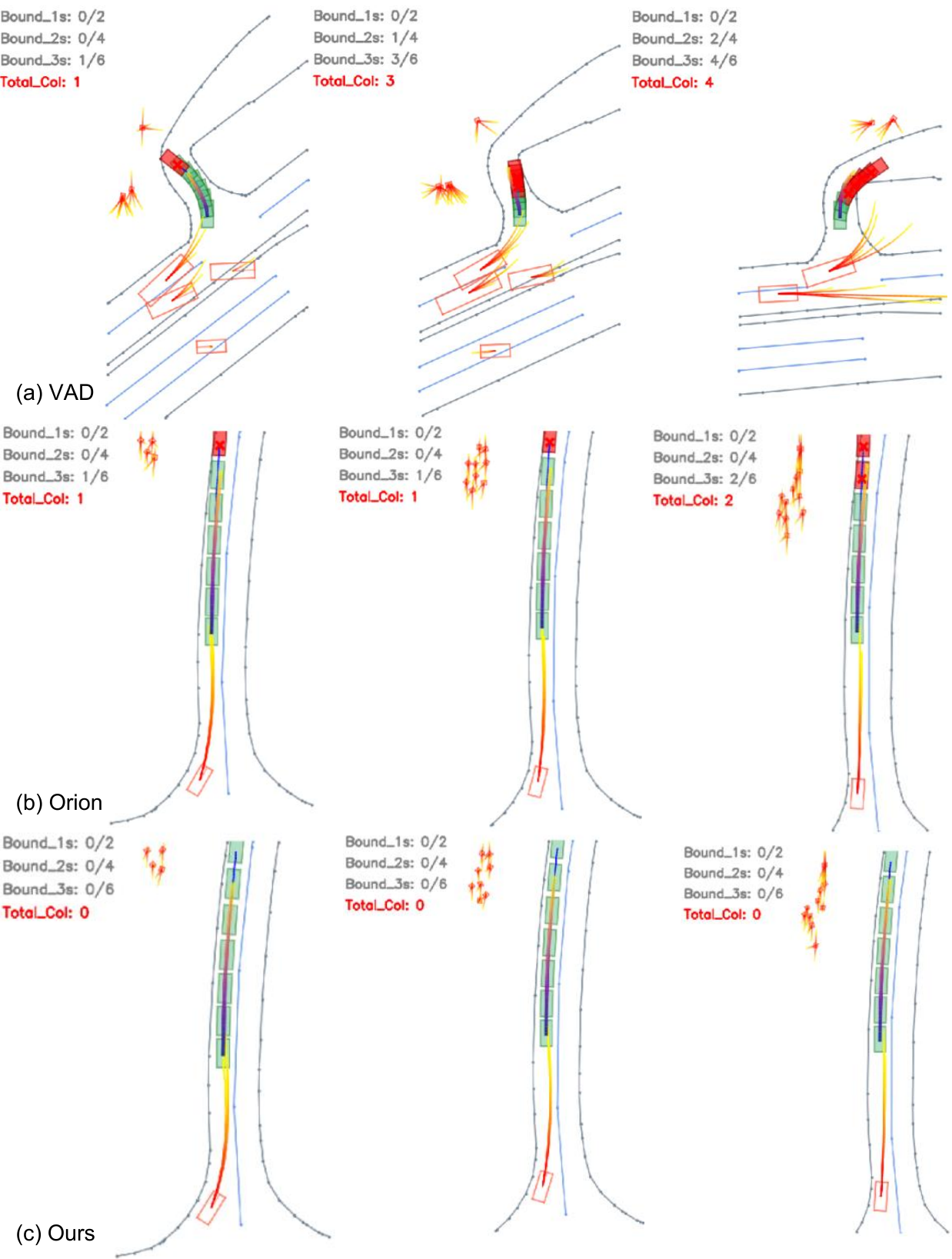}
    \caption{Illustration of a \textbf{limitation of the standard nuScenes metrics}. Using the \textbf{ground-truth HD map}, we additionally count the number of collisions between the predicted ego-vehicle boxes and the lane/map boundaries (shown in the upper-left corner of each panel). (a) VAD, (b) Orion, and (c) Ours. Even in these simple scenes, Ours yields zero boundary collisions, while the other methods still produce boundary violations, showing that L2 and collision rate alone may not fully capture map-consistent planning quality.}
    \label{fig:app-nuscenes boundary metric vis}
\end{figure}

During evaluation, we observe an important limitation of the standard nuScenes metrics. The reported collision rate mainly measures whether the predicted future trajectory collides with other dynamic objects within the prediction horizon. However, this metric does not capture another clearly undesirable behavior: \emph{driving out of the valid drivable region or crossing map boundaries}. As illustrated in Fig.~\ref{fig:app-nuscenes boundary metric vis}, such boundary violations can still yield a zero collision rate, despite being obviously unsafe and map-inconsistent.

To better reflect this failure mode, we additionally count \emph{boundary collisions}, defined as intersections between the predicted ego boxes and the HD-map boundary. Quantitative results in Tab.~\ref{tab:app_boundary_collision} show that our method consistently reduces boundary collision rate at all horizons. The qualitative examples in Fig.~\ref{fig:app-nuscenes boundary metric vis} show that even in simple straight-road scenarios, baseline methods may still collide with the boundary, whereas our method produces zero boundary collisions. We attribute this to the introduced \textbf{topology cues}, which provide stronger lane-level structural constraints and lead to safer, more map-consistent planning.

\begin{table}[t]
\centering
\small
\setlength{\tabcolsep}{6pt}
\renewcommand{\arraystretch}{1.1}
\caption{\textbf{Boundary collision rate} on nuScenes. Lower is better.}
\label{tab:app_boundary_collision}
\begin{tabular}{lcccc}
\toprule
Method & \multicolumn{4}{c}{Boundary Collision Rate (\%) $\downarrow$} \\
\cmidrule(lr){2-5}
 & 1s & 2s & 3s & Avg. \\
\midrule
VAD  & 1.09 & 1.81 & 2.85 & 1.92 \\
\rowcolor{myblue}
Ours & \textbf{1.01} & \textbf{1.39} & \textbf{1.93} & \textbf{1.44} \\
\bottomrule
\end{tabular}
\end{table}

\section{Limitations \& Future Work}
\label{apd:limitations & future work}
Our method still depends on the quality of upstream \textbf{traffic-element detection} and \textbf{depth estimation}, especially on datasets without native TE annotations where pseudo labels are required. Errors in either stage may propagate to the constructed 3D TE representation and weaken the downstream planning benefit. A natural next step is to move from the current staged pipeline toward \textbf{joint end-to-end optimization}, so that 2D TE detection, depth estimation, 3D TE lifting, and planning can be trained together under planning-oriented supervision.

In its current form, our framework conditions planning mainly on the \textbf{current TE and topology state}. While effective, this design does not explicitly model temporal evolution, which can be important when the scene state changes over time, such as traffic-light transitions or temporally ambiguous rule cues. Future work could incorporate \textbf{historical TE/topology memory} or temporal consistency modules, allowing the planner to reason over state transitions rather than only instantaneous observations.

Our current formulation focuses on traffic lights, traffic signs, and lane topology, which already provide strong rule-critical cues, but the semantic scope remains limited. Many \textbf{additional scene factors} may also be relevant for safe and compliant planning, such as lane markings, stop lines, temporary traffic control, construction cues, or richer map semantics. Extending the framework to a broader set of structured scene semantics is therefore an important direction.

Finally, although we have validated the method across multiple benchmarks and planners, the closed-loop evaluation scale is still limited compared with the diversity of real-world long-tail driving. In particular, it remains valuable to test whether the gains from TE and topology continue to grow with \textbf{larger-scale training}, \textbf{more challenging corner cases}, and \textbf{long-tail rule-critical scenarios}, and whether these benefits remain stable under cross-dataset transfer. This motivates future work on larger closed-loop data generation and targeted scenario construction for stress-testing TE- and topology-aware planning.

\end{document}